\documentclass[lettersize,journal]{IEEEtran}
\usepackage{amssymb,amsmath,amsfonts}
\usepackage{algorithmic}
\usepackage{array}
\usepackage{textcomp}
\usepackage{stfloats}
\usepackage{url}
\usepackage{verbatim}
\usepackage{graphicx}

\usepackage[hidelinks]{hyperref}
\usepackage{float}
\usepackage{subcaption}
\usepackage[dvipsnames]{xcolor}
\usepackage{enumitem} 
\usepackage{pdflscape}
\usepackage{pdfpages}
\usepackage{booktabs}
\graphicspath{{figures/}}

\newcommand{\question}[1]{\item \textbf{#1}}

\def\BibTeX{{\rm B\kern-.05em{\sc i\kern-.025em b}\kern-.08em
    T\kern-.1667em\lower.7ex\hbox{E}\kern-.125emX}}
\usepackage{balance}
\begin{document}
\title{Facial Emotion Recognition does not detect feeling unsafe in automated driving}
\author{Abel van Elburg, Konstantinos Gkentsidis, Mathieu Sarrazin, Sarah Barendswaard, Varun Kotian, and Riender Happee
\thanks{Manuscript created June, 2025;}
\thanks{A. van Elburg, V. Kotian and R. Happee are with the Faculty of 
Mechanical Engineering, Delft University of Technology, 
Delft, Zuid-Holland, The Netherlands.}
\thanks{K. Gkentsidis, M. Sarrazin and S. Barendswaard are with the Siemens Digital Industries Software,
Leuven, Flemish Brabant, Belgium.}

}

\markboth{}%
{Abel van Elburg, Konstantinos Gkentsidis, Mathieu Sarrazin, Sarah Barendswaard, Varun Kotian, Riender Happee :
Facial Emotion Recognition does not detect feeling unsafe in automated driving}

\maketitle

\begin{abstract}
Trust and perceived safety play a crucial role in the public acceptance of automated vehicles.
To understand perceived risk, an experiment was conducted using a driving simulator under two automated driving styles and optionally introducing a crossing pedestrian.
Data was collected from 32 participants, consisting of continuous subjective comfort ratings, motion, webcam footage for facial expression, skin conductance, heart rate, and eye tracking.

The continuous subjective perceived risk ratings showed significant discomfort associated with perceived risk during cornering and braking followed by relief or even positive comfort on continuing the ride. The dynamic driving style induced a stronger discomfort as compared to the calm driving style. The crossing pedestrian did not affect discomfort with the calm driving style but doubled the comfort decrement with the dynamic driving style. This illustrates the importance of consequences of critical interactions in risk perception.

Facial expression was successfully analyzed for 24 participants but most (15/24) did not show any detectable facial reaction to the critical event. Among the 9 participants who did, 8 showed a Happy expression, and only 4 showed a Surprise expression. Fear was never dominant.
This indicates that facial expression recognition is not a reliable method for assessing perceived risk in automated vehicles.

To predict perceived risk a neural network model was implemented using vehicle motion and skin conductance. The model correlated well with reported perceived risk, demonstrating its potential for objective perceived risk assessment in automated vehicles, reducing subjective bias and highlighting areas for future research.
\end{abstract}


\section{Introduction}
\label{sec:introduction}

Comfort, and more specifically, perceived safety and trust are key in the public adoption of vehicle automation. 
The transition to automation, where the driver becomes a passenger, highlights driving style as a crucial element influencing perceived safety and trust. Interfaces informing users of vehicle automation operation can also enhance perceived safety and trust. To design automated vehicle control systems and interfaces to enhance perceived safety and trust, we need validated methods to assess these in the context of automated driving. 
As outlined in section \ref{subsec:assessment} subjective evaluation is still the golden standard to assess perceived safety and trust. Questions asked during and after experimental sessions can disclose multiple aspects related to perceived risk, trust in automation, understanding of automation, and acceptance including the intention to use automation. However subjective measures are prone to bias, and their collection may influence the user perception in an undesired manner. Hence, objective measures provide a compelling additional approach to measure perceived risk. Several physiological parameters have been shown to correlate with perceived risk. 
Galvanic Skin Response (GSR) has proven to be a reliable indicator for arousal and stress \cite{dawson2007electrodermal, jaiswal2023gsr}. This correlation has also been confirmed in the context of driving \cite{wang2019detection, memar2021stress}, making GSR a promising objective measure of perceived risk. However, GSR responds similarly to different types of arousal such as feeling unsafe, motion sickness, and workload. The same holds for the measurement of heart activity and pupil diameter. In this respect facial emotion recognition (FER) provides a compelling alternative able to capture multiple emotions. Facial emotion recognition is often based on the discrete set of universal emotions defined by 
Paul Ekman \cite{ekman1999basic}. These emotions are expressed with the same facial expressions for all humans, and this set consists of anger, fear, disgust, happiness, surprise, and sadness. Especially fear and surprise are of interest, as these could indicate a person feeling unsafe or uncomfortable. However, as outlined in section \ref{subsec:assessment}, only a few studies explored FER in the context of automated driving.

This paper contributes to this research area by collecting a comprehensive dataset and analyzing subjective comfort ratings, facial expressions, and GSR aiming to demonstrate the effectiveness of FER in measuring feeling unsafe in automated driving and using GSR as an objective measurement for overall comfort.

An experiment was conducted in a driving simulator, where participants were subjected to a scenario as if being driven by an automated vehicle. Perceived risk was varied with a 2x2 within-subject experimental design with a calm and dynamic driving style and stopping for a stop sign without a pedestrian and with a crossing pedestrian.
We measured continuous perceived comfort via a knob, a camera recorded the face, and physiological data was collected. In particular, we explored whether fear and surprise, as detected with facial emotion recognition, represent periods of high perceived risk. Also we explored if GSR could be used as an alternative to subjective measurements.

\section{State of the art}
\label{sec:background}
This section reviews the literature on facial expressions and emotions (section \ref{subsec:emotions}) and current assessment methods of perceived safety and trust in automated driving (section \ref{subsec:assessment}). 

\subsection{Facial emotion recognition}
\label{subsec:emotions}
As humans, we can intuitively read a human face and recognise emotions such as sadness or anger, even in people we never met before. To label facial features representing emotional state, Paul Ekman proposed the Facial Action Coding System (FACS) \cite{ekman1978facial}, which is widely used in the field of computer vision \cite{davoli2020driver}. 


For labelling emotions from facial expressions, one of the most considered sets of general physiologically distinct emotions is proposed by Paul Ekman: anger, fear, disgust, happiness, surprise, and sadness \cite{davoli2020driver, ekman1999basic}. Ekman argues that the facial expressions that display these emotions are universal among all humans, no matter when or where they grew up. This approach has been implemented, resulting in "discrete classification" where emotion is either classified as neutral or one of the above six emotions. The discrete classification approach is widely accepted, but categorizing facial expressions into discrete basic emotions is not undisputed. An important argument is that emotions vary in intensity and often blend together, making discrete classification incomplete \cite{savran2013automatic, leng2007experimental}. 

A popular alternative is the circumplex model of affect \cite{russell1980circumplex}. This 2D  model features arousal on one axis and valence on the other, where arousal represents the intensity, and valence represents the pleasantness of emotion \cite{davoli2020driver}. 
\cite{cai2011modeling} used the arousal-valence model to study the effects of emotion on driving task performance and reported a strong non-linear effect of both arousal and valence where performance drops with extremely positive and negative self-assessed emotions. 
We found no study using facial expression to detect arousal and valence and therefore did not explore the circumplex model in this paper. 

Regarding expressed emotions in real-world situations, research indicated another important concept, commonly referred to as "display rules" \cite{ekman1969pan, matsumoto1990cultural, safdar2009variations}. Display rules are procedures learned to manage facial expressions for specific emotions, i.e. the displayed arousal \cite{ekman1969pan}. Research has shown that display rules can differ significantly between cultures, with multiple experiments showing differences between for example Japanese and North American culture when it comes to the intensity of displaying different emotions \cite{matsumoto1990cultural, safdar2009variations}. 
Research on display rules also tends to make use of the basic set of emotions as defined by Ekman. So the universality of this set seems generally accepted. 
Display rules represent an active (aware) display of emotions and can, therefore, be considered to represent subjective rather than objective measures of emotions. Display rules may be relevant in traffic in eye-contact but seem less suitable to reflect perceived risk in automated driving. Hence display rules are not explored in this paper.

\subsection{Assessment of perceived safety}
\label{subsec:assessment}

The measurement of perceived safety and trust in automated vehicles is predominantly subjective and often through questionnaires during or after experiencing automated driving in a simulator, experiencing a physically automated vehicle or receiving a description of automation without specific experience. However, questionnaire-based research is subject to various biases, including social desirability, misunderstanding, and cognitive dissonance.

Questionnaires do not provide continuous measurement and may require some interruption of the experiment. To address this, researchers used so-called direct input devices for (semi-) continuous subjective assessment. \cite{saffarian2012drivers} used a slider to measure perceived safety in fog in manual and automated car following. \cite{su2023development} evaluated continuous pressing-based and discrete smartphone-based for comfort assessment in an automated driving simulator, where the discrete approach showed better measurement repeatability and lower measurement bias than the continuous approach. \cite{beggiato2020komfopilot} evaluated perceived discomfort during automated driving in a fixed base simulator with a handheld manual input device. Results show discomfort in critical interactions representing elevated perceived risk. 
\cite{he2022modelling} evaluated continuous perceived risk through finger press force measurement in critical events during automation use in a simulator and found a good correspondence of peak finger press force with post-event perceived risk. 

However, both question-based and continuous perceived risk assessments are subjective and, therefore, prone to bias. In addition their assessment may affect the experiment outcome. 
This underscores the need for objective measurement of perceived risk.

\subsection{Objective Measurements}
A range of studies explored physiological measures providing such objective measures of perceived risk, motion sickness, workload and other states in driving conditions. Below we summarize key findings with a focus on perceived risk in automated driving. 
The Galvanic Skin Response (GSR), is also referred to as Electrodermal Activity (EDA), or Skin Conductance (SC). GSR reflects the electrical conductance of the skin and detects fluctuations caused by the triggering of sweat glands, which are controlled by the sympathetic part of the autonomic nervous system and cannot be controlled consciously. GSR has been shown to correlate with arousal, workload and attention \cite{dawson2007electrodermal}. Opposed to other physiological signals like pupil diameter or heart rate, GSR is not controlled by the parasympathetic system. Because of this, it is a promising metric to reflect fear.
Heart activity can be recorded with surface electrodes at the thorax (electrocardiography - ECG) or by measuring blood flow or skin temperature. In driving studies, in particular, the ECG has been evaluated, and various measures derived from the ECG are found to correlate to stress, mental effort and physical effort. In particular, the heart rate (HR) and heart-rate variability (HRV) show good correlations but require somewhat longer observation windows as compared to GSR making ECG less suitable to detect brief moments of high perceived risk. 
Eye activity can also capture cognitive workload and stress through increased pupil diameter and reduced blink rate.
Brain activity (Electroencephalography - EEG) can detect a wealth of processes and states, but is rarely applied in simulators and cars due to electromagnetic interference and intrusiveness of the caps and electrode arrays used in high-end systems.

The previously mentioned research by \cite{beggiato2020komfopilot} found physiological signals to capture events that provoke moderate to high reported discomfort (e.g. close approach
to vehicle driving ahead, intersection with a fast vehicle approaching from the right, approach to red traffic light, entering highway) but not for slowly evolving and long-lasting situations. In particular, they found increased pupil diameter, reduced blink frequency, and reduced HRV in uncomfortable situations. GSR results were inconclusive and dependent on electrode placement. Facial expression analysis was briefly reported to indicate surprise and tension in close approach situations. 
\cite{he2022modelling} found that
pupil diameter correlated with perceived risk during the more risky events. Heart rate also increased during events but no quantified relation between heart rate and perceived risk magnitude was found \cite{he2022modelling}. 
\cite{perez2025perceived} showed pupil diameter to increase with perceived risk in an experimental automated vehicle.
\cite{park2022social} related EEG to emotional states in a self-driving car experiment. Participants experienced five scenarios to measure initial trust, trust escalation, trust reduction, trust mutation, and trust rebuilding. Both subjective rating and EEG showed substantial changes between these phases, where the EEG seemed to reflect changes in trust. 

Facial emotion recognition can be non-invasive as it does not require electrodes or specific devices attached to the body.
\cite{wintersberger2016automated} analysed facial expression of passengers driven by an automated driving system, a male driver, or a female driver in a moving base driving simulator in critical driving conditions. From the total of 31,859 images 31,180 could successfully be classified (97.8\%). The results showed neutral face expressions for 94.3\% and happiness for 4.4\% of all images.
The remaining 1.3\% were attributed to sadness (0.9\%), surprise (0.2\%), contempt (<0.1\%) and anger (<0.1\%) – disgust
and fear were classified not at all.
These facial expressions differed significantly between the 3 driving styles whereas heart rate variability showed no significant differences.
\cite{sini2020automatic} implemented facial emotion recognition and evaluated emotions in eight participants when using automation in urban conditions by showing driving videos with varying maneuvers. Sadness was reported most, followed by neutrality, happiness and surprise. Fear was never detected.

Summarizing the above, physiological signals and facial expressions all correlate to the driver's state and may reflect perceived risk in automated driving. In this paper, we will explore facial expression since it has hardly been explored, and collect GSR, ECG and eye activity as objective reference to validate facial expression. We also note that the above studies indicate challenges for all physiological data collection, and in defining baseline levels. Therefore, we also measure continuous comfort ratings as a robust but subjective reference.

\section{Methods}
\label{sec:methods}


\subsection{Driving Simulator}

\begin{figure}[h]
    \centering
    \includegraphics[width=0.9\linewidth]{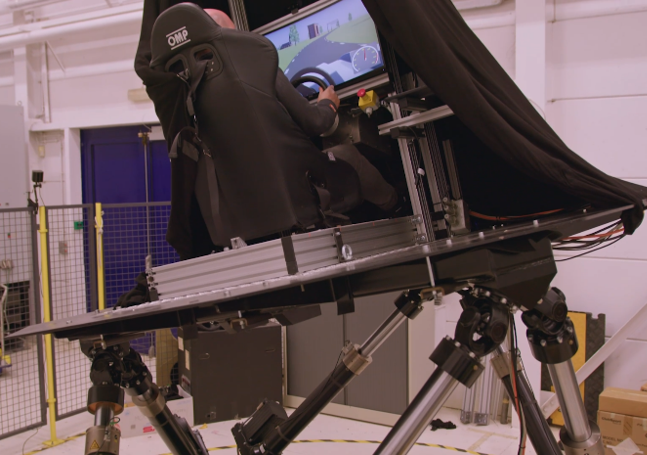}
    \caption{Driving simulator with hexapod, chair and wide-screen.}
    \label{fig:platform}
\end{figure}

The driving simulator used, was built on a MOOG Hexapod, located in the Siemens facility in Leuven, with six degrees of freedom (Figure \ref{fig:platform}). 
The scene is displayed on a wide-screen monitor with a resolution of 3440 by 1440 and a diameter of 34 inch / 86 centimeter. 
On the bottom right there was a big red emergency stop button for the participant. All participants were made aware of this option, but none used it or mentioned feeling a need to do so. 

\subsection{Sensors}

During this experiment, measurements from five different sensors were recorded. 

\subsubsection{Comfort-Knob}
The Comfort-Knob is a rotational potentiometer 
to continuously measure the subjective comfort of the participant (Figure \ref{fig:knob}). It is a turning knob where the participant gives a value between 0 and 10. The scale was defined as 0 being extremely uncomfortable, 5 being neutral, and 10 being extremely comfortable. This is based on \href{https://www.sae.org/standards/content/j1060_201405/}{SAE J1060} for subjective ratings related to ride comfort in motor vehicles, where 0 up to and including 4 is considered unacceptable, 5 is borderline, and 6 to 10 is acceptable \cite{cieslak2020accurate}. Except for the knob not turning past the limits of 0 and 10, there was no haptic feedback. Figure \ref{fig:smiley} was constantly present in the top-middle of the screen to remind the participant which way to turn the knob according to their state. This illustration was made simple  to limit distraction. 
On the device itself is a LED screen that shows the current value in 2 decimals. Participants were instructed not to focus on the exact value, as this would be distracting. 


\subsubsection{Galvanic Skin Response (GSR), heart activity (ECG) \& eye tracking}
The Galvanic Skin Response was measured with two electrodes at the fingers, heart activity was measured with three electrodes at the thorax, and eye gaze and pupil diameter were measured with a head mounted eye racker as detailed in appendix C.

\subsubsection{Webcam}
The face was recorded via a webcam located above the screen (see Figure \ref{fig:firstperson}). 
The frames are in color and saved as a video of 30 frames per second with a resolution of 640 by 480 pixels. 


\subsection{Driving scenario}

\begin{figure*}[h]
    \centering
    \begin{subfigure}[b]{0.5\textwidth}
        \centering
        \includegraphics[width=.9\textwidth]{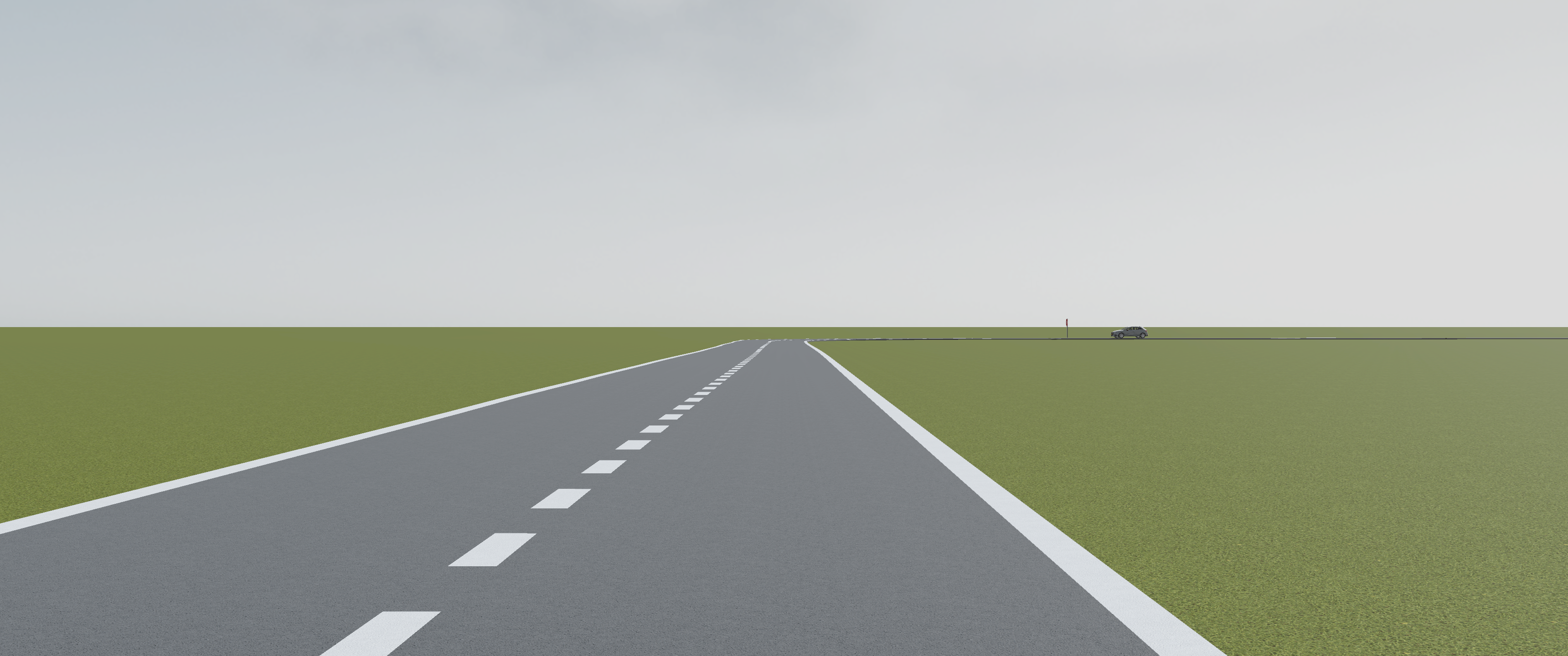}
        \caption{Starting position.}
        \label{fig:start}
    \end{subfigure}%
    \begin{subfigure}[b]{0.5\textwidth}
        \centering
        \includegraphics[width=.9\textwidth]{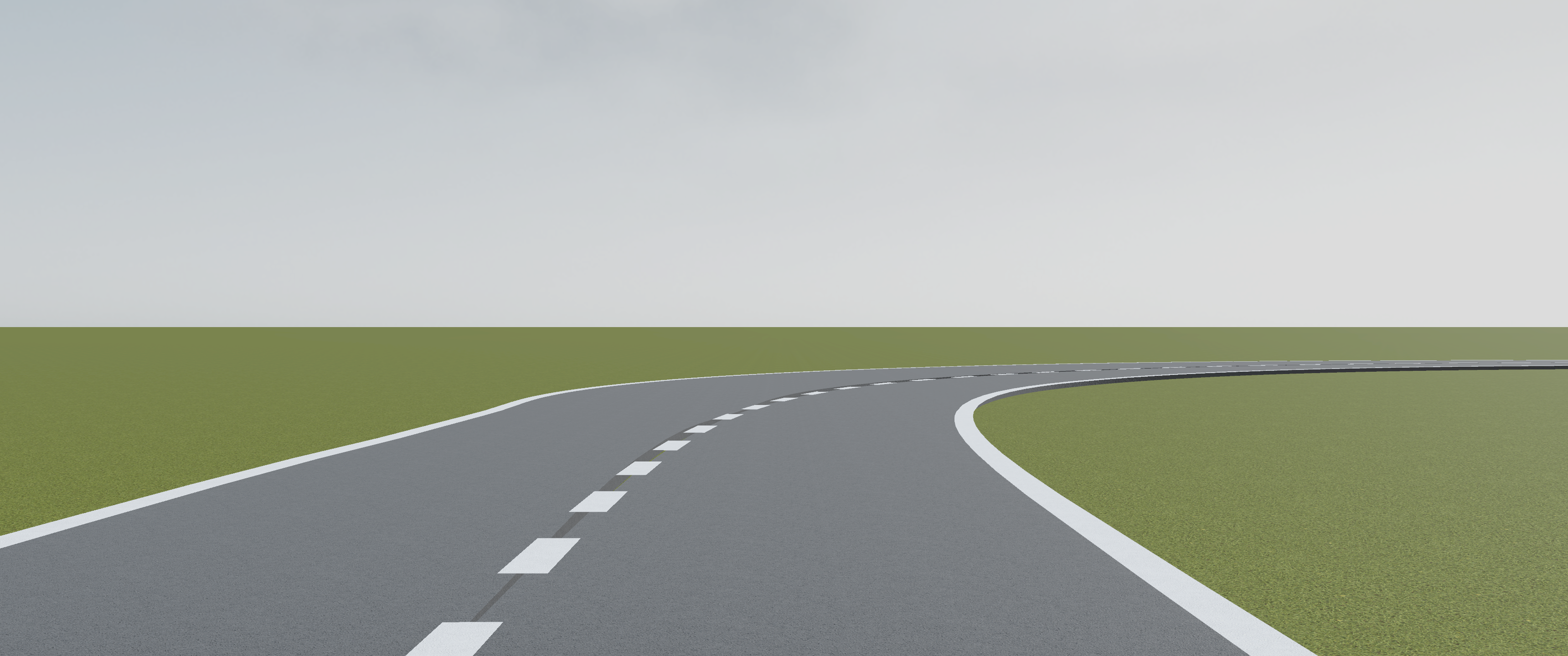}
        \caption{Beginning of the right-hand corner.}
        \label{fig:sub2}
    \end{subfigure}
    \begin{subfigure}[b]{0.5\textwidth}
        \centering
        \includegraphics[width=.9\textwidth]
        {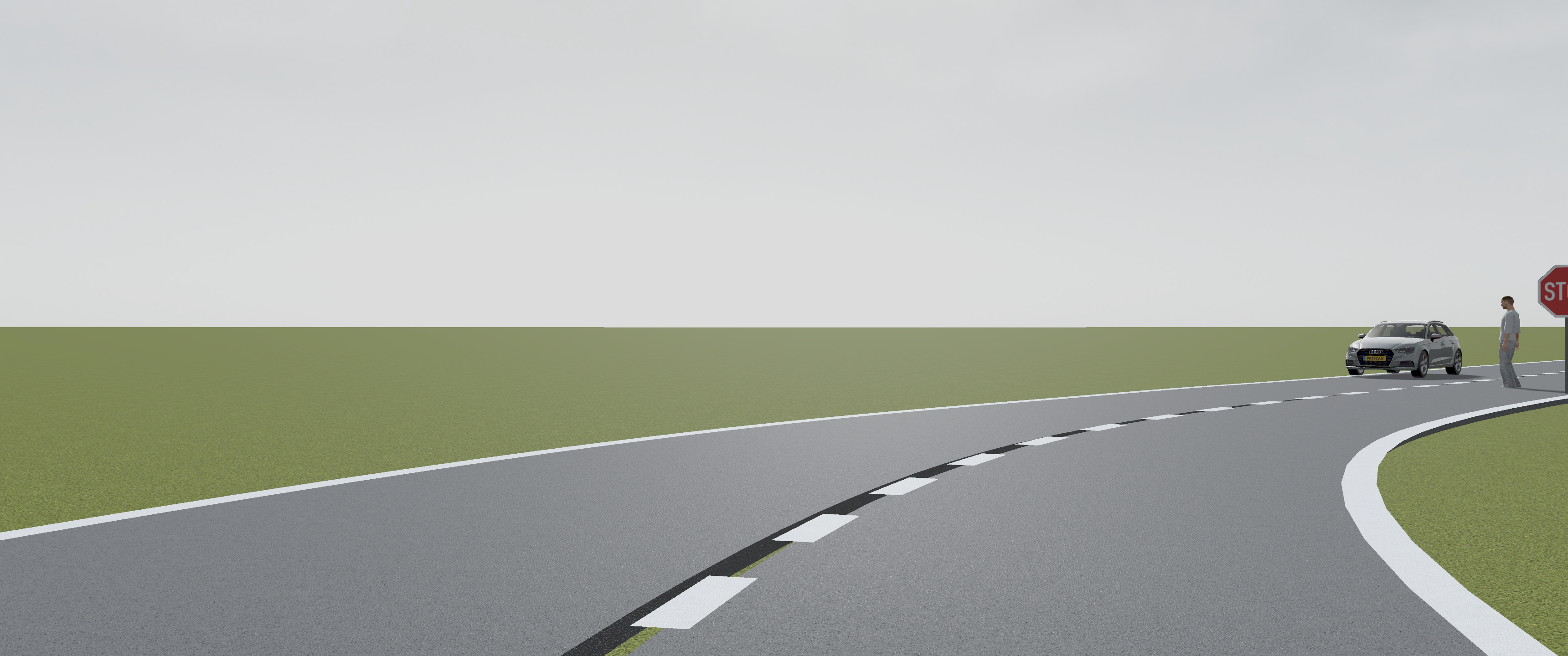}
        \caption{The pedestrian and the stop sign enter the visual frame of the participant (the vehicle is still cornering).}
        \label{fig:pedjoins}
    \end{subfigure}%
    \begin{subfigure}[b]{0.5\textwidth}
        \centering        
        \includegraphics[width=.9\textwidth]{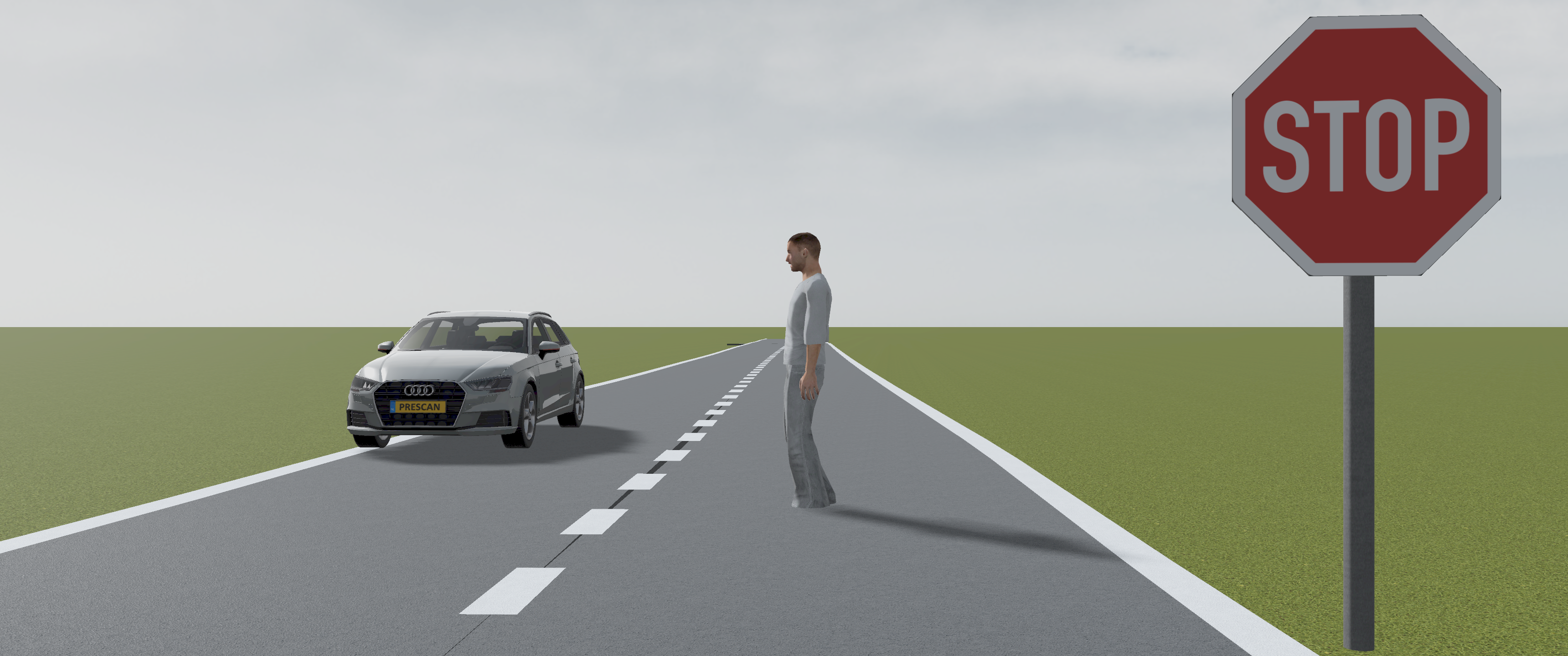}
        \caption{After stopping with the pedestrian crossing.\\ \vspace{\baselineskip}}
        \label{fig:pedestrian}
    \end{subfigure}
    \caption{Screen captures from the participants perspective.}
    \label{fig:scene}
\end{figure*}

The scenarios are based on a real-vehicle experiment conducted at the Griesheim airfield where the same driver consistently drove the vehicle interacting with a real pedestrian \cite{Scharringa2025}. 

\begin{figure}
    \centering
    \includegraphics[width=0.5\textwidth]{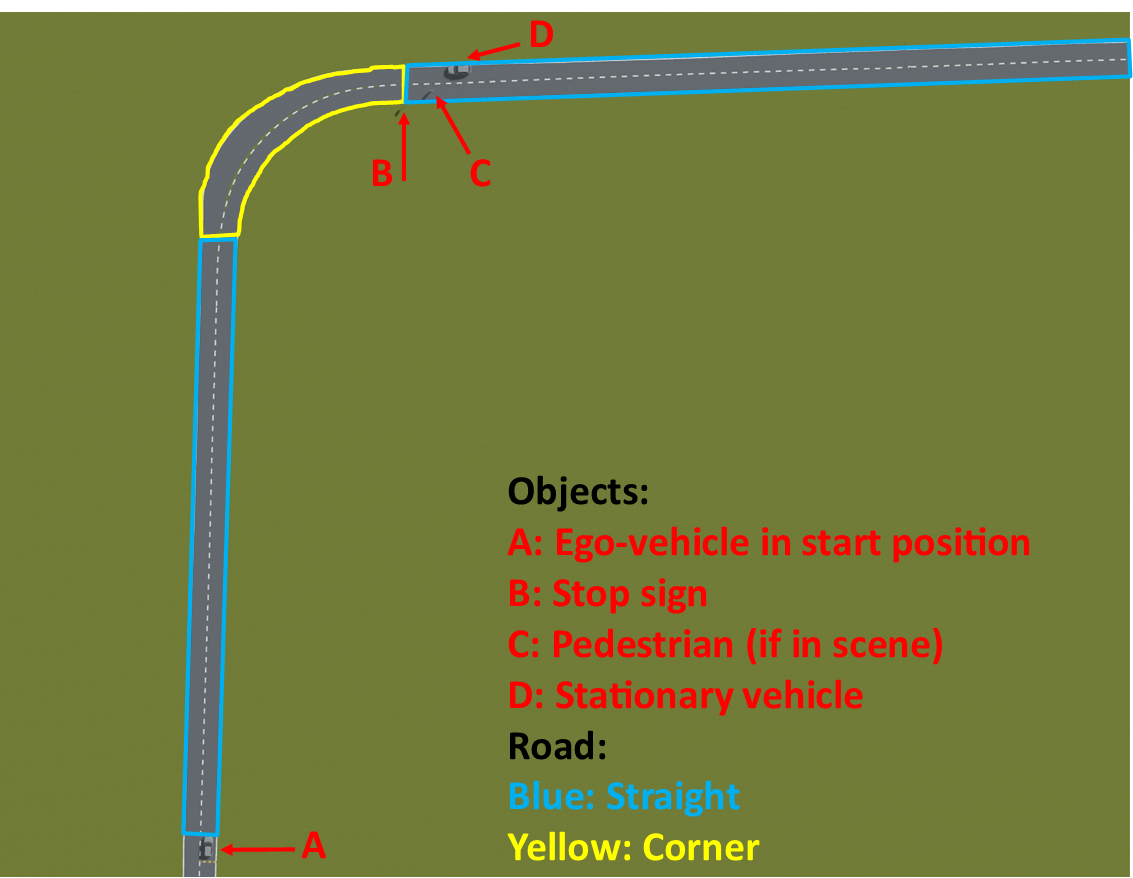}
    \caption{Birds-eye view of the scene, with the different road types, straight line, and right-hand turn, indicated with blue and yellow boxes respectively. The objects in the scene are indicated with red arrows and letters: A is the ego vehicle in the starting position, B is the stop sign, C is the pedestrian before crossing, and D is the stationary vehicle. Note that the pedestrian is not in the scene for every condition.}
    \label{fig:birdseye}
\end{figure}

\begin{figure*}
    \centering
    \includegraphics[width=0.9\textwidth]{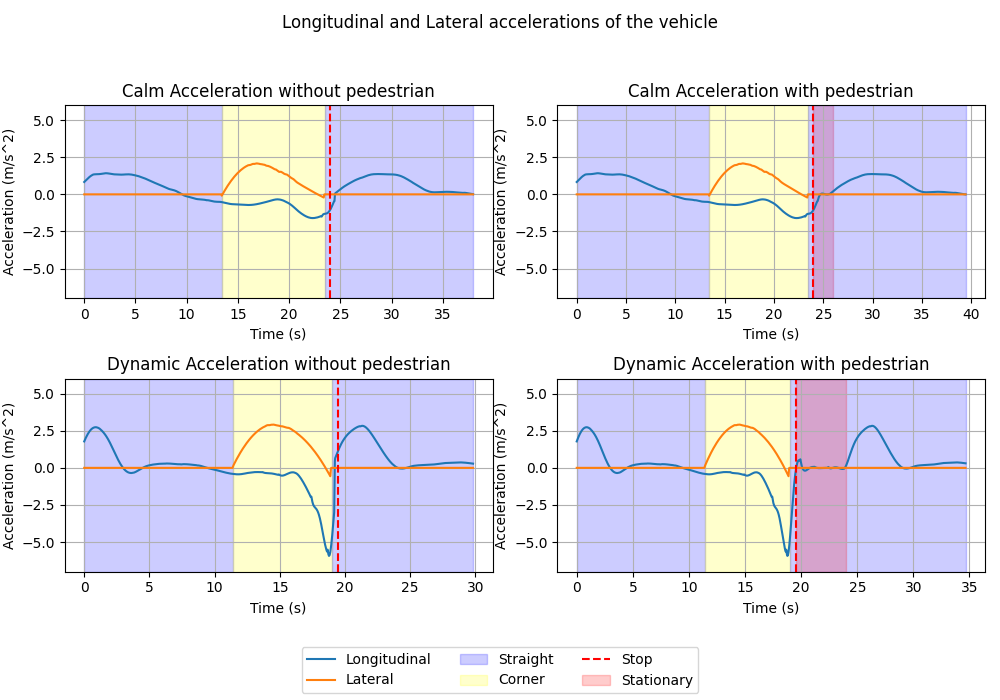}
    \caption{Linear Acceleration profiles of the four different conditions. The colors are the same as on the birds-eye view in Figure \ref{fig:birdseye}. Blue is the straight line, yellow is the right-hand turn, and the red striped line marks the point where the vehicle stops. If the pedestrian is in the scene, the red part is the stationary time for the pedestrian to cross.}
    \label{fig:accelerations}
\end{figure*}

\begin{figure*}
    \centering
    \includegraphics[width=\textwidth]{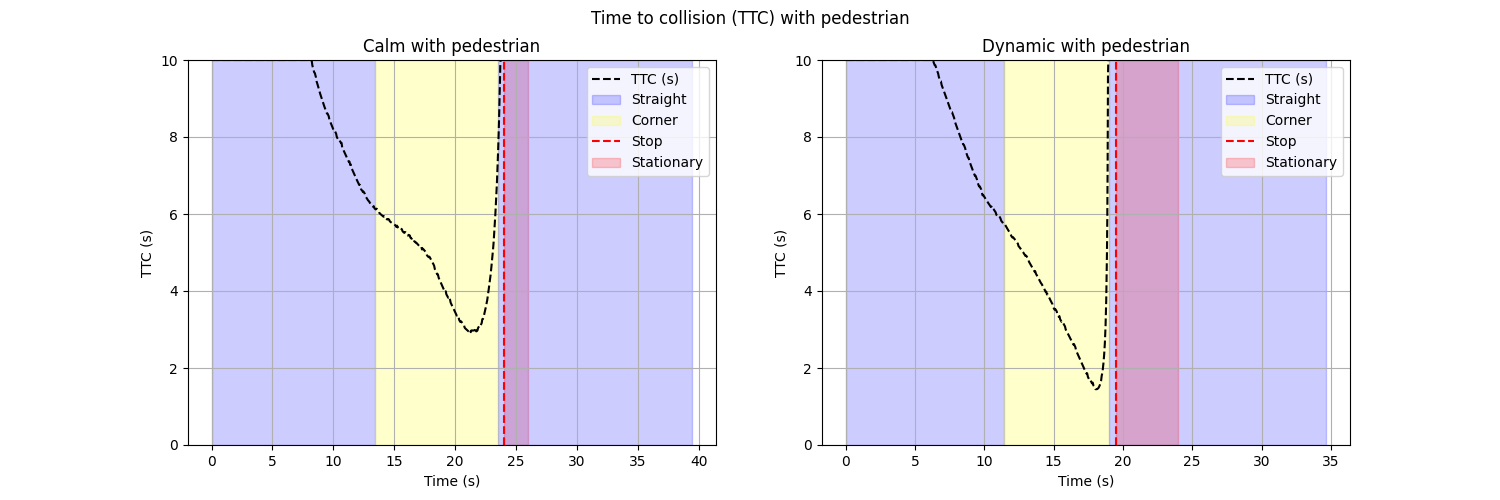}
    \caption{Time-to-collision with pedestrian for both driving styles. Both axes are in seconds, with the x-axis representing the time point in the run, and the y-axis the TTC. The colors are the same as on the birds-eye view in Figure \ref{fig:birdseye}. Blue is the straight line, yellow is the right-hand turn, and the red striped line marks the point where the vehicle stops. If the pedestrian is in the scene, the red part is the stationary time for the pedestrian to cross.}
    \label{fig:ttc}
\end{figure*}

The experiment scenario starts with a straight line, followed by a wide right-hand corner, right after which a stop sign is placed where a stationary vehicle is facing the other direction. After that, the drive continues along a straight line. A birds-eye view of the experiment with pointers and indicators for the different objects and roads can be found in Figure \ref{fig:birdseye}. In the experiment, the participants experienced this scenario in 4 different conditions designed to vary the level of perceived risk. We tested two driving styles, one being more dynamic and one being more calm. Both these styles were experienced with and without a pedestrian crossing at the stop sign. The view of the pedestrian crossing the road from the perspective of the participant can be seen in Figures \ref{fig:pedjoins} and  \ref{fig:pedestrian}. In terms of dynamics, the only difference between conditions with and without a pedestrian is the stationary period. Other than that the dynamics are the same. This can be seen in the acceleration profiles in Figure \ref{fig:accelerations}. In Figure \ref{fig:ttc} the time-to-collision (TTC) is plotted for both driving styles. Here it is visible that the dynamic driving style is more critical, with the TTC decreasing faster and reaching a lower value before the vehicle stops.

The simulator scenarios are created in Simcenter Prescan. These were generated using an in-house real2sim pipeline developed by Siemens. Real2sim refers to the process of translating real-world driving data—captured by a fully instrumented vehicle equipped with lidar, radar, and camera sensors—into high-fidelity simulation environments. As each object in the Simcenter Prescan environment is individually configurable, the reconstructed scenes could be selectively modified; for example, the pedestrian was removed for the no-pedestrian condition. This approach enables controlled variations while maintaining a close link to the original real-world conditions.

The simulator motion platform was controlled using measured vehicle motion from the track test, including linear accelerations and angular velocities from the IMU installed in the vehicle, forward velocity from the GPS and positions in the GNSS  coordinate system that was also used to build the road network in Simcenter Prescan, and create the visualization for the simulator. 

\subsection{Procedure}
At first, participants were asked to fill out a form asking for consent followed by their data regarding demographics and pre-existing trust in automated vehicles (see appendix A).


The experiment started with two conditions of `without pedestrian' (dynamic and calm), in random order and without informing the order of the participant.
After the two of these conditions, participants were asked to identify each of the driving styles, whether is was calm or dynamic.
All but one participant identified these first two runs correctly. Then they were told that these two driving styles would be repeated, and the order would be randomized. They were not informed about the pedestrian crossing the road. This added effect of surprise was intended to stimulate  physiological reactions. With the pedestrian, when asked to identify the driving style, all participants were correct.

Then, all four different conditions (calm with pedestrian, calm without pedestrian, dynamic with pedestrian, dynamic without pedestrian) were repeated in random order, where participants were asked to identify the driving style after each run. 
All participants identified the driving styles correctly when the pedestrian was present. Only seven participants out of thirty-two, incorrectly identified the driving styles without the pedestrian. 
Before starting these four runs, participants were told that now it was possible that they would impact a pedestrian. This was not true, the presented experiments were identical, but aimed to keep participants engaged rather than assuming an identical and safe performance. However, one participant stated that participants should not have been told about this possibility, as now the participant was "warned" about the dynamic driving style being more aggressive. When this participant was told that it was the same run as they had experienced before, the reaction was surprised as the experience was that it was more aggressive. 

After the last run the participants were assisted in removing the sensors and getting out of the simulator, and asked how they felt about the experience. Afterwards, they were given the tablet again with a Google Form containing the same questions on trust and comfort in automated vehicles as before the experiment.

\subsection{Questionnaire}

Before the experiment demographic data was collected. Participants were asked for gender and age,
if they had ever experienced a driving simulator before, and how familiar they were with automated
driving systems. 
A specific questionnaire was designed to assess pre-existing trust in automation, as well as
post-experiment trust. The questions were based on \cite{weigl2021development} who performed a literature study and a focus group with experts, and iteratively developed a questionnaire on acceptance of automated driving, for SAE level 3 and level 5. We adopted 8 questions from their questionnaire for SAE level 5. Each question was given as a statement, to which the participant would give their answer on a 5-point scale from 1 (fully disagree) to 5 (fully agree). After each question, the option was given to explain the answer. The same questions were presented after the experiment to asses a possible change. Participants were asked to elaborate only in case any response had changed.

\subsection{Facial emotion recognition}
\label{subsec:fermodel}
 The model selected for this paper is a Dual-Direction Attention Mixed Feature Network (DDAFN) by \cite{zhang2023dual}. This model achieves start-of-the-art results on multiple datasets, and the developers released both code and the trained weights on \href{https://github.com/simon20010923/DDAMFN}{GitHub}, with the confusion matrices showing the performance on different datasets. They recently released an improved version under the name DDAMFN++ in the same repository, which is used in this paper. It was validated on the Real-world Affective Faces DataBase (RAF-DB) \cite{li2017reliable, li2019reliable}. 
 This is a dataset with almost 30-thousand images from the internet, annotated with emotion labels by 40 independent persons, making it an extremely valuable dataset to train and validate emotion recognition models. Overall, the performance is good, but the accuracy for Fear and Disgust is relatively low. Disgust is not expected to be relevant for this application, but Fear is. According to the confusion matrix, Fear is relatively often classified as Sad or Surprise. This can be taken into account for the data analyses. They also tested for cross-database performance, and achieved 75.6\% accuracy on the RAF-DB with a model that was trained on AffectNet-7. This is a very strong achievement for cross-database performance \cite{zhang2023dual} where other models like the PAtt-lite model described below struggle. The cross-database performance suggests this is a suitable model for our application, where it has to classify unseen data.

 We also implemented the lightweight patch and attention network (PAtt-lite) by \cite{ngwe2023patt}. They report state-of-the-art performance on multiple datasets and published code on \href{https://github.com/JLREx/PAtt-Lite?tab=readme-ov-file}{GitHub}, including pre-trained weights which were tested preparing this paper. However on our simulator data, it always predicted Neutral, and even using self-recorded webcam footage while performing multiple very expressive facial expressions, all were labeled as Neutral. 

\begin{figure}[H]
    \centering    \includegraphics[width=\linewidth]{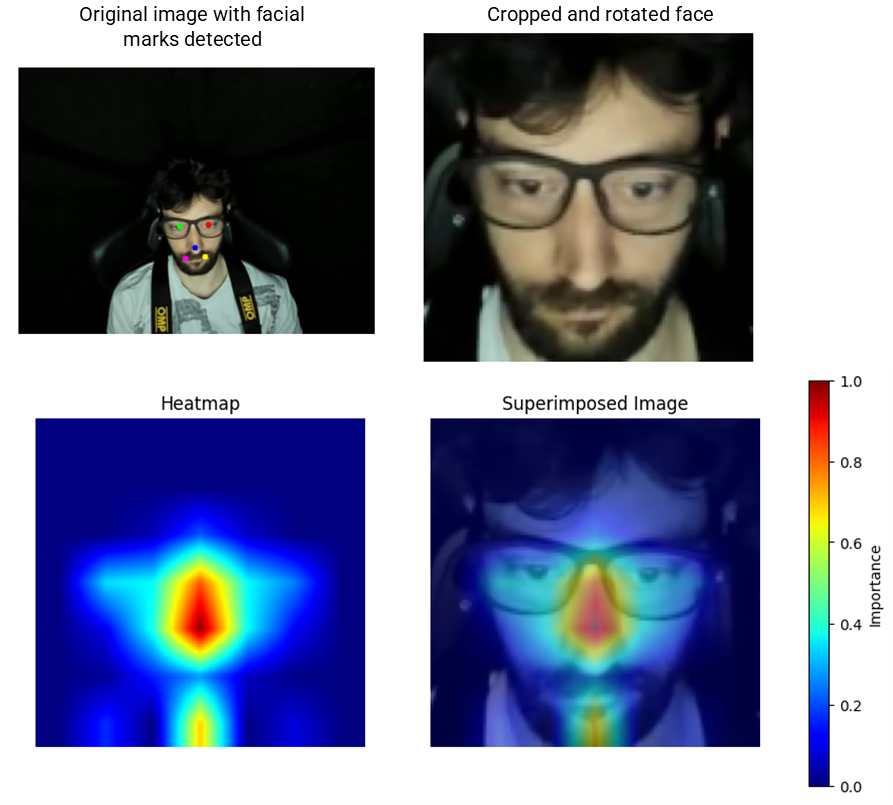}
    \caption{Original image with facial landmarks detected (top left), extracted and rotated face (top right), heatmap showing the focus areas of feature extraction (bottom left) and face and heatmap superimposed (bottom right). }
    \label{fig:groupedheatmap}
\end{figure}

The Dual-Direction Attention Mixed Feature Network (DDAMFN++) by \cite{zhang2023dual} is trained on extracted faces from the dataset, which means using the model also requires to first extracting the faces from the frames. To detect and extract the faces, RetinaFace was used \cite{deng2019retinaface}. Retinaface is the same model as the developers of the DDAMFN++ model used for the pre-processing of their dataset \cite{zhang2023dual}. Retinaface is state-of-the-art for face detection. It detects 5 facial landmarks: eyes, nose, and the corners of the mouth. These landmarks are given in a dictionary with coordinates, in Figure \ref{fig:groupedheatmap}-upper left the detected facial landmarks are plotted, showingthat the model can detect these also for faces with glasses or facial hair.

After this detection, the face has to be cropped out of the frame. The DDAMFN++ model expects an input of size 112 by 112 pixels, which is also enforced by the pre-processing function. The facial area, as detected by RetinaFace, is not always square, and when it is not, it means the extracted face gets deformed a bit by the function. So instead of using the facial area given by RetinaFace, a different approach is used. Pre-defined coordinates are set, of where the facial landmarks should approximately be in the bounding box. Then a spatial transformation matrix is generated that brings the facial landmarks as close to these coordinates as possible, without deforming the image. With this spatial transformation matrix, the bounding box is used to extract the face from the image, which can be seen in Figure \ref{fig:groupedheatmap}-upper right. 

The architecture of the DDAMFN model is explained in more detail by \cite{zhang2023dual}. To visualize the decisions that CNN-based models make, a technique called Gradient-weighted Class Activation Mapping (Grad-CAM) is applied. This is a technique to visualize the regions in an image that are important for the prediction the model makes by generating a heatmap \cite{selvaraju2017grad}. This heatmap is made by taking the gradients of the target class, which in this case is the predicted emotion, with respect to the features of the last convolutional layer. Each feature map is weighted, and by multiplying the feature maps with their weights a map is created highlighting the important areas. Figure \ref{fig:groupedheatmap}-lower right shows the extracted face that is given as input, with the generated heatmap plotted over it, illustrating that the model is well-focused on facial features.

\section{Results}
\label{sec:analyses}

Subjective continuous comfort ratings are presented, confirming that the experiment was successful in eliciting different levels of comfort associated with perceived risk. Then the facial expression recognition results are presented, to answer the main research question. Finally, a comfort prediction model is presented combining vehicle motion and GSR data. 

\subsection{Descriptive analyses}
The experiment was conducted with 32 participants.
Many said they enjoyed the experience, and only two participants mentioned some unease that could be classified as minor motion sickness, but they did not mention sickness and reported no symptoms that would be classified higher than 1 on the MIsery SCale (MISC).

\subsection{Continuous subjective comfort ratings}
All four conditions were experienced twice by the participants, and all individual continuous comfort ratings are plotted in Figures \ref{fig:conscalmnoped}, \ref{fig:conscalmped}, \ref{fig:consdynamicnoped}, and \ref{fig:consdynamicped}.
These figures show a mostly reasonable to good consistency for the two repeated runs within participants. As described in appendix B inconsistent responses and outliers were removed based on dynamic time warping and visual inspection.

\begin{figure}[]
    \centering
    \begin{subfigure}[b]{\linewidth}
        \centering
        \includegraphics[width=0.9\linewidth]{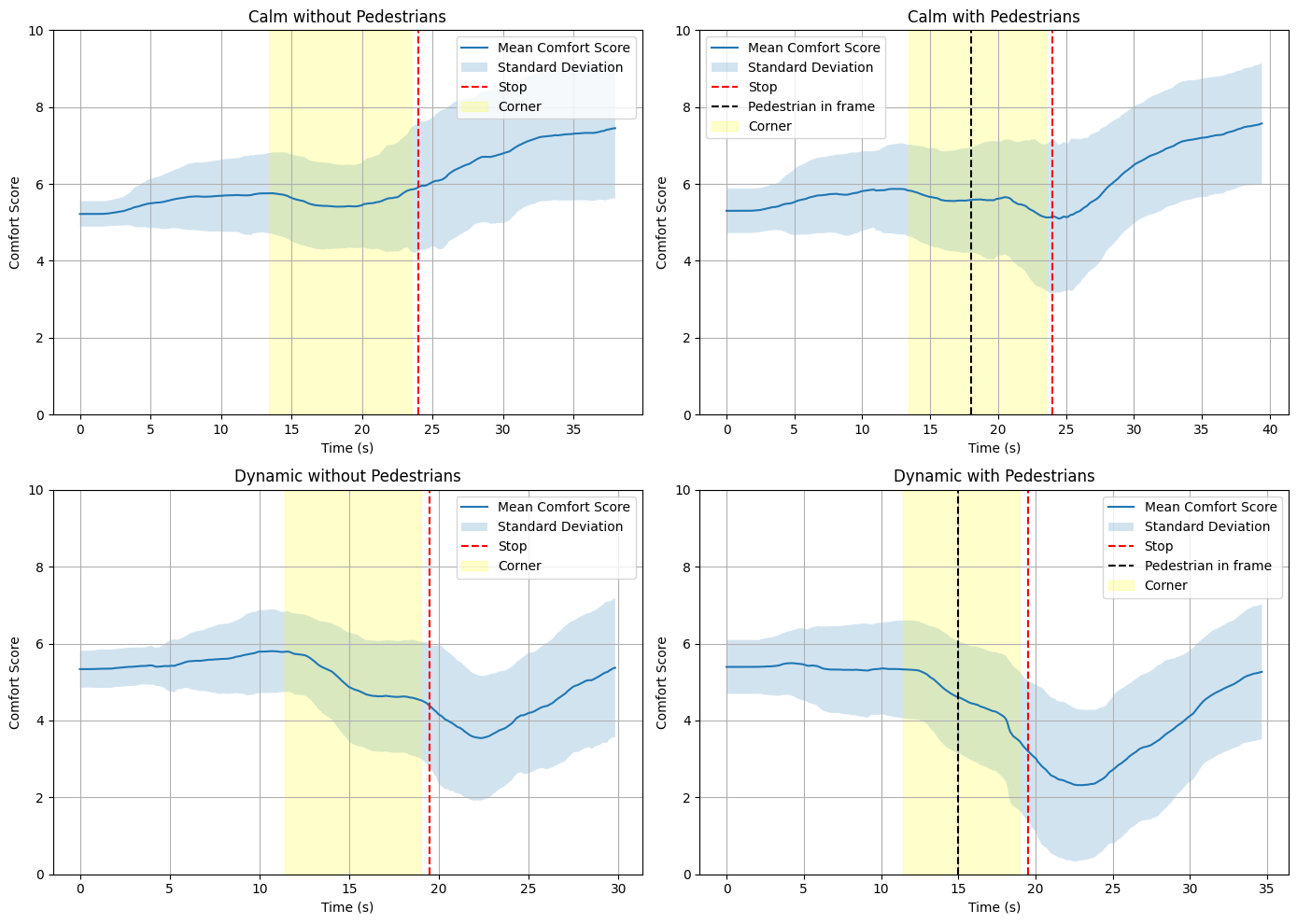}
        \caption{with the standard deviation over participants in light blue.}
        \label{fig:averageall}
    \vspace{\baselineskip}
    \end{subfigure}
    \begin{subfigure}[b]{\linewidth}
        \centering
        \includegraphics[width=0.9\linewidth]{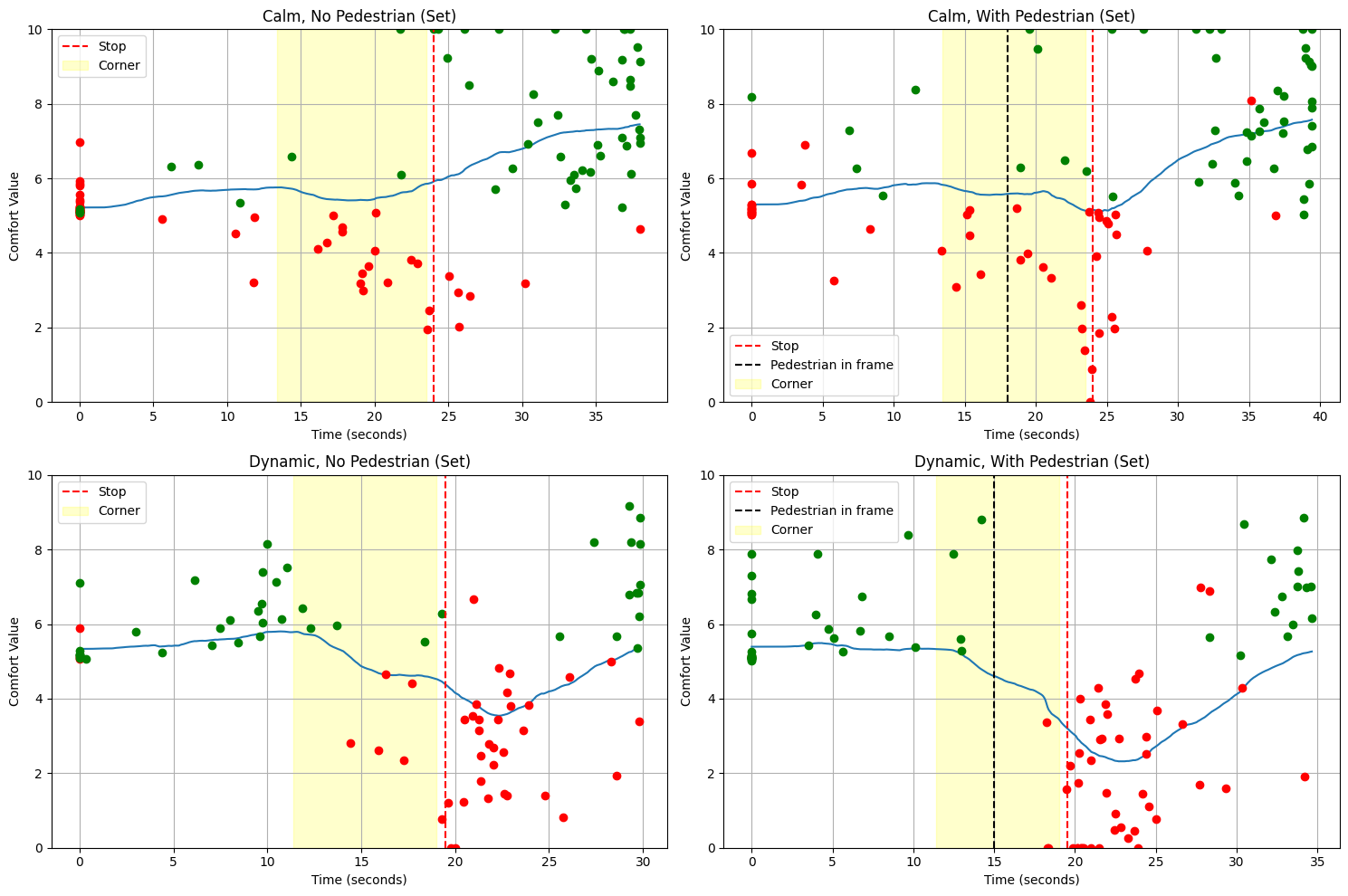}
        \caption{with individual minima (red) and maxima (green) versus the average response (blue).}
        \label{fig:minmax}
    \end{subfigure}
    \caption{Continuous comfort ratings}
\end{figure}

The average and standard deviation over participants after this outlier removal can be found in Figure \ref{fig:averageall}. All conditions show a highly significant divergence from their baseline (p $<$ .001). This derives from t-tests, using the first 10 seconds as a baseline, where the braking event is not yet anticipated. 
Hence, the desired effect of eliciting time-varying comfort levels is achieved. In particular, the calm driving style without pedestrian is perceived as much more comfortable compared to the dynamic style with pedestrian. 

\begin{figure}[H]
    \centering    \includegraphics[width=\linewidth]{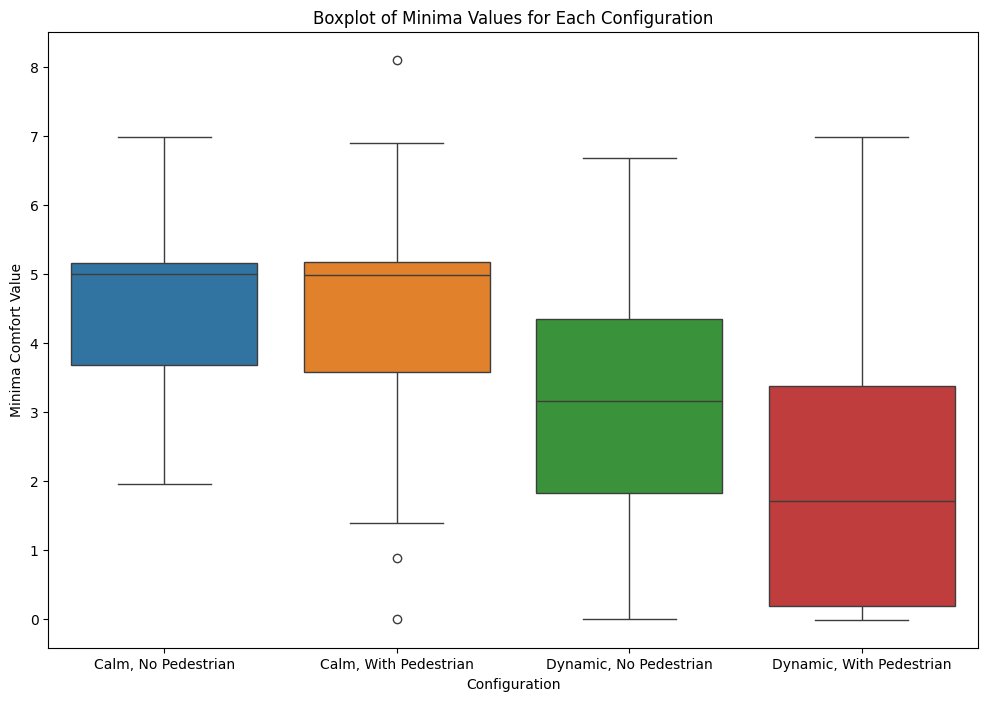}
    \caption{Distribution of absolute minima of the continuous comfort rating.}
    \label{fig:boxplot}
\end{figure}

The above averaging over participants smoothened out some dynamic effects as individual plots were not perfectly aligned in time. Hence we analysed extreme values in Figure \ref{fig:minmax} where each dot represents an extreme value for one run for one participant. The substantial scatter again illustrates a high variety in timing and magnitude of individual responses.
The absolute minima per condition are compared in Figure \ref{fig:boxplot}. 
To quantify the effect of driving styles and pedestrian presence on the minimal perceived comfort, both an ANOVA and a pair-wise t-test test were performed. All effects were highly significant (p<.008) with the exception of Calm no pedestrian, versus Calm with pedestrian.

This illustrates a strong interaction where the pedestrian has no effect in the calm driving style, but leads to the most severe discomfort in the dynamic driving style. 

It was noted by some participants that the car braking timely for the pedestrian with the calm driving style, actually increased their trust. This is also visible in the acquired data, as we see that, even though there is a small dip in the perceived risk, generally comfort increases after the vehicle has successfully stopped. This confirms what was stated by \cite{park2022social}, that trust can be brought down by undesired behavior of the vehicle, but also be increased again if the vehicle shows trustworthy behavior, like braking in time for the pedestrian. 

\subsection{Facial emotion recognition}
\label{subsec:FER}

For facial emotion recognition, we only analyzed the most critical condition, which is the dynamic driving style with the crossing pedestrian. This condition elicits the strongest discomfort during cornering and braking, so if there is any effect on facial expression it will be most evident in this condition. 

The state-of-the-art DDAFN++ model was successfully implemented. Of the 32 participants, 24 were included in the analyses. Exclusions were due to inconsistencies in subjective ratings, as detailed in appendix B, or the inability of the facial recognition software to reliably detect expressions, caused by factors such as participants wearing glasses, looking downward, or skin tone.
For the remaining 24, with minimal lighting conditions, the face could always be detected, and expressions were successfully classified with corresponding emotion labels from the universal set of emotions. However for most of these participants (15/24) the facial expression was always classified as Neutral, and they did not show any detectable reaction in their facial expressions in the two repeated events. 

\begin{figure}[H]
    \centering    \includegraphics[width=\linewidth]{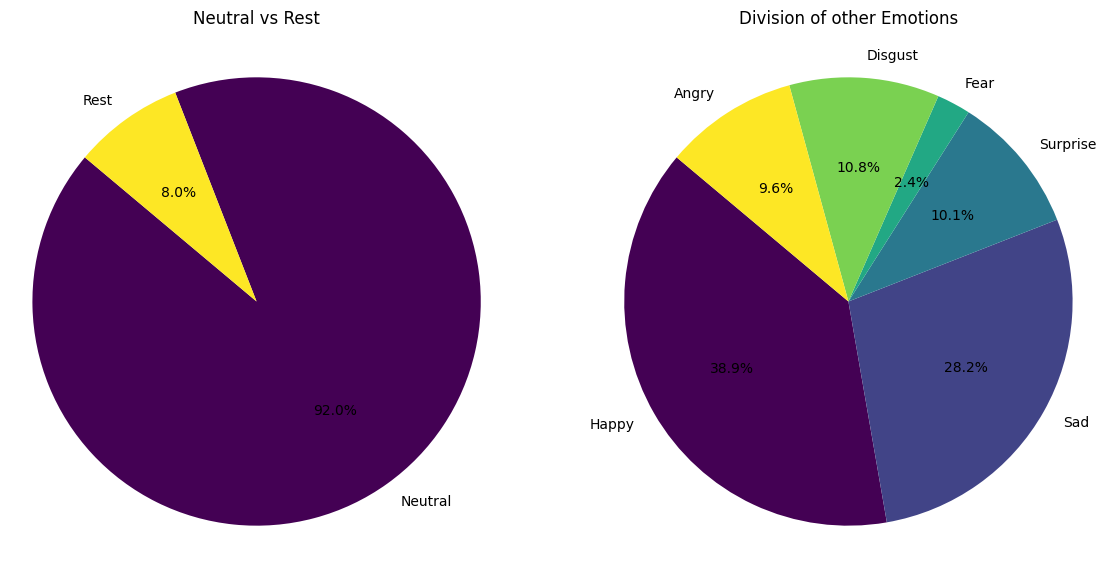}
\caption{Pie charts illustrating the distribution of detected emotions based on accumulated confidence scores. The left chart shows the proportion of Neutral versus all non-neutral expressions combined; the right chart provides a detailed breakdown of the non-neutral categories. Percentages (rounded to two decimals) are: Neutral (92.02\%), Happy (3.10\%), Sad (2.25\%), Disgust (0.86\%), Surprise (0.80\%), Angry (0.77\%), and Fear (0.19\%).}

    \label{fig:piechart}
\end{figure}

\begin{figure}[H]
    \centering    
    \begin{subfigure}[h]{0.48\linewidth}
        \includegraphics[width=0.9\linewidth]{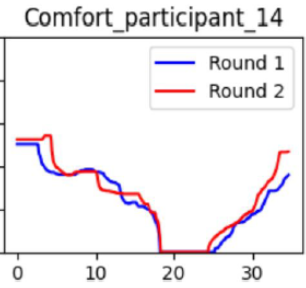}
        \caption{Comfort rating}
        \label{fig:rating14}
    \end{subfigure}
    \begin{subfigure}[h]{0.48\linewidth}
        \includegraphics[width=0.9\linewidth]{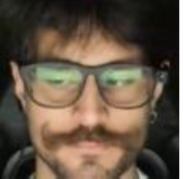}
        \caption{Neutral}
        \label{fig:neutral14}
    \end{subfigure}
    \begin{subfigure}[h]{0.48\linewidth}
        \includegraphics[width=0.9\linewidth]{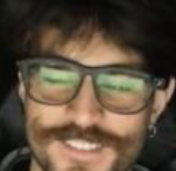}
        \caption{Happy Round 1 t=22s}
        \label{fig:happy14}
    \end{subfigure}
    \caption{Facial emotion recognition participant 14, where the happy expression was absent in round 2.}
    \label{fig:emotion14}
\end{figure}

\begin{figure}[H]
    \centering    
    \begin{subfigure}[h]{0.48\linewidth}
        \includegraphics[width=0.9\linewidth]{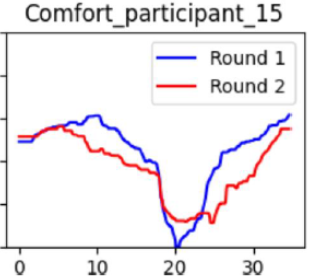}
        \caption{Comfort rating}
        \label{fig:rating15}
    \end{subfigure}%
    \begin{subfigure}[h]{0.48\linewidth}
        \includegraphics[width=0.9\linewidth]{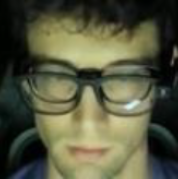}
        \caption{Neutral}
        \label{fig:neutral15}
    \end{subfigure}
    \begin{subfigure}[h]{0.48\linewidth}
        \includegraphics[width=0.9\linewidth]{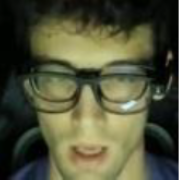}
        \caption{Surprise Round 1 t=33s}
        \label{fig:surpriseR1_P15}
    \end{subfigure}%
   \begin{subfigure}[h]{0.48\linewidth}
        \includegraphics[width=0.9\linewidth]{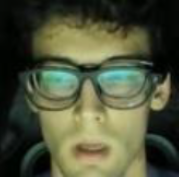}
        \caption{Surprise Round 2 t=6s}
        \label{fig:surpriseR2P15}
    \end{subfigure}
    \caption{Facial emotion recognition participant 15, which was the only case where the same emotion was recognised in Round 2.}
    \label{fig:emotion15}
\end{figure}

Example results are shown in Figures \ref{fig:emotion14} and \ref{fig:emotion15}. The emotion detection model gives a confidence score to each emotion, for each frame in the video. Figure \ref{fig:piechart} shows the total percentage of each emotion, after accumulating all confidence scores. As expected, Neutral is the dominant emotion. However, it was also expected to see Fear and Surprise during and after the critical event. After Neutral, the most dominant emotions are Happy and Sad. Disgust, Surprise, and Anger contribute very little, where Fear is the least detected emotion of all. We know from previous performance that the  DDAFN++ model can sometimes perceive Fear as Sad, but even when we take this into account the contribution of the facial expression for Fear seems to be minimal. 

Representative emotions as a function of time are shown in Figure \ref{fig:allemotions1417} and all plots can be found in  appendix E.
The emotions are plotted together with the time-to-collision, as measure of criticality. 
The moment during the right-hand turn where the pedestrian enters the visual frame, is marked by a red dot (see Figure \ref{fig:pedjoins}).  

Figure \ref{fig:allemotions1417} shows major differences between the first and the second exposure to the same scenario in four participants and  appendix E shows similar differences in other participants. Of the participants showing any reaction, only one participant repeated this reaction with the same intensity during the second round (see \ref{fig:emotion15}). Of the others, half showed a similar reaction but less intense, and half did not show a similar reaction at all. This suggests that the element of not knowing what will happen is important. This is important for future research, as it is common to perform repetitions in these type of experiments, which may enhance statistical power, but hamper realism.
This also highlights a disparity with the continuous comfort rating which was quite consistent for the two repetitions of each condition. 

To further validate the detected facial expressions, a qualitative analysis was conducted for each participant by cross-referencing the continuous comfort ratings with the recorded video footage. The analysis confirmed that the DDAFN++-based emotion detection accurately reflected the visible expressions and did not overlook any apparent facial reactions during the events (see Figures \ref{fig:emotion14} and \ref{fig:emotion15}).



\begin{figure}[H]
    \centering    \includegraphics[width=\linewidth]{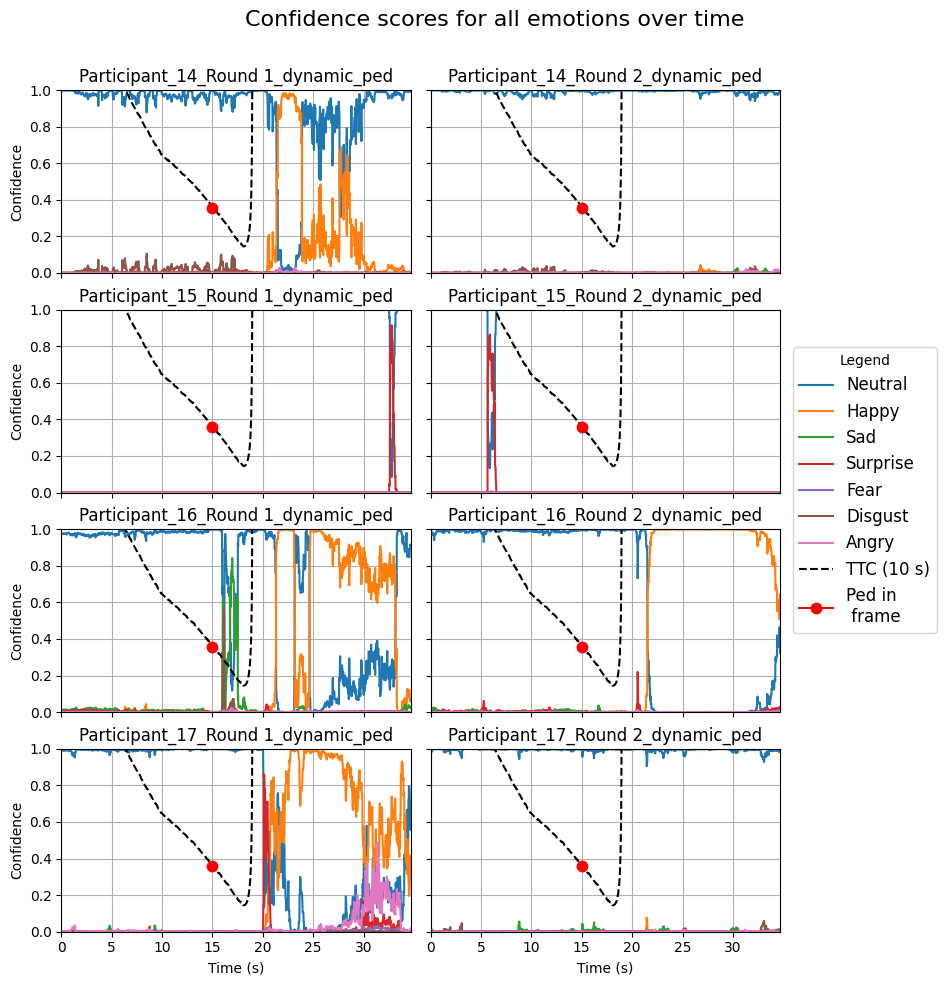}
    \caption{Facial emotion recognition confidence scores over time for Participants 14, 15, 16, and 17 during Rounds 1 and 2. The plots show emotion confidence scores during the dynamic condition involving a pedestrian in the scene, after applying a moving average filter. The red dot marks the moment when the pedestrian and stop sign enter the participant’s visual frame (see Fig.~\ref{fig:pedjoins}). Time-to-collision (TTC) is scaled by a factor of 0.1 for visualization purposes.}
    \label{fig:allemotions1417}
\end{figure}

\subsection{Neural network for comfort prediction}
\ref{subsec:GSR_driving_style} shows that GSR can discriminate the two driving styles, which indicates that GSR is a physiological measurement that can be used for comfort prediction in the tested conditions.
Experiments and models have shown that vehicle motion can predict perceived risk as an important component of comfort in automated driving \cite{he2022modelling,HE2024PCAD}. Within such models, the time to collision (TTC), defined as distance divided by approach speed, is a well-known surrogate safety measure associated with both actual and perceived risk (see \cite{Mulakkal_Safety_Indicators2017,he2022modelling}). This section presents a model predicting the measured continuous comfort using vehicle motion, TTC and/or GSR as inputs.

As elaborated in appendix D, GSR is split into Tonic and Phasic components and both are used as possible features for the model to train on, together with the time-to-collision (TTC) and longitudinal and lateral vehicle accelerations.


GSR, TTC and accelerations are passed through their own feature extraction models consisting of three Convolutional layers and two LSTM layers. The outputs of these models are combined in the attention layer. Finally, a fully connected layer is used to obtain a prediction for the continuous comfort rating. The mean absolute error of the prediction versus the true value is used as a loss function to train the model. 
For an input of 5 seconds of data, which had a frequency of 10Hz, the model outputs one score to predict the perceived risk. As a true label, the score given by the participants at the end of these 5 seconds was taken. 
The participant perceived risk scores were available in 2 decimal precision but rounded to 1 decimal, as turning the knob by a smaller amount on purpose was practically impossible. 

In pre-processing, the data was resampled to 10Hz and then the sequences of 5 seconds were created. The data was then standardized using a standard scalar, which gives the data zero mean and a standard deviation of 1. For the GSR data, this was not done on all data combined, because this would make the GSR signal for participants with a low mean almost zero. For this reason, the GSR signals were standardized individually before splitting into tonic and phasic components. 20\% of the data was set aside as a validation set, and 80\% was used as training data. In this split, it was ensured that both sets contained roughly the same distribution for the labels. This can be seen in Figure \ref{fig:distribution}. We also see that the dataset is not balanced, with much less data near the extremes. To account for this, different strides were applied when sampling the data, resulting in a division shown in Figure \ref{fig:distributionequal}.



The performance of the model was compared for the different features, looking at the loss (mean absolute error) and accuracy on both the fitted training set and the validation set. For the accuracy, a margin of error was given, since it can be considered correct if the model predicts a value close to the label. 
This margin of error was chosen to be plus or minus 1 around label 5, and linearly building up to 2 on both outer labels, 0 and 10. This variable margin was selected because at the edges the exact number becomes less important, a perceived risk rating below 2 means extremely low perceived risk and above 8 extremely high perceived risk. 


The different feature combinations were compared in Table \ref{tab:featureperformance}. We see that in terms of accuracy, the best performance is the model that is trained on both vehicle motion and GSR. However, only vehicle motion does reach a slightly lower mean absolute error, although this difference is very small with the model that combines the two. In terms of fitting the model to the training data, using only vehicle motion seems to work the least well, and only GSR provides a better fit than the combined model. Here it should be taken into account, that the accelerations were the same for every run, meaning that also between the training and testing data, there will be a lot of highly similar samples, if not exactly the same. This is generally not desired, as it will make the model less applicable to data outside of this experiment, thus it is not possible to really test its validity. Another thing that can be noted is that because the labels were from different participants, giving different ratings, all while experiencing the same accelerations, the model gets exactly the same input, but with different output labels to converge to. This limits the extent to which the model can fit to the data. This was confirmed when trying to run the model for 1000 epochs, which improved the fitting on the vehicle motion data only slightly both for the MAE (-0,023) and the accuracy (+1.4\%). When doing the same with the model only trained on GSR data (which varies between participants), a much bigger improvement was found on both the MAE (-0,248) and accuracy (+7.1\%). This shows that to predict individual perceived risk levels, it helps to incorporate physiological data, and GSR is a viable predictor. 
TTC is another feature that always shows the same pattern in time within a driving style, because TTC is taken from environmental variables and so does not change between participants. In this NN a time varying TTC was only defined when the pedestrian was visible (in all other cases it was set at its maximum of 10 s). It can be seen that TTC has no positive effect on the performance, it just results in more overfitting with an improved fit of the training set but a detoriated fit of the test set. TTC was thus left out in further processing. 
GSR could be used to discriminate between the two driving styles, which was shown to be the biggest factor between low perceived risk and high perceived risk drives, and it shows promising results for predicting individual perceived risk levels.

\begin{table}[H]
\centering
\caption{Performance for different feature sets. Accelerations are longitudinal and lateral. GSR consists of the tonic and the phasic component. TTC is time to collision.}
\label{tab:featureperformance}
\begin{tabular}{@{}lcccc@{}}
\toprule
Features &
  \begin{tabular}[c]{@{}c@{}}MAE\\ train\end{tabular} &
  \begin{tabular}[c]{@{}c@{}}MAE\\ test\end{tabular} &
  \begin{tabular}[c]{@{}c@{}}Accuracy\\ train\end{tabular} &
  \begin{tabular}[c]{@{}c@{}}Accuracy\\ test\end{tabular} \\ \midrule
Accelerations             & 1,306 & 1,330 & 64,8\% & 62,9\% \\
GSR                       & 1,234 & 1,660 & 70,6\% & 55,5\% \\
Accelerations + GSR       & 1,321 & 1,352 & 65,4\% & 65,1\% \\
Accelerations + GSR + TTC & 1,029 & 1,426 & 76,2\% & 60,9\% \\ \bottomrule
\end{tabular}%
\end{table}

To demonstrate the performance on individual data, Leave-One-Out-Cross-Validation (LOOCV) was applied. The model was trained omitting data from one participant, and then correlated to the omitted data.
For the GSR only model, all correlations were positive, and 11/14 were significant (p<0.01). For only vehicle motion, all correlations were significant and much stronger than for the GSR. For the combined feature model, all 14 correlations were positive and significant, showing highly similar performance to the model with only vehicle motion. This shows that these models capture data of unseen individual participants.

\section{Discussion}
\label{sec:discussion}
Developments in automated driving promise to enhance the quality of life with fewer traffic accidents, and more time that can be spent on other activities. These benefits rely on acceptance by the public, and thereby require high levels of perceived safety and trust in automated driving. This is why understanding perceived risk in automated vehicles is critical for further development in this area. This paper contributes to the development and validation of objective methods for assessing emotional states of users of vehicle automation.

\subsection{Data collection}
\label{subsec:data_collection}
A comprehensive dataset was collected from a vehicle simulator experiment that elicited different levels of comfort in relation to perceived risk. This dataset includes continuous subjective comfort ratings, vehicle motion from the real-world drive on which the simulation was based, webcam footage monitoring the person's facial expression, and Galvanic Skin Response.  Heart rate (variability), and eye-tracking were collected but not yet analyzed. 

The continuous subjective comfort ratings were analyzed and their relation to calm and dynamic driving styles and the presence of a crossing pedestrian were quantified. Results demonstrate that the experiment successfully elicited different levels of comfort evidenced in the continuous comfort rating and GSR but not through facial emotion recognition. In the braking event without the pedestrian crossing, the dynamic driving style elicited a stronger discomfort as compared to the calm driving style. Adding the crossing pedestrian did not significantly affect discomfort with the calm driving style but leads to the most severe discomfort with the dynamic driving style. With or without pedestrian the accelerations are identical and the stop sign, which is present in both cases indicates the desired stopping location. Hence the perception of relative motion is equivalent. The main difference lies in the perceived consequence of inadequate braking which is fundamentally more severe with a pedestrian crossing the road. This aligns with ample studies relating perceived risk both to the probability and the severity of consequences of events. Complex interactions between effects of driving style and pedestrian crossing are also reported  by \cite{peintner2024driving}, for a simulator experiment with an aggressive and a defensive driving style with pedestrians crossing while asking about the participants' desire for control, trust in automation, and acceptance. They found that the preferred driving style depends strongly on the crossing event type where participants did not always prefer the defensive style. A significant difference between their experiment and ours is that they did not have a moving base simulator, whereby the dynamics of the driving styles were not physically experienced by the participants. 

\subsection{Facial emotion recognition for perceived risk detection}
\label{subsec:facial_emotion_recog}
For facial emotion recognition (FER), the state-of-the-art Dual-Direction Attention Mixed Feature Network (DDAFN++) model was successfully implemented \cite{zhang2023dual}. Under minimal lighting conditions, with the simulator monitor as the only light source, faces could consistently be detected and emotions classified for 24 participants. This demonstrates the feasibility of applying FER in a driving simulator setting mimicking automated driving with road-monitoring users.

FER assigned labels frame-by-frame using the widely accepted basic emotions: Neutral, Happy, Sad, Angry, Disgust, Surprise, and Fear. However, most participants (15 out of 24) did not exhibit detectable changes in facial expressions during the most critical events, even though their subjective comfort ratings and GSR responses showed clear signs of discomfort. This lack of facial reaction suggests that FER, in its current form, does not reliably capture (dis)comfort related to perceived risk.

Contrary to expectations, Fear was the least detected emotion (see Figure \ref{fig:piechart}). Among the nine participants showing non-neutral emotions, eight displayed a dominant Happy expression, typically after the pedestrian crossing. This may reflect a sense of relief following the stressful event (decreasing TTC), although no signs of Fear were detected.

The absence of Fear and the occurrence of Happiness may be partly explained by participants’ awareness of the simulation environment, reducing their perception of real danger. Additionally, technical factors may have influenced FER performance: the webcam was positioned above the monitor, often above participants’ eye level, while some participants wore large eye-tracking glasses or tilted their heads forward, obscuring key facial features. Reflections from the monitor further complicated expression detection, especially around the eyes—critical areas for emotion recognition \cite{zhang2023dual}.

Overall, these results suggest that FER is currently not a reliable method for assessing comfort and perceived risk in automated driving. Similar findings from other studies \cite{wintersberger2016automated,sini2020automatic} indicate that this limitation may be general, rather than specific to the present experiment.

\subsection{GSR as a proxy for comfort}
\label{subsec:gsr_proxy_comfort}
A neural network was implemented to predict continuous comfort from vehicle motion and GSR. It was shown that both GSR and vehicle motion can successfully fit the dataset, where vehicle motion shows better performance on the test set, but also limitations on how well it can fit the training set. More importantly, the LOOCV analyses showed the model was able to follow trends of subjective perceived risk well for data from unseen participants, where for all participants, the self-asserted perceived risk and the predicted perceived risk show a positive correlation. This was the case both with and without including the GSR features. This might be caused by the fact that the vehicle motions are highly correlated with the subjective perceived risk in this experiment. Since the dynamics are always the same, the model learns this relation and the varying GSR signal becomes of little influence. This model was taken from literature and not fine-tuned, meaning it is likely possible to get better results from this dataset in future research. These findings do already show the potential of achieving objective perceived risk assessment in automated vehicles, losing the biases that are inherent to subjective assessment, and paving the way for future studies in this area. 

\subsection{Future work}
\label{subsec:future}

State-of-the-art models can classify emotions from facial expressions, but most research is based on laboratory settings or internet images, where display rules—learned behaviors influenced by culture, sex, and age—may exaggerate expressions \cite{li2020deep, ekman1969repertoire}. In our experiment, many participants showed no detectable facial reaction despite strong subjective discomfort, highlighting the greater reliability of physiological signals like GSR. Future research could explore facial expressions in real-world or more unpredictable simulated environments, as repeated exposure to the same scenario may have reduced emotional responses in our study. Nevertheless, the consistency of continuous comfort ratings and GSR supports the validity of our design.

This study used discrete facial emotion classification due to the availability of rich datasets and pre-trained models. However, alternative approaches, such as analyzing Facial Action Units (AUs) \cite{beggiato2020facial}, could provide more sensitive measures, as they capture subtle facial movements linked to emotions like surprise. Mapping expressions to arousal and valence dimensions, already explored in other modalities like eye tracking \cite{mou2023driver}, also offers promise but would require specialized datasets for automated driving contexts.

Combining GSR and facial expressions, as suggested by \cite{meza2021driver}, may help distinguish positive and negative arousal. However, as our study observed, participants sometimes smiled during discomfort, complicating interpretation. Future studies could refine objective methods for perceived risk assessment, minimizing reliance on subjective ratings. Herein haptic feedback devices  as currently being developed at Siemens, instead of visual comfort knobs, can provide less intrusive, bias-free measurement.

\section{Conclusion} \label{sec:conclusion}

This study investigated the potential of facial emotion recognition (FER) and Galvanic Skin Response (GSR) as objective measures for perceived risk and comfort in automated driving. A comprehensive simulator experiment successfully elicited varying levels of discomfort associated with perceived risk, confirmed by subjective comfort ratings and GSR data. Although a state-of-the-art FER model (DDAFN++) was successfully implemented under minimal lighting conditions, facial expressions largely remained neutral during critical events, and fear-related emotions were rarely detected. In some cases, Happy expressions were observed following the pedestrian crossing, potentially reflecting relief rather than heightened risk perception. These findings suggest that FER, under current conditions, is not a reliable proxy for assessing perceived risk. Technical factors, including camera placement and limited facial visibility due to eyewear and lighting reflections, may have impacted detection performance. In contrast, GSR signals consistently tracked discomfort levels and, when combined with vehicle motion data, enabled promising predictions of perceived risk through a neural network model. Overall, these results highlight the limitations of FER for comfort assessment in automated driving and underscore the value of physiological signals, particularly GSR, as objective and reliable indicators. Future work should further explore physiological and behavioral markers, improve facial expression analysis methods, and refine experimental protocols to better capture emotional responses under more realistic and unpredictable conditions.

\section*{Acknowledgement} 
This research is supported by VLAIO (Flanders Innovation \& Entrepreneurship Agency) within the BECAREFUL (Belgium consortium for enhanced safety \& comfort perception of future autonomous vehicles) project, no. HBC.2021.0939 and by the European Union’s Horizon Europe research and innovation program under grant agreement No. 101076754 – project AIthena.

\bibliographystyle{ieeetr}

\begin{thebibliography}{10}

\bibitem{dawson2007electrodermal}
M.~E. Dawson, A.~M. Schell, D.~L. Filion, {\em et~al.}, ``The electrodermal system,'' {\em Handbook of psychophysiology}, vol.~2, pp.~200--223, 2007.

\bibitem{jaiswal2023gsr}
D.~Jaiswal, D.~Chatterjee, M.~B~s, R.~K. Ramakrishnan, and A.~Pal, ``Gsr based generic stress prediction system,'' in {\em Adjunct Proceedings of the 2023 ACM International Joint Conference on Pervasive and Ubiquitous Computing \& the 2023 ACM International Symposium on Wearable Computing}, pp.~433--438, 2023.

\bibitem{wang2019detection}
K.~Wang, Y.~L. Murphey, Y.~Zhou, X.~Hu, and X.~Zhang, ``Detection of driver stress in real-world driving environment using physiological signals,'' in {\em 2019 IEEE 17th International Conference on Industrial Informatics (INDIN)}, vol.~1, pp.~1807--1814, IEEE, 2019.

\bibitem{memar2021stress}
M.~Memar and A.~Mokaribolhassan, ``Stress level classification using statistical analysis of skin conductance signal while driving,'' {\em SN Applied Sciences}, vol.~3, no.~1, p.~64, 2021.

\bibitem{ekman1999basic}
P.~Ekman {\em et~al.}, ``Basic emotions,'' {\em Handbook of cognition and emotion}, vol.~98, no.~45-60, p.~16, 1999.

\bibitem{ekman1978facial}
P.~Ekman and W.~V. Friesen, ``Facial action coding system,'' {\em Environmental Psychology \& Nonverbal Behavior}, 1978.

\bibitem{davoli2020driver}
L.~Davoli, M.~Martal{\`o}, A.~Cilfone, L.~Belli, G.~Ferrari, R.~Presta, R.~Montanari, M.~Mengoni, L.~Giraldi, E.~G. Amparore, {\em et~al.}, ``On driver behavior recognition for increased safety: a roadmap,'' {\em Safety}, vol.~6, no.~4, p.~55, 2020.

\bibitem{savran2013automatic}
A.~Savran, R.~Gur, and R.~Verma, ``Automatic detection of emotion valence on faces using consumer depth cameras,'' in {\em Proceedings of the IEEE International Conference on Computer Vision Workshops}, pp.~75--82, 2013.

\bibitem{leng2007experimental}
H.~Leng, Y.~Lin, and L.~Zanzi, ``An experimental study on physiological parameters toward driver emotion recognition,'' in {\em Ergonomics and Health Aspects of Work with Computers: International Conference, EHAWC 2007, Held as Part of HCI International 2007, Beijing, China, July 22-27, 2007. Proceedings}, pp.~237--246, Springer, 2007.

\bibitem{russell1980circumplex}
J.~A. Russell, ``A circumplex model of affect.,'' {\em Journal of personality and social psychology}, vol.~39, no.~6, p.~1161, 1980.

\bibitem{cai2011modeling}
H.~Cai and Y.~Lin, ``Modeling of operators' emotion and task performance in a virtual driving environment,'' {\em International Journal of Human-Computer Studies}, vol.~69, no.~9, pp.~571--586, 2011.

\bibitem{ekman1969pan}
P.~Ekman, E.~R. Sorenson, and W.~V. Friesen, ``Pan-cultural elements in facial displays of emotion,'' {\em Science}, vol.~164, no.~3875, pp.~86--88, 1969.

\bibitem{matsumoto1990cultural}
D.~Matsumoto, ``Cultural similarities and differences in display rules,'' {\em Motivation and emotion}, vol.~14, no.~3, pp.~195--214, 1990.

\bibitem{safdar2009variations}
S.~Safdar, W.~Friedlmeier, D.~Matsumoto, S.~H. Yoo, C.~T. Kwantes, H.~Kakai, and E.~Shigemasu, ``Variations of emotional display rules within and across cultures: A comparison between canada, usa, and japan.,'' {\em Canadian Journal of Behavioural Science/Revue canadienne des sciences du comportement}, vol.~41, no.~1, p.~1, 2009.

\bibitem{saffarian2012drivers}
M.~Saffarian, R.~Happee, and J.~d. Winter, ``Why do drivers maintain short headways in fog? a driving-simulator study evaluating feeling of risk and lateral control during automated and manual car following,'' {\em Ergonomics}, vol.~55, no.~9, pp.~971--985, 2012.

\bibitem{su2023development}
H.~Su, J.~Brooks, and Y.~Jia, ``Development and evaluation of comfort assessment approaches for passengers in autonomous vehicles,'' tech. rep., SAE Technical Paper, 2023.

\bibitem{beggiato2020komfopilot}
M.~Beggiato, F.~Hartwich, P.~Ro{\ss}ner, A.~Dettmann, S.~Enhuber, T.~Pech, D.~Gesmann-Nuissl, K.~M{\"o}{\ss}ner, A.~C. Bullinger, and J.~Krems, ``Komfopilot—comfortable automated driving,'' {\em Smart automotive mobility: reliable technology for the mobile human}, pp.~71--154, 2020.

\bibitem{he2022modelling}
X.~He, J.~Stapel, M.~Wang, and R.~Happee, ``Modelling perceived risk and trust in driving automation reacting to merging and braking vehicles,'' {\em Transportation research part F: traffic psychology and behaviour}, vol.~86, pp.~178--195, 2022.

\bibitem{perez2025perceived}
E.~P{\'e}rez-Moreno, J.~E. Naranjo, M.~J. Hern{\'a}ndez, T.~Ru{\'\i}z, A.~Valle, A.~Cruz, F.~Serradilla, and F.~Jim{\'e}nez, ``Perceived risk and acceptance of automated vehicles users to unexpected hazard situations in real driving conditions,'' {\em Behaviour \& Information Technology}, pp.~1--18, 2025.

\bibitem{park2022social}
C.~Park and M.~Nojoumian, ``Social acceptability of autonomous vehicles: unveiling correlation of passenger trust and emotional response,'' in {\em International Conference on Human-Computer Interaction}, pp.~402--415, Springer, 2022.

\bibitem{wintersberger2016automated}
P.~Wintersberger, A.~Riener, and A.-K. Frison, ``Automated driving system, male, or female driver: Who'd you prefer? comparative analysis of passengers' mental conditions, emotional states \& qualitative feedback,'' in {\em Proceedings of the 8th international conference on automotive user interfaces and interactive vehicular applications}, pp.~51--58, 2016.

\bibitem{sini2020automatic}
J.~Sini, A.~C. Marceddu, and M.~Violante, ``Automatic emotion recognition for the calibration of autonomous driving functions,'' {\em Electronics}, vol.~9, no.~3, p.~518, 2020.

\bibitem{cieslak2020accurate}
M.~Cieslak, S.~Kanarachos, M.~Blundell, C.~Diels, M.~Burnett, and A.~Baxendale, ``Accurate ride comfort estimation combining accelerometer measurements, anthropometric data and neural networks,'' {\em Neural Computing and Applications}, vol.~32, pp.~8747--8762, 2020.

\bibitem{Scharringa2025}
J.~Scharringa, K.~Gkentsidis, M.~Sarrazin, R.~Happee, and K.~Janssens, ``Perceived comfort and safety in automated driving based on physiological signals: Findings from a proving ground study,'' submitted.

\bibitem{weigl2021development}
K.~Weigl, C.~Schartm{\"u}ller, A.~Riener, and M.~Steinhauser, ``Development of the questionnaire on the acceptance of automated driving (qaad): Data-driven models for level 3 and level 5 automated driving,'' {\em Transportation research part F: traffic psychology and behaviour}, vol.~83, pp.~42--59, 2021.

\bibitem{zhang2023dual}
S.~Zhang, Y.~Zhang, Y.~Zhang, Y.~Wang, and Z.~Song, ``A dual-direction attention mixed feature network for facial expression recognition,'' {\em Electronics}, vol.~12, no.~17, p.~3595, 2023.

\bibitem{li2017reliable}
S.~Li, W.~Deng, and J.~Du, ``Reliable crowdsourcing and deep locality-preserving learning for expression recognition in the wild,'' in {\em 2017 IEEE Conference on Computer Vision and Pattern Recognition (CVPR)}, pp.~2584--2593, IEEE, 2017.

\bibitem{li2019reliable}
S.~Li and W.~Deng, ``Reliable crowdsourcing and deep locality-preserving learning for unconstrained facial expression recognition,'' {\em IEEE Transactions on Image Processing}, vol.~28, no.~1, pp.~356--370, 2019.

\bibitem{ngwe2023patt}
J.~L. Ngwe, K.~M. Lim, C.~P. Lee, and T.~S. Ong, ``Patt-lite: Lightweight patch and attention mobilenet for challenging facial expression recognition,'' {\em arXiv preprint arXiv:2306.09626}, 2023.

\bibitem{deng2019retinaface}
J.~Deng, J.~Guo, Y.~Zhou, J.~Yu, I.~Kotsia, and S.~Zafeiriou, ``Retinaface: Single-stage dense face localisation in the wild,'' 2019.

\bibitem{selvaraju2017grad}
R.~R. Selvaraju, M.~Cogswell, A.~Das, R.~Vedantam, D.~Parikh, and D.~Batra, ``Grad-cam: Visual explanations from deep networks via gradient-based localization,'' in {\em Proceedings of the IEEE international conference on computer vision}, pp.~618--626, 2017.

\bibitem{HE2024PCAD}
X.~He, R.~Happee, and M.~Wang, ``A new computational perceived risk model for automated vehicles based on potential collision avoidance difficulty (pcad),'' {\em Transportation Research Part C: Emerging Technologies}, vol.~166, p.~104751, 2024.

\bibitem{Mulakkal_Safety_Indicators2017}
F.~A. Mullakkal-Babu, M.~Wang, H.~Farah, B.~van Arem, and R.~Happee, ``Comparative assessment of safety indicators for vehicle trajectories on highways,'' {\em Transportation Research Record}, vol.~2659, no.~1, pp.~127--136, 2017.

\bibitem{peintner2024driving}
J.~Peintner, C.~Himmels, T.~Rock, C.~Manger, O.~Jung, and A.~Riener, ``Driving behavior analysis: A human factors perspective on automated driving styles,'' in {\em 2024 IEEE Intelligent Vehicles Symposium (IV)}, pp.~3325--3330, IEEE, 2024.

\bibitem{li2020deep}
S.~Li and W.~Deng, ``Deep facial expression recognition: A survey,'' {\em IEEE transactions on affective computing}, vol.~13, no.~3, pp.~1195--1215, 2020.

\bibitem{ekman1969repertoire}
P.~Ekman and W.~V. Friesen, ``The repertoire of nonverbal behavior: Categories, origins, usage, and coding,'' {\em semiotica}, vol.~1, no.~1, pp.~49--98, 1969.

\bibitem{beggiato2020facial}
M.~Beggiato, N.~Rauh, and J.~Krems, ``Facial expressions as indicator for discomfort in automated driving,'' in {\em Intelligent Human Systems Integration 2020: Proceedings of the 3rd International Conference on Intelligent Human Systems Integration (IHSI 2020): Integrating People and Intelligent Systems, February 19-21, 2020, Modena, Italy}, pp.~932--937, Springer, 2020.

\bibitem{mou2023driver}
L.~Mou, Y.~Zhao, C.~Zhou, B.~Nakisa, M.~N. Rastgoo, L.~Ma, T.~Huang, B.~Yin, R.~Jain, and W.~Gao, ``Driver emotion recognition with a hybrid attentional multimodal fusion framework,'' {\em IEEE Transactions on Affective Computing}, 2023.

\bibitem{meza2021driver}
B.~Meza-Garc{\'\i}a and N.~Rodr{\'\i}guez-Ib{\'a}{\~n}ez, ``Driver's emotions detection with automotive systems in connected and autonomous vehicles (cavs).,'' in {\em CHIRA}, pp.~258--265, 2021.

\bibitem{keogh2005exact}
E.~Keogh and C.~A. Ratanamahatana, ``Exact indexing of dynamic time warping,'' {\em Knowledge and information systems}, vol.~7, pp.~358--386, 2005.

\bibitem{salvador2007toward}
S.~Salvador and P.~Chan, ``Toward accurate dynamic time warping in linear time and space,'' {\em Intelligent Data Analysis}, vol.~11, no.~5, pp.~561--580, 2007.

\bibitem{Makowski2021neurokit}
D.~Makowski, T.~Pham, Z.~J. Lau, J.~C. Brammer, F.~Lespinasse, H.~Pham, C.~Schölzel, and S.~H.~A. Chen, ``{NeuroKit}2: A python toolbox for neurophysiological signal processing,'' {\em Behavior Research Methods}, vol.~53, pp.~1689--1696, 2 2021.

\bibitem{carlos2015biosppy}
C.~Carreiras, A.~P. Alves, A.~Louren\c{c}o, F.~Canento, H.~Silva, A.~Fred, {\em et~al.}, ``{BioSPPy}: Biosignal processing in {Python},'' 2022.
\newblock [Online; accessed 25/07/24].

\bibitem{smith1997scientist}
S.~W. Smith {\em et~al.}, ``The scientist and engineer's guide to digital signal processing,'' 1997.

\bibitem{posada2020innovations}
H.~F. Posada-Quintero and K.~H. Chon, ``Innovations in electrodermal activity data collection and signal processing: A systematic review,'' {\em Sensors}, vol.~20, no.~2, p.~479, 2020.

\bibitem{kumar2023comparative}
P.~S. Kumar, P.~K. GOVARTHAN, N.~GANAPATHY, and J.~F.~A. RONICKOM, ``A comparative analysis of eda decomposition methods for improved emotion recognition,'' {\em Journal of Mechanics in Medicine and Biology}, vol.~23, no.~06, p.~2340043, 2023.

\bibitem{lutin2021feature}
E.~Lutin, R.~Hashimoto, W.~De~Raedt, and C.~Van~Hoof, ``Feature extraction for stress detection in electrodermal activity.,'' in {\em BIOSIGNALS}, pp.~177--185, Vienna, Austria, 2021.

\bibitem{greco2015cvxeda}
A.~Greco, G.~Valenza, A.~Lanata, E.~P. Scilingo, and L.~Citi, ``cvxeda: A convex optimization approach to electrodermal activity processing,'' {\em IEEE transactions on biomedical engineering}, vol.~63, no.~4, pp.~797--804, 2015.

\bibitem{kim2004emotion}
K.~H. Kim, S.~W. Bang, and S.~R. Kim, ``Emotion recognition system using short-term monitoring of physiological signals,'' {\em Medical and biological engineering and computing}, vol.~42, pp.~419--427, 2004.

\bibitem{braithwaite2013guide}
J.~J. Braithwaite, D.~G. Watson, R.~Jones, and M.~Rowe, ``A guide for analysing electrodermal activity (eda) \& skin conductance responses (scrs) for psychological experiments,'' {\em Psychophysiology}, vol.~49, no.~1, pp.~1017--1034, 2013.

\end{thebibliography}

\begin{IEEEbiographynophoto}{Abel van Elburg}
received his MSc degree in Robotics Engineering from Delft University of Technology in 2024. This paper is based on his final thesis. Currently, he is working at a start-up in the Netherlands in the energy transition. 
\end{IEEEbiographynophoto}

\begin{IEEEbiographynophoto}{Konstantinos Gkentsidis}
holds a Master of Engineering (MEng) in Electrical and Computer Engineering from Democritus University of Thrace and an Advanced Master's degree (MSc) in Artificial Intelligence from KU Leuven. Since 2022, he has been involved in multiple European and Belgian-funded research projects on behalf of Siemens Digital Industries Software, specializing in occupant comfort assessment, and perception for autonomous vehicles.
\end{IEEEbiographynophoto}

\begin{IEEEbiographynophoto}{Mathieu Sarrazin}
received his first M.Sc. in Electromechanical-Electrotechnical Engineering in 2009 from the University of Gent, Campus Kortrijk, and a second M.Sc. in Mechanical Engineering with a minor in Automotive Engineering in 2012 from KU Leuven. In 2013, he has received the Marie Curie Fellowship award, recognizing his excellence in advanced research. Since 2012, he has been a key member of LMS International, now Siemens Industry Software (SISW). With over a decade of experience in applied research, Mathieu has made significant contributions to technologies related to electro-mechanical drivetrains, NVH, model-based system testing, hybrid and electric vehicles, autonomous systems, mechatronics, system identification, converter-machine interactions, control strategies, and condition monitoring.
In 2014, he received the Siemens Technology Award, and in 2016, he was honored as Siemens Researcher of the Month for his outstanding research achievements. Previously, he served as an R\&D manager, leading the definition, planning, and execution of national and European research projects. Currently, Mathieu is  Senior Technical Specialist and focuses on developing compelling value propositions and executing effective marketing strategies to accelerate the adoption and success of Siemens' innovative products in the marketplace.

\end{IEEEbiographynophoto}

\begin{IEEEbiographynophoto}{Sarah Barendswaard}
received her B.Sc. and M.Sc. degrees (cum laude) in Aerospace Engineering from Delft Univeristy of Technoogy. Her doctoral research focused on developing innovative haptic shared control systems for automotive applications, specifically creating a personalised lane-keeping assistance system that achieved a five-fold improvement in driver acceptance rates. Currently, she serves as a Research Engineer at Siemens, where her work centers on advancing occupant comfort modeling and comfort enhancement strategies for autonomous vehicles. Her research interests include human-machine interfaces, haptic feedback systems, passenger comfort optimization and modelling in automated driving systems.
\end{IEEEbiographynophoto}

\begin{IEEEbiographynophoto}{Varun Kotian}
received his MSc degree in Mechanical Engineering from Delft University of Technology in 2021. Currently, he is a PhD candidate at the Delft University of Technology, specializing in motion perception and motion sickness modelling. His research focuses on enhancing human comfort and safety in vehicles, particularly in the context of autonomous vehicles. 
\end{IEEEbiographynophoto}

\begin{IEEEbiographynophoto}{Riender Happee}
received the Ph.D. degree from TU Delft, The Netherlands, in 1992. He investigated road safety and introduced biomechanical human models for impact and comfort at TNO Automotive (1992-2007). Currently, he investigates the human interaction with automated vehicles focusing on motion comfort, perceived safety and acceptance at the Delft University of Technology, the Netherlands, where he is full Professor.
\end{IEEEbiographynophoto}

\clearpage
\begin{appendices}
    

\section{Questionnaire}
\label{appendix:questionnaire}


Rate the following 8 statements from 1 (do not agree at all) to 5 (completely  agree). When an automated vehicle is mentioned, automation level SAE4+ is meant. This means that the vehicle can drive fully automated, without human supervision. After every question you have the option to provide comments/context to your given answer. If anything is unclear, please ask the researcher.

\begin{enumerate}
    \question{I trust the technology behind automated vehicles to work correctly.} \\
    Not at all \quad 1 \quad 2 \quad 3 \quad 4 \quad 5 \quad Completely\\
    Why?
    \vspace{\baselineskip}
    
    \question{I think riding in an automated vehicle would be comfortable.}\\
    Not at all \quad 1 \quad 2 \quad 3 \quad 4 \quad 5 \quad Completely\\
    Why?
    \vspace{\baselineskip}
    
    \question{I would feel relaxed while being in an automated vehicle.}\\
    Not at all \quad 1 \quad 2 \quad 3 \quad 4 \quad 5 \quad Completely\\
    Why?
    \vspace{\baselineskip} 
    
    \question{I believe automated vehicles could reduce traffic accidents.}\\
    Not at all \quad 1 \quad 2 \quad 3 \quad 4 \quad 5 \quad Completely\\
    Why?
    \vspace{\baselineskip}
    
    \question{I would rather trust a fully automated vehicle than today’s drivers. \cite{weigl2021development}} \\
    Not at all \quad 1 \quad 2 \quad 3 \quad 4 \quad 5 \quad Completely\\
    Why?
    \vspace{\baselineskip}
    
    \question{I would be concerned about safety, if I was driven by a fully automated vehicle. \cite{weigl2021development}} \\
    Not at all \quad 1 \quad 2 \quad 3 \quad 4 \quad 5 \quad Completely\\
    Why?
    \vspace{\baselineskip}
    
    \question{I would not engage in non-driving related activities, but monitor the driving system. \cite{weigl2021development}} \\
    Not at all \quad 1 \quad 2 \quad 3 \quad 4 \quad 5 \quad Completely\\
    Why?
    \vspace{\baselineskip}
    
    \question{I would use a fully automated vehicle if they are available.\cite{weigl2021development}}\\
    Not at all \quad 1 \quad 2 \quad 3 \quad 4 \quad 5 \quad Completely\\
    Why?
    \vspace{\baselineskip}
    
    \question{Feel free to share any additional thoughts or comments on fully automated vehicles:}
    \vspace{1cm}
\end{enumerate}

\section{Continuous Rating Repeatability}
\label{appendix:rating_repeatability}

\begin{landscape}
    \begin{figure}[H]
        \centering
        \includegraphics[width=0.8\paperheight]{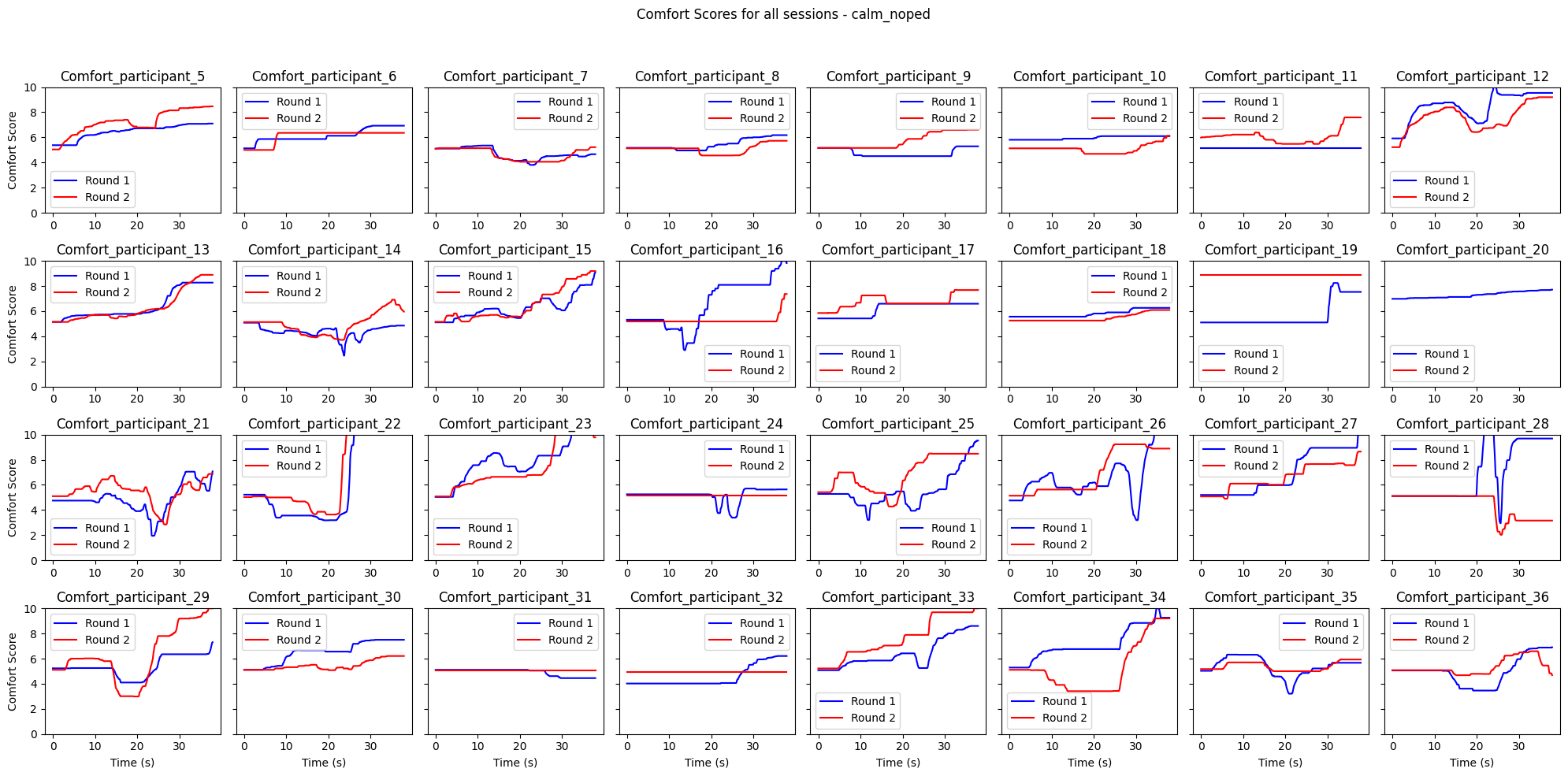}
        \caption{Individual subjective perceived risk responses for the calm drive without  pedestrian in the scene. Blue lines represent the first and red lines the the second round in this condition.}
        \label{fig:conscalmnoped}
    \end{figure}
\end{landscape}

\begin{landscape}
    \begin{figure}[H]
        \centering
        \includegraphics[width=0.8\paperheight]{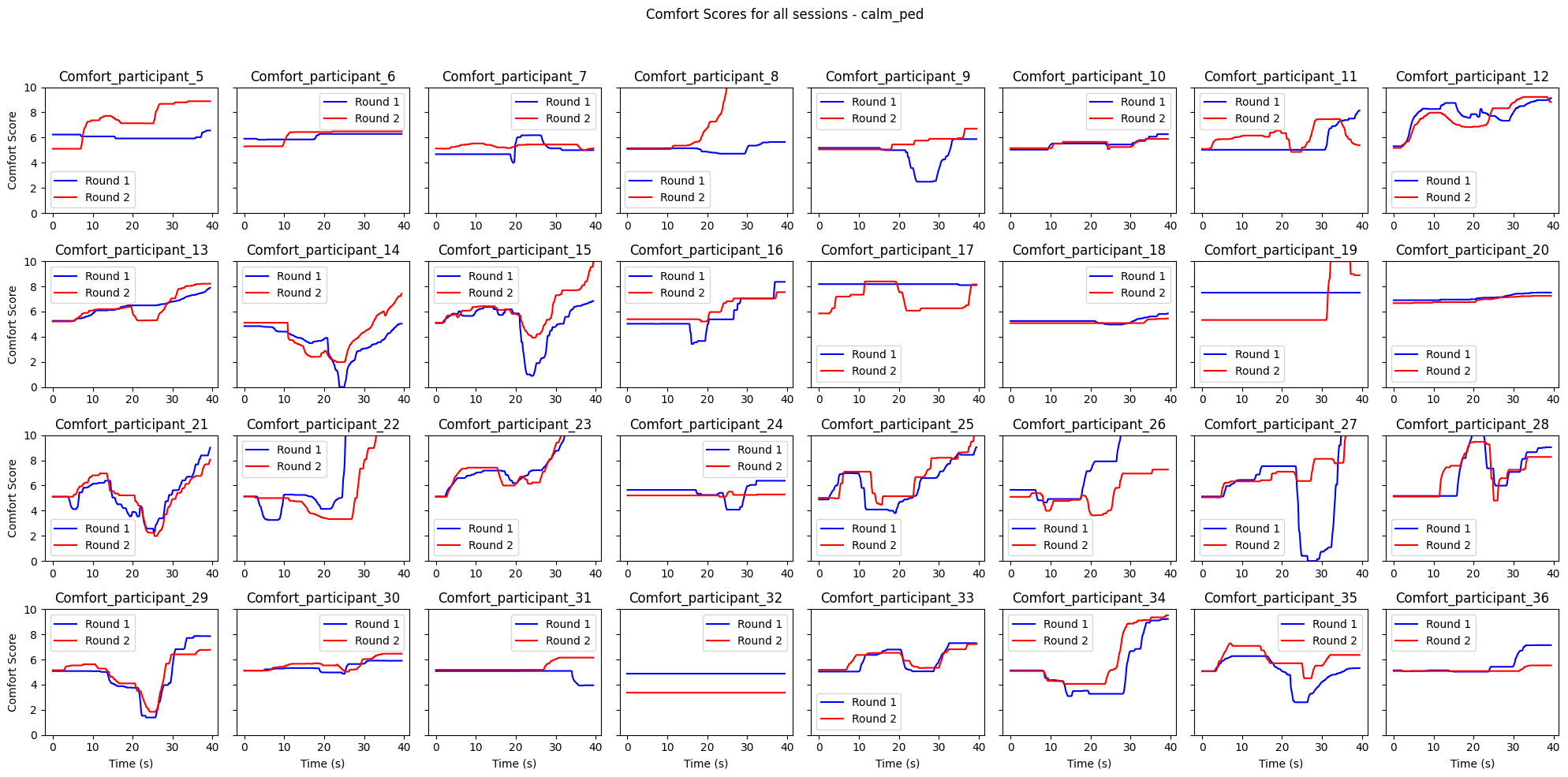}
        \caption{All subjective perceived risk responses for the calm drive with the pedestrian in the scene. Each blue line represents the response during the first round, and the red line the response during the second round.}
        \label{fig:conscalmped}
    \end{figure}
\end{landscape}

\begin{landscape}
    \begin{figure}[H]
        \centering
        \includegraphics[width=0.8\paperheight]{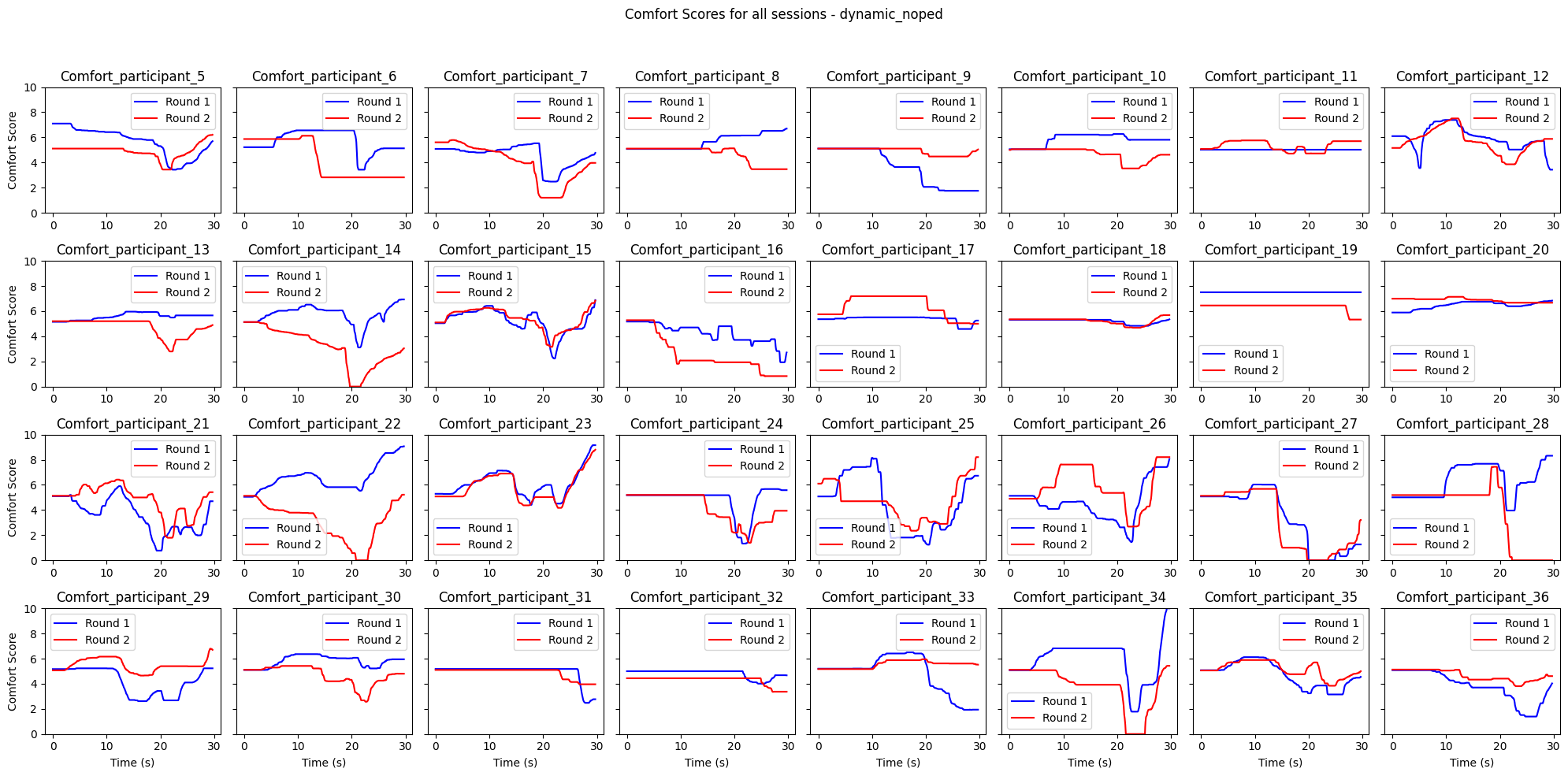}
        \caption{All subjective perceived risk responses for the dynamic drive without the pedestrian in the scene. Each blue line represents the response during the first round, and the red line the response during the second round.}
        \label{fig:consdynamicnoped}
    \end{figure}
\end{landscape}

\begin{landscape}
    \begin{figure}[H]
        \centering
        \includegraphics[width=0.8\paperheight]{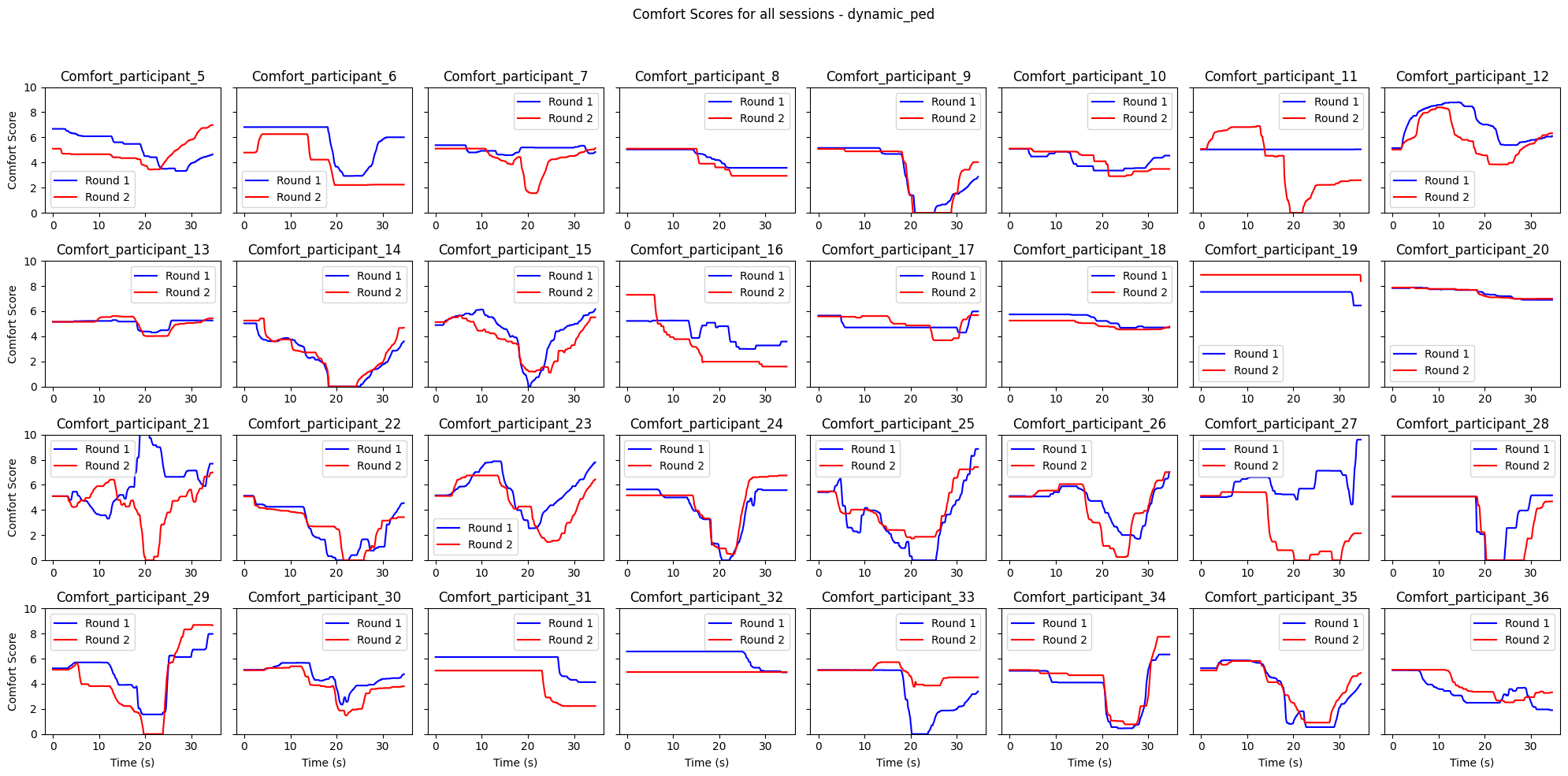}
        \caption{All subjective perceived risk responses for the dynamic drive with the pedestrian in the scene. Each blue line represents the response during the first round, and the red line the response during the second round.}
        \label{fig:consdynamicped}
    \end{figure}
\end{landscape}

All conditions were experienced two times by the participants, and all individual results are plotted in Figures \ref{fig:conscalmnoped}, \ref{fig:conscalmped}, \ref{fig:consdynamicnoped}, and \ref{fig:consdynamicped} where each figure captures one condition. The x-axis is time in seconds, and the y-axis the perceived risk score between 0 and 10. The blue line shows the score given during the first round, and the red line represents the second round. 

The data for Participant 20 during the second round for the calm driving style without pedestrian was lost, which is why this plot also has no red line. Participant 11 misunderstood the working of the knob during the first round, meaning this knob data is also not viable. All data for participant 19 was also excluded from further analyses, showing unrealistic scores and after the experiment the participant commented that it was normal if a car also "nudged" a pedestrian. This was considered very exceptional, and not relevant to how we desire automated vehicles to behave. Participants 31 and 32 seem to only adjust the perceived risk knob at the very end of the run, long after the critical condition presented itself, or not at all, leading to believe that they might have misunderstood the assignment, or forgot to use the knob till the very end. Their knob data was also excluded.  

The main purpose of repeating conditions was to check for consistency. If the subjective score in time is consistent over runs, it is more likely to show a true representation of the participants' state. Consistency was assessed visually by looking at the graphs, and dynamic time-warping (DTW) was applied starting at 10 seconds to account for initial offsets.

\subsection{DTW method}
The reliability of the continuous subjective scores was evaluated by analyzing consistency over the two runs for the same condition within participants using a method called Dynamic Time Warping (DTW) \cite{keogh2005exact}. DTW is a method used to measure similarity between time-series data, that may vary in time, speed, or both. It is generally used for applications like speech recognition and bio-informatics. Given two sequences \( X = (x_1, x_2, \ldots, x_n) \) and \( Y = (y_1, y_2, \ldots, y_m) \), the DTW algorithm works as follows \cite{salvador2007toward}:

1. \textbf{Distance Matrix}: Compute the distance matrix \( D \), where each element \( D(i, j) \) represents the squared difference between \( x_i \) and \( y_j \):
   \[
   D(i, j) = (x_i - y_j)^2
   \]

2. \textbf{Cumulative Distance Matrix}: Construct the cumulative distance matrix \( C \). The element \( C(i, j) \) represents the minimum cumulative distance to reach \( D(i, j) \):
   \[
   C(i, j) = D(i, j) + \min \left\{ 
       C(i-1, j),   
       C(i, j-1),   
       C(i-1, j-1)  
   \right\}
   \]
   with boundary conditions:
   \[
   C(0, 0) = D(0, 0)
   \]
   \[
   C(i, 0) = \sum_{k=1}^i D(k, 0) \quad \text{for } i \in [1, n]
   \]
   \[
   C(0, j) = \sum_{k=1}^j D(0, k) \quad \text{for } j \in [1, m]
   \]

3. \textbf{Warping Path}: The optimal warping path \( P \) is found by tracing back from \( C(n, m) \) to \( C(0, 0) \):
   \[
   P = (p_1, p_2, \ldots, p_L)
   \]
   where \( p_k = (i_k, j_k) \), \( i_L = n \), \( j_L = m \), \( i_1 = 1 \), and \( j_1 = 1 \).

\begin{figure}[htbp]
    \centering
    \begin{subfigure}[b]{0.45\textwidth}
        \centering
        \includegraphics[width=\textwidth]{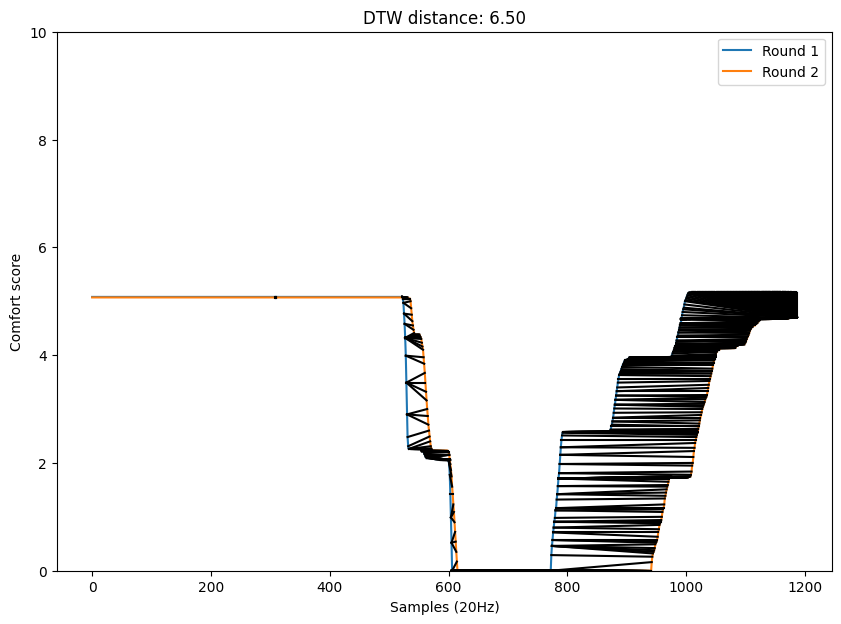}
        \caption{Low DTW distance, suggesting high consistency.}
        \label{fig:low_dtw}
    \end{subfigure}
    \hfill
    \begin{subfigure}[b]{0.45\textwidth}
        \centering
        \includegraphics[width=\textwidth]{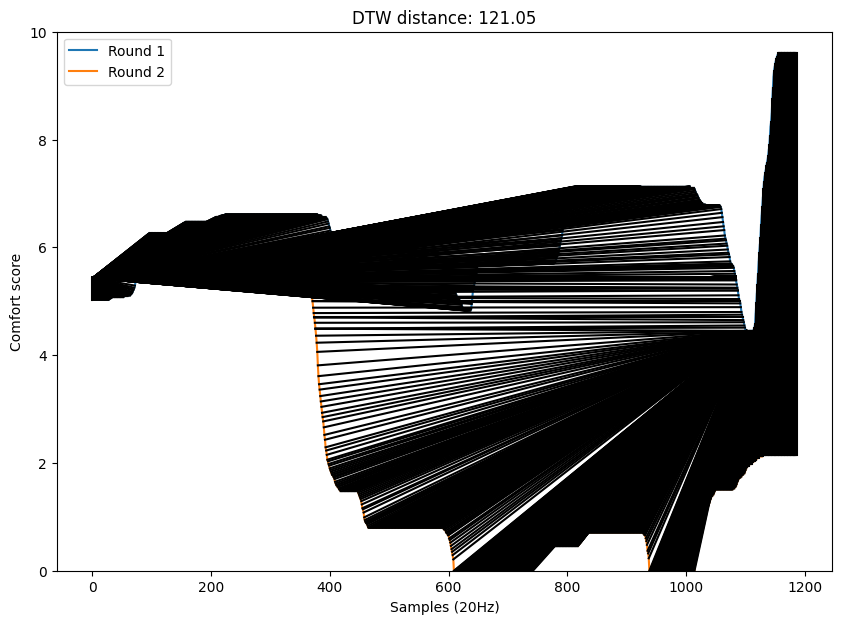}
        \caption{High DTW distance, suggesting low consistency.}
        \label{fig:high_dtw}
    \end{subfigure}
    \caption{Two plots showing the calculated paths between the two plots. The more distance these paths have to cover, the less aligned the two plots are. }
    \label{fig:dtw_comparison}
\end{figure}

The process is visualized for two of the runs in Figure \ref{fig:dtw_comparison}. The left plot shows a run with high consistency, resulting in a low DTW distance. The right plot shows a run with low consistency, resulting in a high DTW distance. The paths between the two plots are drawn. The shorter this accumulated path, the more similar the two plots are. This analyses is applied to all runs and the results are presented in section \ref{sec:analyses}.

\subsection{DTW-results}
The resulting DTW distances are presented in Table \ref{tab:dtw}. The higher the distance, the less similar the two plots for that condition are according to the DTW method. Participant 21 seems to have given the exact opposite rating during the first round of the dynamic drive with the pedestrian. Looking at their consistency in the other conditions, it is likely that they inverted the perceived risk scale during that run. 

\begin{table*}[h]
\centering
\caption{DTW distances for the continuous rating per participant per condition. Higher values indicate a mismatch between round 1 and 2 in the same condition. The excluded participant-condition combinations are made bold}
\label{tab:dtw}
\begin{tabular}{|c|l|l|l|l|}
\hline
\multicolumn{1}{|l|}{} &
  \multicolumn{1}{c|}{\textbf{calm\_noped}} &
  \multicolumn{1}{c|}{\textbf{calm\_ped}} &
  \multicolumn{1}{c|}{\textbf{dynamic\_noped}} &
  \multicolumn{1}{c|}{\textbf{dynamic\_ped}} \\ \hline
participant 5  & 30,21           & \textbf{66,97}  & 18,85           & 34,83           \\ \hline
participant 6  & 15,01           & 6,13            & 38,7            & 61,53           \\ \hline
participant 7  & 6,47            & 18,31           & 19,56           & \textbf{37,47}  \\ \hline
participant 8  & 10,85           & \textbf{110,08} & \textbf{53,61}  & 14,19           \\ \hline
participant 9  & 35,05           & \textbf{42,33}  & \textbf{56,95}  & 10,76           \\ \hline
participant 10 & 28,04           & 5,67            & 46              & 13,69           \\ \hline
participant 12 & 13,35           & 12,9            & 20,21           & 25,26           \\ \hline
participant 13 & 8,07            & 11,88           & 38,52           & 6,72            \\ \hline
participant 14 & 28,55           & 29,18           & \textbf{75,99}  & 7,71            \\ \hline
participant 15 & 9,26            & 43,7            & 8,79            & 15,41           \\ \hline
participant 16 & \textbf{38,66}  & 22,5            & \textbf{57,96}  & 41,51           \\ \hline
participant 17 & 24,86           & 48,38           & 34,92           & 16,83           \\ \hline
participant 18 & 7,83            & 5,6             & 3,29            & 8,67            \\ \hline
participant 20 &                 & 5,36            & 7,5             & 1,45            \\ \hline
participant 21 & 21,04           & 9,9             & 15,42           & \textbf{62,89}  \\ \hline
participant 22 & 19,17           & 15,54           & \textbf{115,36} & 9,83            \\ \hline
participant 23 & 26,09           & 5,69            & 9,72            & 20,93           \\ \hline
participant 24 & 21,44           & 24,93           & 25,01           & 19,93           \\ \hline
participant 25 & 21,72           & 9,73            & 28,6            & 28,61           \\ \hline
participant 26 & 32,78           & \textbf{64,16}  & 46,8            & 17,05           \\ \hline
participant 27 & 14,08           & \textbf{108,59} & 8,77            & \textbf{120,66} \\ \hline
participant 28 & \textbf{143,46} & 13,81           & \textbf{120,31} & 6,5             \\ \hline
participant 29 & 43,42           & 17,53           & 40,05           & 30,64           \\ \hline
participant 30 & 36,73           & 11,42           & 38,85           & 12,88           \\ \hline
participant 33 & 30,17           & 8,46            & \textbf{56,27}  & \textbf{65,94}  \\ \hline
participant 34 & \textbf{81,51}  & 18,21           & \textbf{63,86}  & 19,76           \\ \hline
participant 35 & 18,58           & 34,63           & 11,13           & 9,92            \\ \hline
participant 36 & 29,04           & 28,24           & 33,62           & 23,56           \\ \hline
\end{tabular}
\end{table*}

\begin{figure}[H]
    \centering
    \includegraphics[width=\linewidth]{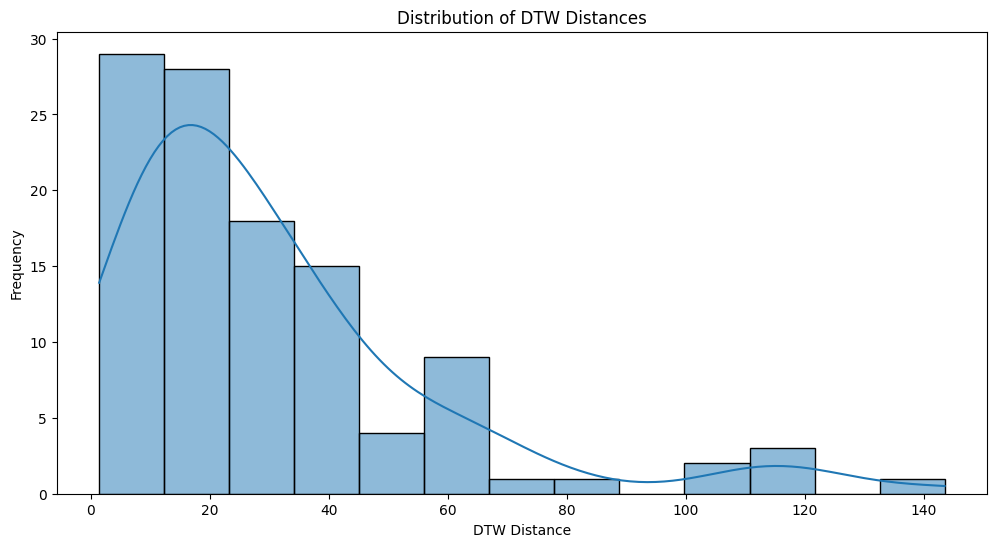}
    \caption{The distribution of the DTW distances.}
    \label{fig:dtwdist}
\end{figure}

The distribution of the DTW analyses is shown in Figure \ref{fig:dtwdist}. This gives an overview and a general idea of what can be reasonably expected from the participants. A soft cut-off was selected, at the 85th percentile, corresponding to a value of 54.94. This was then combined with some visual inspection. Some exceptions were made, and explained here:

\begin{itemize}
    \item \textbf{Participant 16, calm noped:} A score of 38.88 but round 2 barely changes and round 1 a lot. the long straights make it seem similar, but the absolute difference between the two ratings is big for most of the run. 
    \item \textbf{Participant 9, calm ped:} A score of 42,37. The score is not too low only due to the exact overlap up to almost 20 seconds into the run, however, after that the rounds diverge, feeling perceived risk in round 1 and slight perceived risk in round 2. 
    \item \textbf{Participant 8, dynamic noped:} A score of 53.61, almost the cut-off. The moment the person changes their perceived risk, they do the exact opposite in either round. It is possible that they inverted the scale mistakenly in one of the rounds.
    \item \textbf{Participant 6, dynamic ped:} A score of 61.53, mainly because at the end they set their perceived risk back up in round 1, and not in round 2. However, before that during the condition the trends are similar, so it was decided to keep this in. 
    \item \textbf{Participant 7, dynamic ped:} A score of 37.47. Again, barely changing perceived risk in round 1, but in round 2 reporting strong perceived risk around 20 seconds. 
\end{itemize}

The final selection of participants whose perceived risk score is excluded from further processing is in Table \ref{tab:excluded}. 

\begin{table}[H]
\centering
\caption{Excluded \textbf{comfort knob data} for participants per condition based on DTW and visual inspection of the plots. *for this condition, only the first round of participant 21 is removed. }
\begin{tabular}{|l|l|}
\hline
condition & Excluded Participants \\ \hline
calm\_noped & 11, 16, 19, 28, 31, 32, 34 \\ \hline
calm\_ped & 5, 8, 9, 11, 19, 27, 31, 32 \\ \hline
dynamic\_noped & 8, 9, 11, 14, 19, 22, 28, 31, 32, 33, 34 \\ \hline
dynamic\_ped & 7, 11, 19, 21*, 27, 31, 32, 33 \\ \hline
\end{tabular}
\label{tab:excluded}
\end{table}

\section{Sensors for the experiment}
\label{appendix:Phys_sensors}

GSR and heart activity (ECG) were measured using a NeXus-10 MKII from MindMedia. 

\begin{figure}[H]
    \centering
    \includegraphics[angle=-90, width=0.9\linewidth]{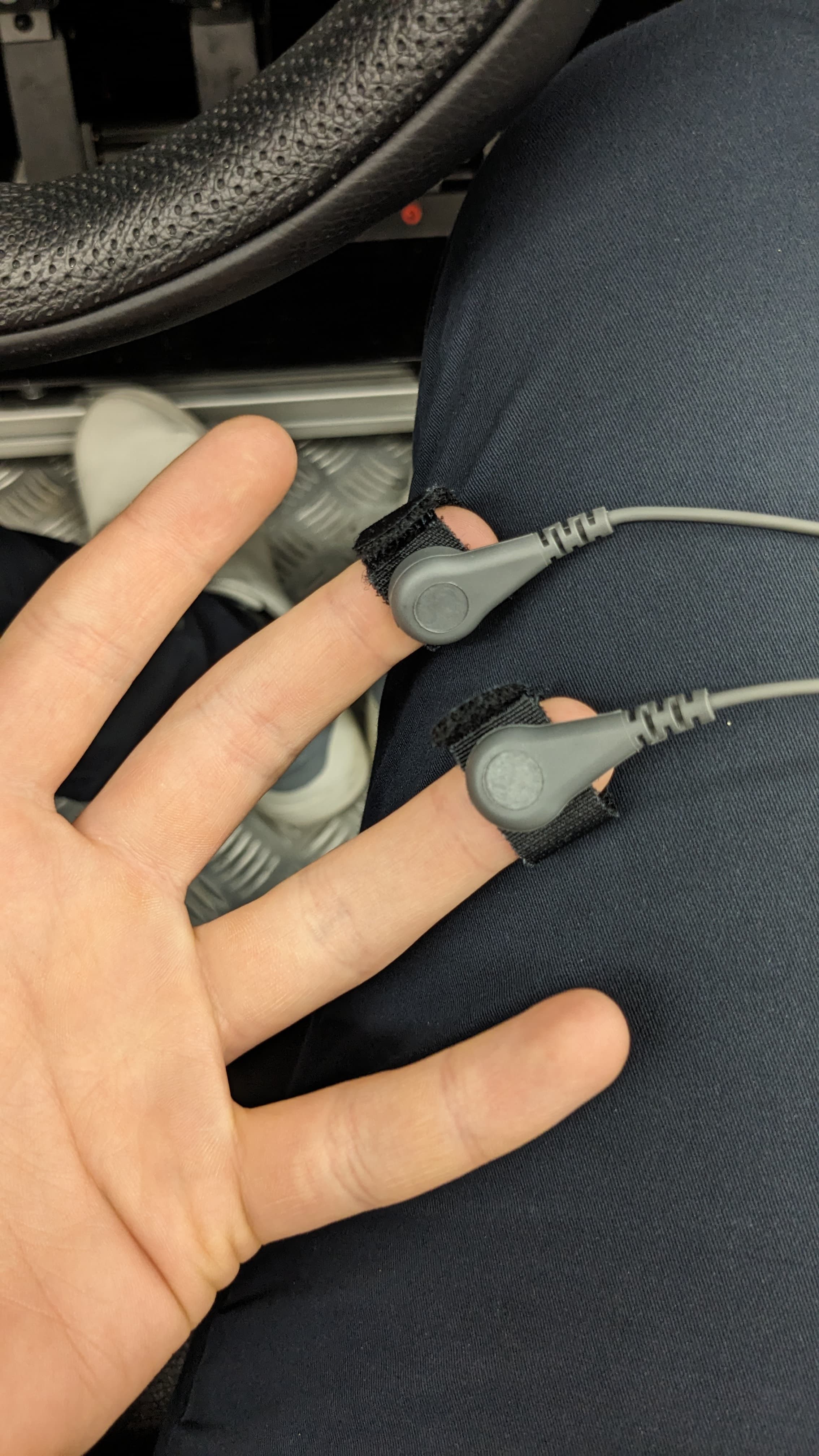}
    \caption{GSR sensors attached to the fingertips of the non-dominant hand. They were not color-coded because it does not matter which goes on which finger, as long as they are on the same hand.}
    \label{fig:velcro}
\end{figure}

The GSR was measured via two electrodes attached to two fingertips of the same hand with velcro straps, as can be seen in Figure \ref{fig:velcro}. 
For this experiment, the participant was asked what was their dominant hand, and how they would operate the comfort knob. The velcro straps were attached to the fingertips on the hand that would not operate the comfort knob, to limit electrode movements that could affect the measurements. With the GSR electrodes, it is important that there is some time (1-2 minutes) between attaching the sensor and starting the measurement, as the skin under and around the velcro straps has to accommodate to the presence of the straps. During this experiment, after the velcro straps were attached, the other sensors were explained, and a test run was performed, ensuring sufficient time before starting the measurements. 

The GSR data was processed using the \href{https://neuropsychology.github.io/NeuroKit/introduction.html}{NeuroKit2} python package \cite{Makowski2021neurokit} as described in appendix D. 

\begin{figure}[H]
    \centering
    \includegraphics[width=0.9\linewidth]{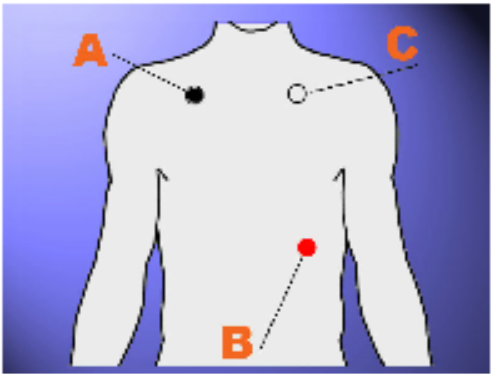}
    \caption{Illustration on where to place the electrodes for ECG measurement. This is the front view of the torso, so from the perspective of the participant: A is located top-right, B bottom-left, and C top-left. A (black electrode) is the negative input, and B (the red electrode) is the positive input. C (the white electrode) is the ground electrode.}
    \label{fig:hrelectrodes}
\end{figure}

The ECG was measured via three electrodes placed on the torso of the participant. The electrodes were color-coded and placed as illustrated in Figure \ref{fig:hrelectrodes}. 
The specific sticky electrodes used in this experiment were the F9060 electrodes (for adults, 48x50 millimeter) produced by \href{https://www.fiab.it/en/results.php?s=F9060}{FIAB}. The participant attached these stickers themselves, under clear instruction from the researcher. 
ECG was collected for future research and not analysed in this paper.

Eye-tracking was performed with the Pupil Invisible glasses by Pupil-Labs. 
These glasses contain a 6 degrees of freedom IMU, a scene camera for first person view, a microphone to capture audio, and two infrared eye cameras to capture eye videos and perform real-time gaze estimation. The glasses are connected via USB-C to a smartphone that has the Invisible Companion application installed to set-up measurements and start data acquisition. 

\begin{figure}[H]
    \centering    
    \begin{subfigure}[b]{0.9\linewidth}
    \centering  
    \includegraphics[width=\linewidth]{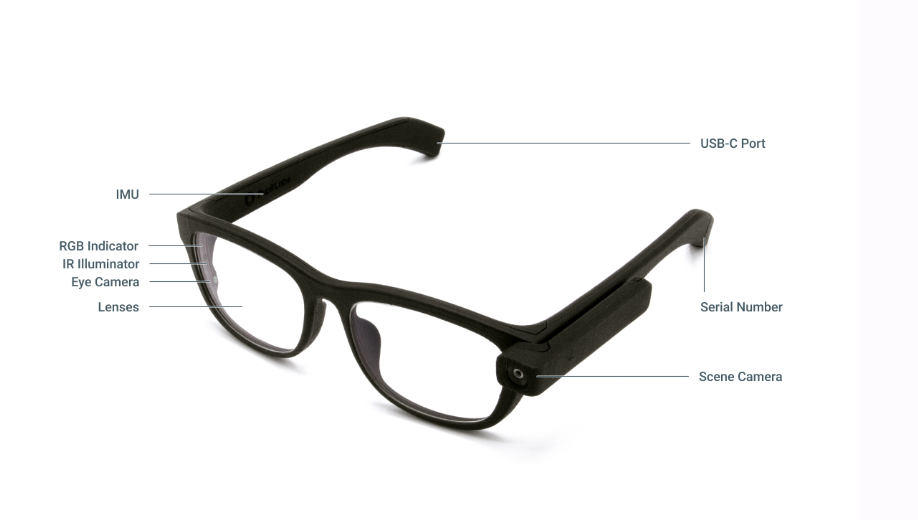}
    \label{fig:pupilhardware}
    \end{subfigure}
    \hfill
    \caption{The Pupil Invisible glasses with eye tracker from Pupil-Labs.}
    \label{fig:pupilinvisible}
\end{figure}

\begin{figure}[H]
    \centering
    \includegraphics[width=0.9\linewidth]{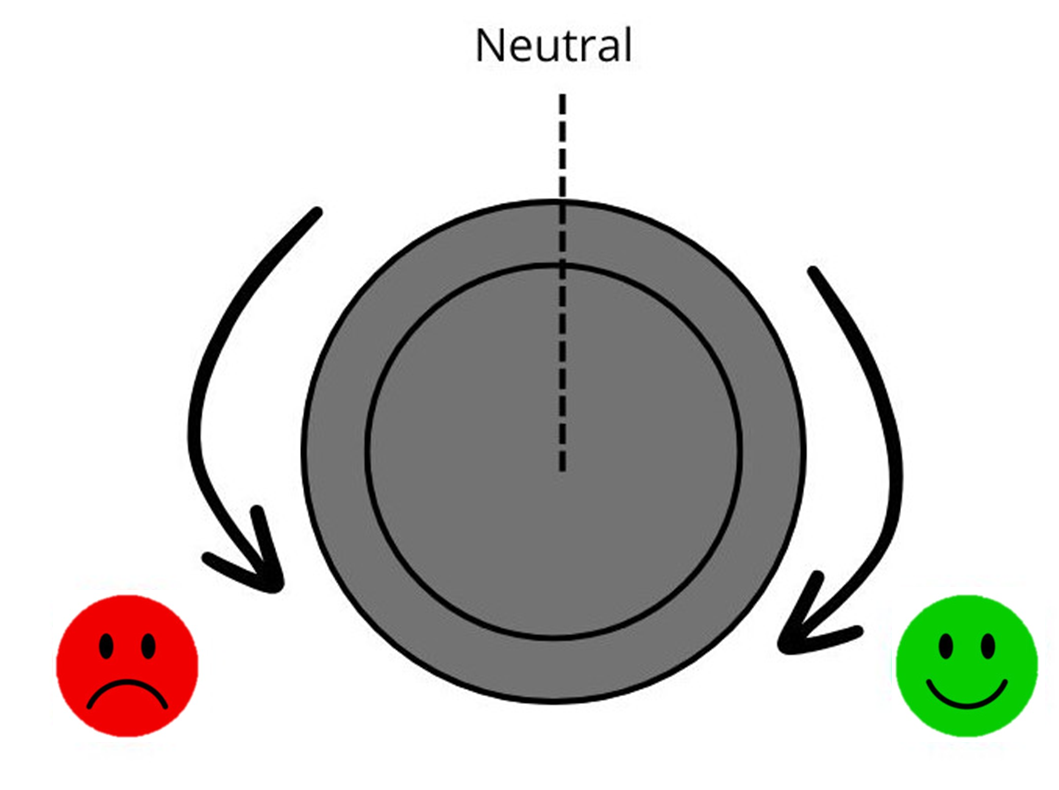}
    \caption{The illustration that was in the top middle of the screen for the participants, to remind the participant which way to turn the knob.}
    \label{fig:smiley}
\end{figure}
\begin{figure}[H]
    \centering    
    \includegraphics[width=0.8\linewidth,angle=90]{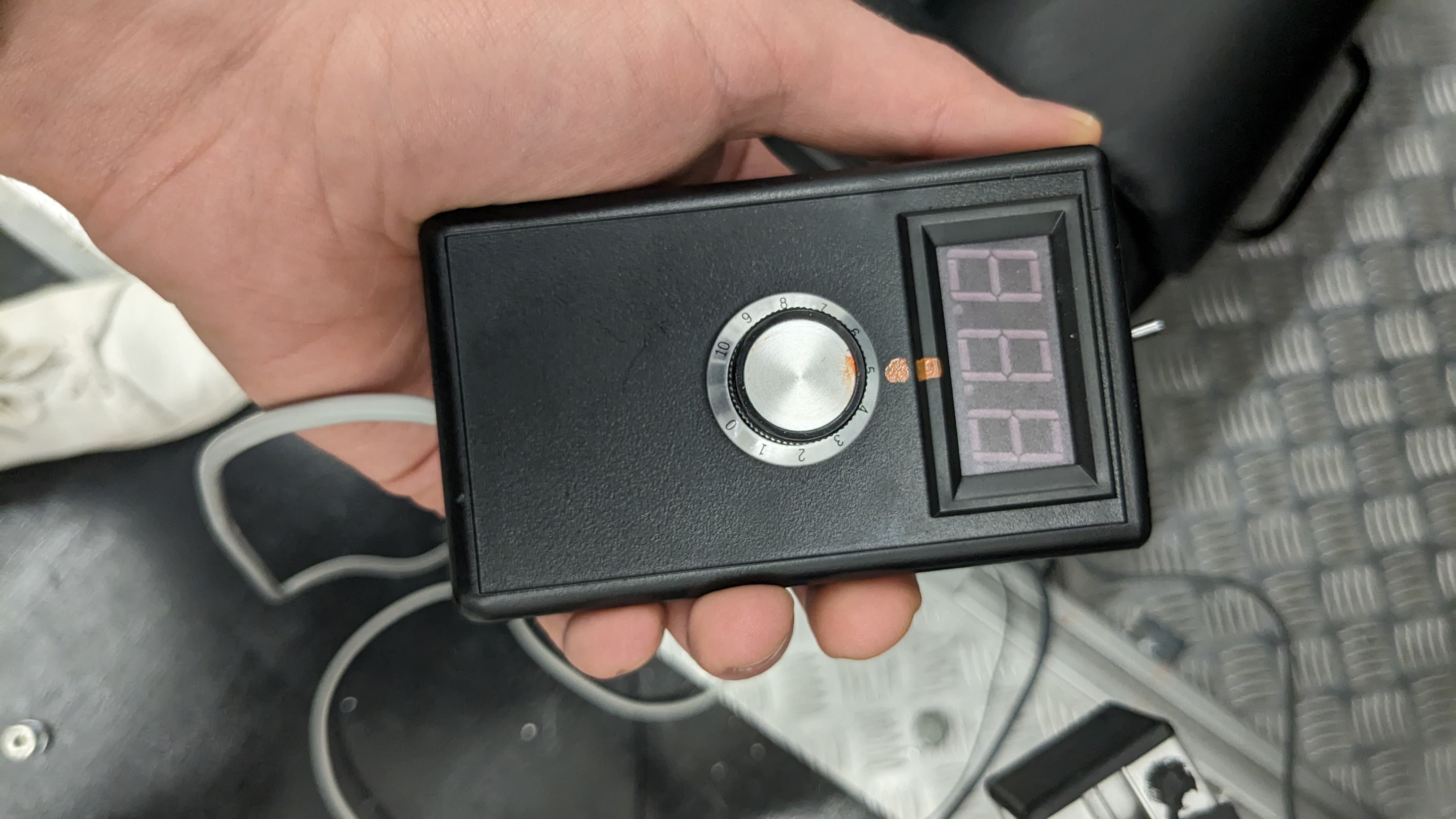}
    \caption{The Comfort Knob. A marking was made on the neutral position, and a figure was placed in the screen to help the participant remember which way to turn the knob.}
    \label{fig:knob}
\end{figure}

The Comfort-Knob is a rotational potentiometer connected 
to continuously measure the subjective comfort of the participant. It is a turning knob where the participant gives a value between 0 and 10. The scale was defined as 0 being extremely uncomfortable, 5 being neutral, and 10 being extremely comfortable. This is based on the \href{https://www.sae.org/standards/content/j1060_201405/}{SAE J1060} for subjective ratings related to ride comfort in motor vehicles, where 0 up to and including 4 is considered unacceptable, 5 is borderline, and 6 to 10 is acceptable \cite{cieslak2020accurate}. Except for the knob not turning past the limits of 0 and 10, there was no haptic feedback for the participant to feel how far they had turned the knob. Figure \ref{fig:smiley} was constantly present in the top-middle of the screen to remind the participant which way to turn the knob according to their state. This illustration was made very simple  to limit distraction. 
On the device itself is a LED screen that also shows the current value in 2 decimals. Participants were instructed not to focus on the exact value, as this would be distracting. A picture of the comfort knob is presented in Figure \ref{fig:knob}.

\begin{figure}[H]
    \centering    
    \includegraphics[width=0.9\linewidth]{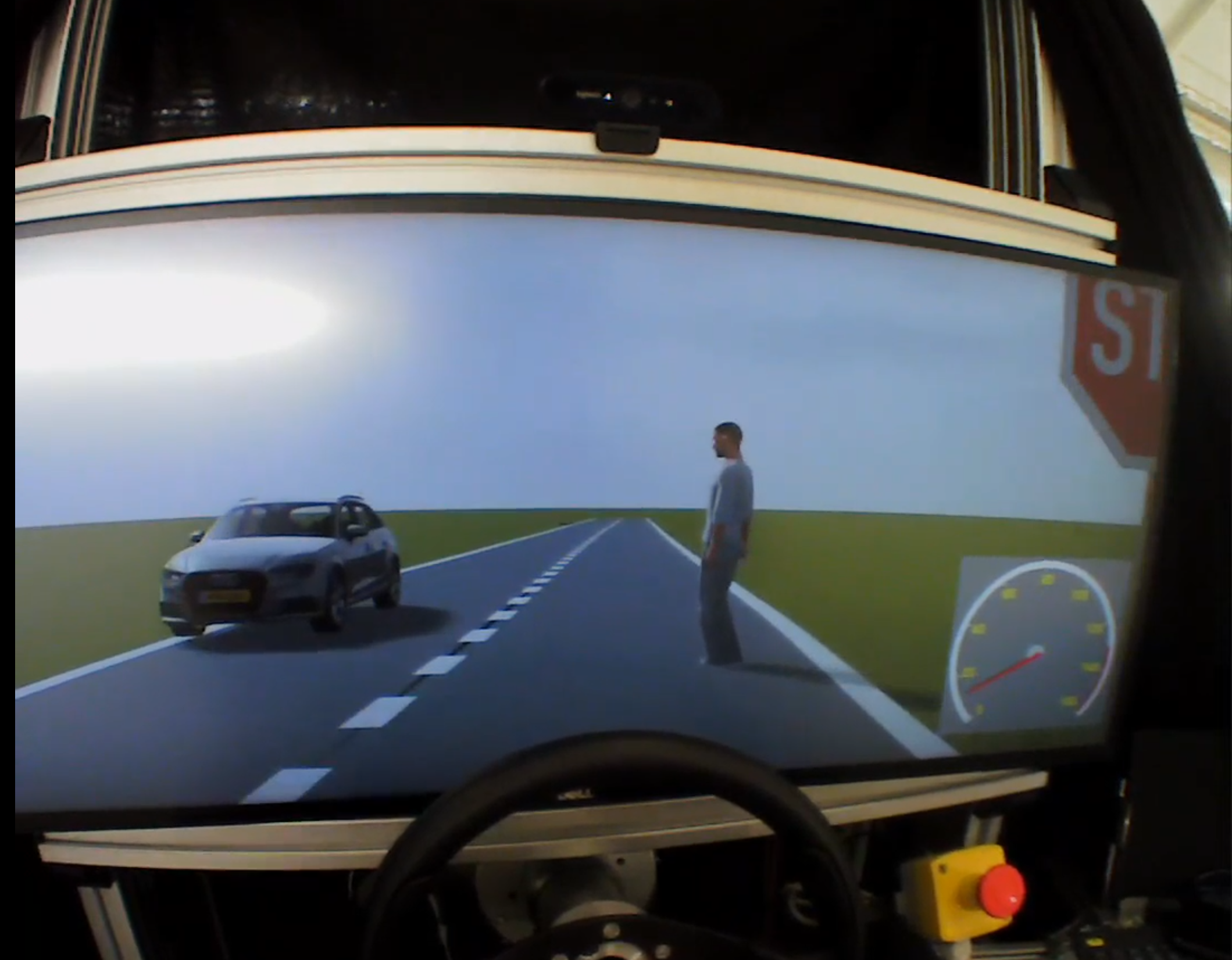}
    \caption{First person view from the participants. The webcam recording the face was located in the top middle above the screen.}
    \label{fig:firstperson}
\end{figure}

Figure \ref{fig:firstperson} shows the first-person view of the participant recorded from the head of the participant.
The figure shows the scenario with the steering wheel, speedometer and emergency button. 

\section{GSR analysis}
\label{appendix:GSR_analysis}

For processing of the GSR data, the \href{https://neuropsychology.github.io/NeuroKit/introduction.html}{NeuroKit2} python package was used \cite{Makowski2021neurokit}. This is an open source package for processing physiological signals. It is completely open-source, and invites anyone to contribute. Contributors are from all over the world and can be found with their affiliation \href{https://neuropsychology.github.io/NeuroKit/authors.html}{here}. They incorporate different processing methods, both self-developed and implemented from literature. They also incorporate methods from BioSPPy, another python package for physiological signal processing \cite{carlos2015biosppy}. BioSPPy is marked as archived on github, whereas the Neurokit package is still being updated. 

\begin{figure}[H]
    \centering    \includegraphics[width=0.9\linewidth]{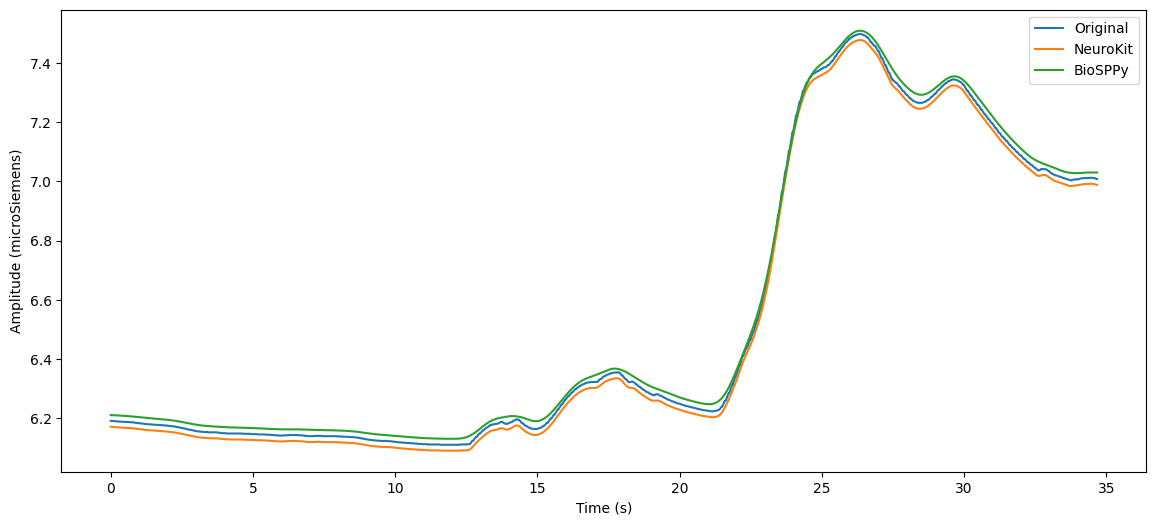}
    \caption{The raw GSR signal and the two cleaned signals. Both methods use a Butterworth filter, but BioSPPy is more aggressive and also includes smoothing. The offset both signals have from the original signal is implemented manually. }
    \label{fig:clean}
\end{figure}

First a low-pass Butterworth filter was applied to clean the signal. As movement artifacts are likely to occur due to the dynamic nature of the experiment, the decision was made to go for a more aggressive approach, also smoothing the signal after the Butterworth filter conform \cite{smith1997scientist}. The difference between these two approaches can be seen in Figure \ref{fig:clean}. The Neurokit plot is only a Butterworth filter, and the BioSPPy plot also has smoothing. Note that a small offset was given to the plots, to make them better visible. This offset was implemented manually purely for this image, and is not present in the data itself. 

\begin{figure}[H]
    \centering    \includegraphics[width=0.9\linewidth]{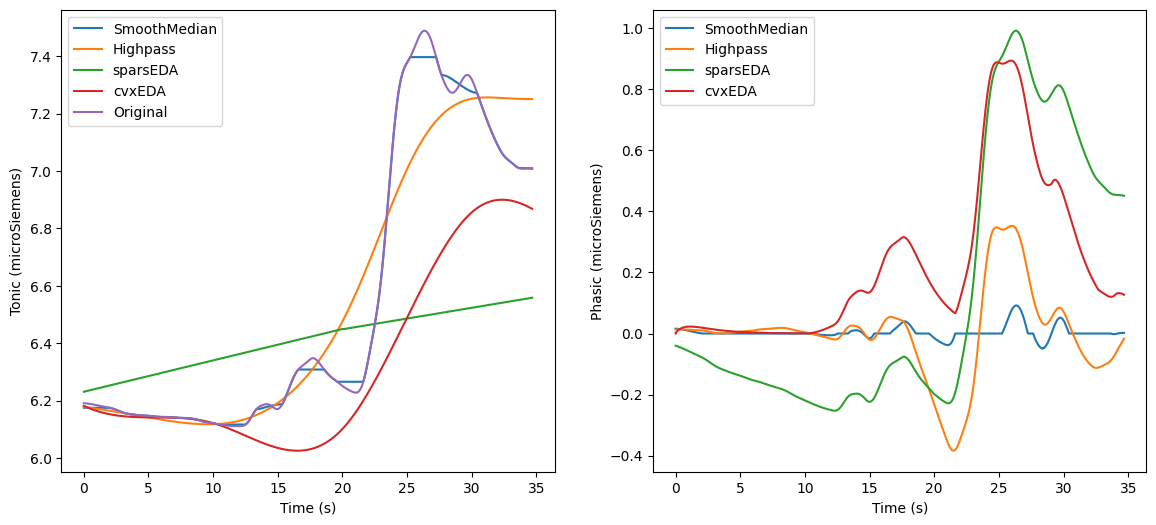}
    \caption{The Tonic (left) and Phasic (right) components of a signal, split through different methods.}
    \label{fig:tonicphasic}
\end{figure}

GSR is generally split in a Tonic and a Phasic component. The Tonic component, also known as the Skin Conductance Level (SCL) represents slow changes. The Phasic component represents fast responses, which are generally called Skin Conductance Responses (SCRs). Event-related SCRs generally arises 1-5 seconds after the event. To split these components different methods are used (see \cite{posada2020innovations} for a discussion). Different methods can give varying outputs, as can also be seen in Figure \ref{fig:tonicphasic}. According to literature sparsEDA performs best for classifying event-related stress, also specifically in the context of active driving through the Phasic component \cite{kumar2023comparative, lutin2021feature}. However, a downside is a less accurate Tonic component \cite{lutin2021feature}. CvxEDA \cite{greco2015cvxeda} has a better tonic component and also performs well in the classification of the Phasic component, and was therefore selected for this paper. 

\begin{figure}[H]
    \centering    \includegraphics[width=0.9\linewidth]{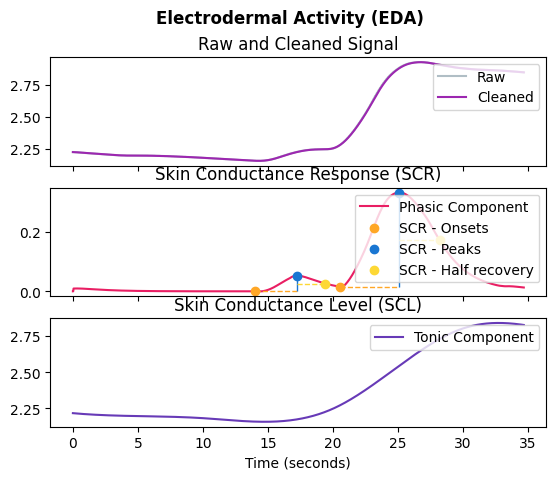}
    \caption{An overview of the processing of a GSR signal. First the signal is cleaned (top), then the Phasic (middle) and Tonic (bottom) components are extracted, and peaks are defined (middle). }
    \label{fig:processed}
\end{figure}

In the Phasic component, peak detection was applied using the popular method by \cite{kim2004emotion}, whom performed emotion classification from short physiological signals. These peaks represent the Skin Conductance Responses, and are expressed in three components. The onset, where the rise starts, the amplitude at its peak, and the half-recovery time, when the level is halfway back to the base level. A summary of this processing is given in Figure \ref{fig:processed}. Here we also see that SCR has a zero baseline due to the applied high pass filtering which simplifies the analysis. 

\begin{figure}[H]
    \centering    \includegraphics[width=0.9\linewidth]{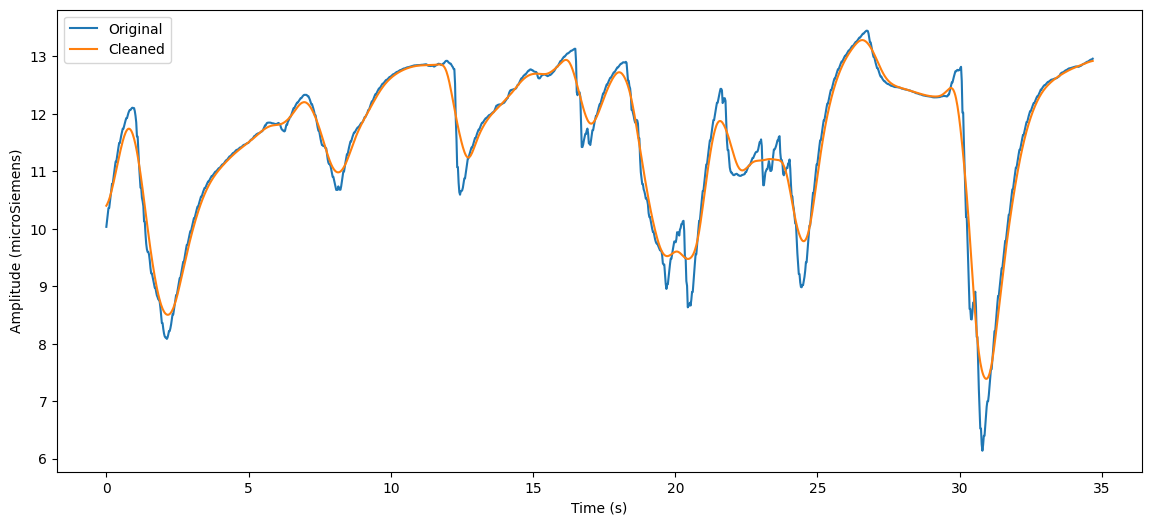}
    \caption{Faulty GSR with extreme drops. }
    \label{fig:faultyGSR}
\end{figure}

Before analysis, the GSR data was inspected. From one participant, the data showed a response that was unlikely to be physiological. This response, of which one run is shown in Figure \ref{fig:faultyGSR}, showed abnormally fast and large drops in the GSR, also showing response amplitudes much higher than 3 \textmu Siemens \cite{braithwaite2013guide}. Something else that was noticed, was that a lot of responses were relatively low. Normal GSR values do not go below 1\textmu Siemens \cite{braithwaite2013guide}, which did occur in this dataset. A problem could have been that the velcro straps were not tight enough, resulting in the electrode to momentarily lose contact with the skin. A cut-off was implemented, excluding participants whose data dropped below 1 \textmu Siemens, even if it was momentarily because the full response was deemed no longer reliable. This resulted in excluding 6 participants, which is close to 20\% of the data. To avoid this, one-time-use sticky electrodes could be used. 

\subsection{GSR relates to driving style}
\label{subsec:GSR_driving_style}
The Facial Expression analyses did not provide the desired results, even though the continuous subjective ratings showed significant drops in comfort, in particular during the dynamic driving style. To validate whether this effect was only present in the subjective ratings, or also present in the physiological data, the GSR was analysed.

The most significant variable of the experiment was found to be the driving style, and to not let the validation of GSR depend on the individual subjective perceived risk scores, an event-based analysis was done. As explained above the phasic component, or the Skin Conductance Response (SCR), is known to be an indicator of events that cause arousal or stress. Peak detection was performed, resulting in dictionaries with onsets, the moment in time that the peak starts, peaks, the moment the phasic component is at its highest, and the amplitudes, the difference between the peak value and the onset value. This detection is also depicted in Figure \ref{fig:processed}.

As features, the times and amplitude of the highest peak for each run were given to a Random Forest Classifier. The classifier then predicted if this was a calm or dynamic drive. Because results for such classifiers can depend on the random seed that is used, the fitting and prediction were repeated 10 times with a different seed. The classifier could always reach a perfect fit on the training data, with a mean accuracy of 100\% and a standard deviation of 0. On the testing data, the mean accuracy was 74.3\% with a standard deviation of 4.7\%. This analysis was not fine-tuned, as the goal was merely to show that GSR can be utilized to classify events with high perceived risk in this dataset. By fine-tuning further, both in feature selection and the classifier itself, it is likely possible to achieve better results. 

These GSR analyses confirm that the comfort drop, though not visible in the facial expressions of most participants, is present in physiological data. 

\section{Facial expression as function of time}
\label{appendix:plots}

\begin{figure}[H]
    \centering    \includegraphics[width=\linewidth]{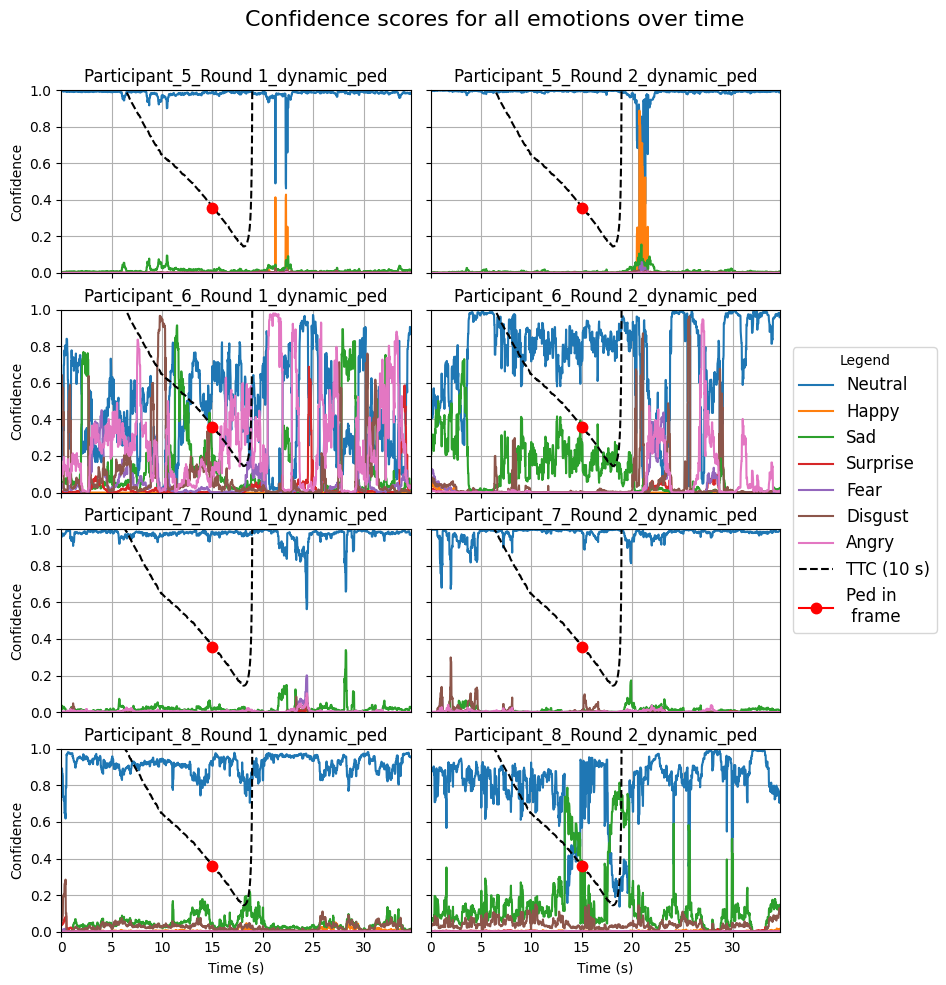}
    \caption{All emotions confidence scores during the dynamic scenario with the pedestrian in the scene, after applying the moving average filter, for participants 5 through 8.}
    \label{fig:allemotions58}
\end{figure}

\begin{figure}[H]
    \centering
    \includegraphics[width=\linewidth]{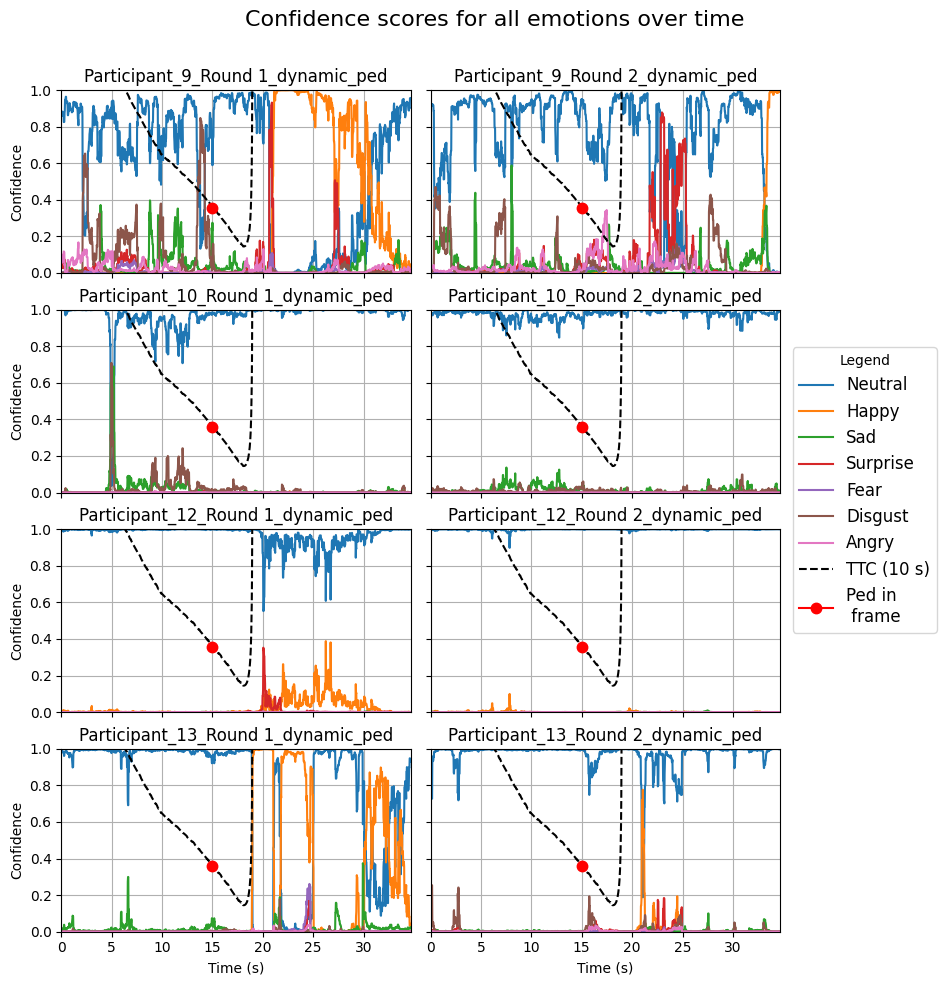}
    \caption{All emotions confidence scores during the dynamic scenario with the pedestrian in the scene, after applying the moving average filter, for participants 9 through 13, excluding 11.}
    \label{fig:allemotions913}
\end{figure}

\begin{figure}[H]
    \centering
    \includegraphics[width=\linewidth]{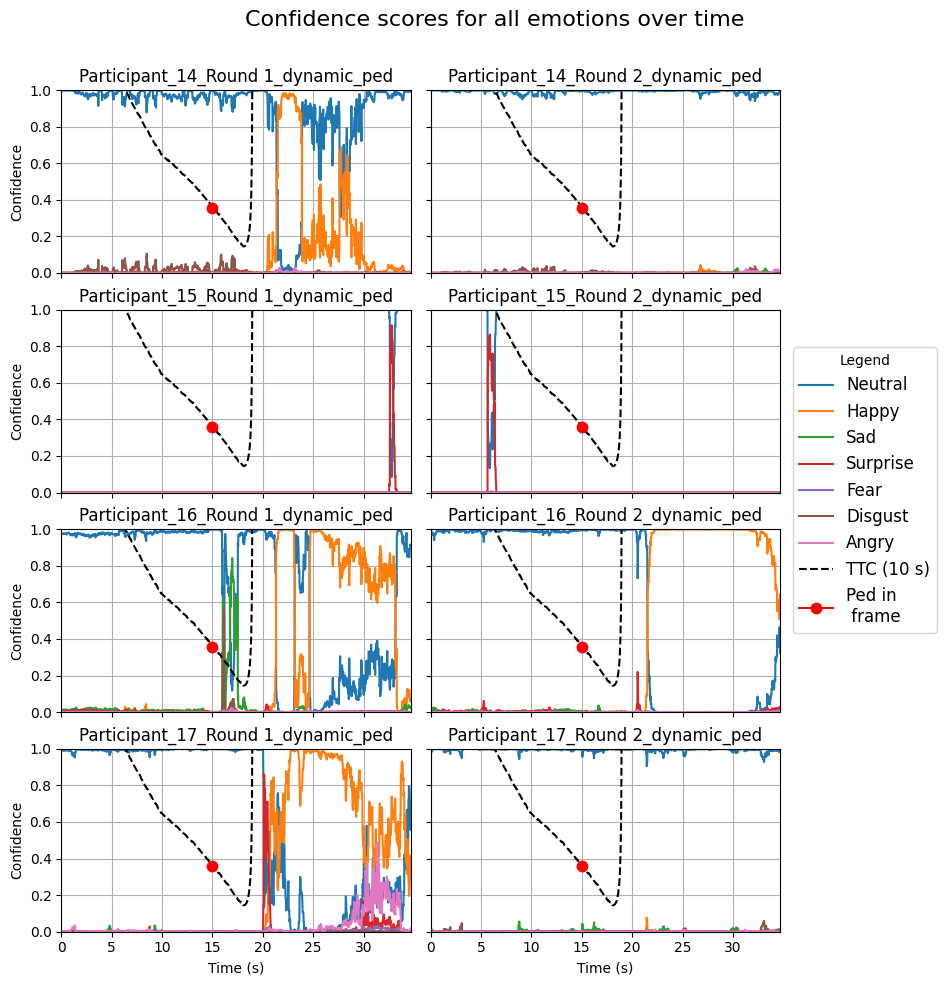}
    \caption{All emotions confidence scores during the dynamic scenario with the pedestrian in the scene, after applying the moving average filter, for participants 14 through 17.}
    \label{fig:allemotions1417app}
\end{figure}

\begin{figure}[H]
    \centering
    \includegraphics[width=\linewidth]{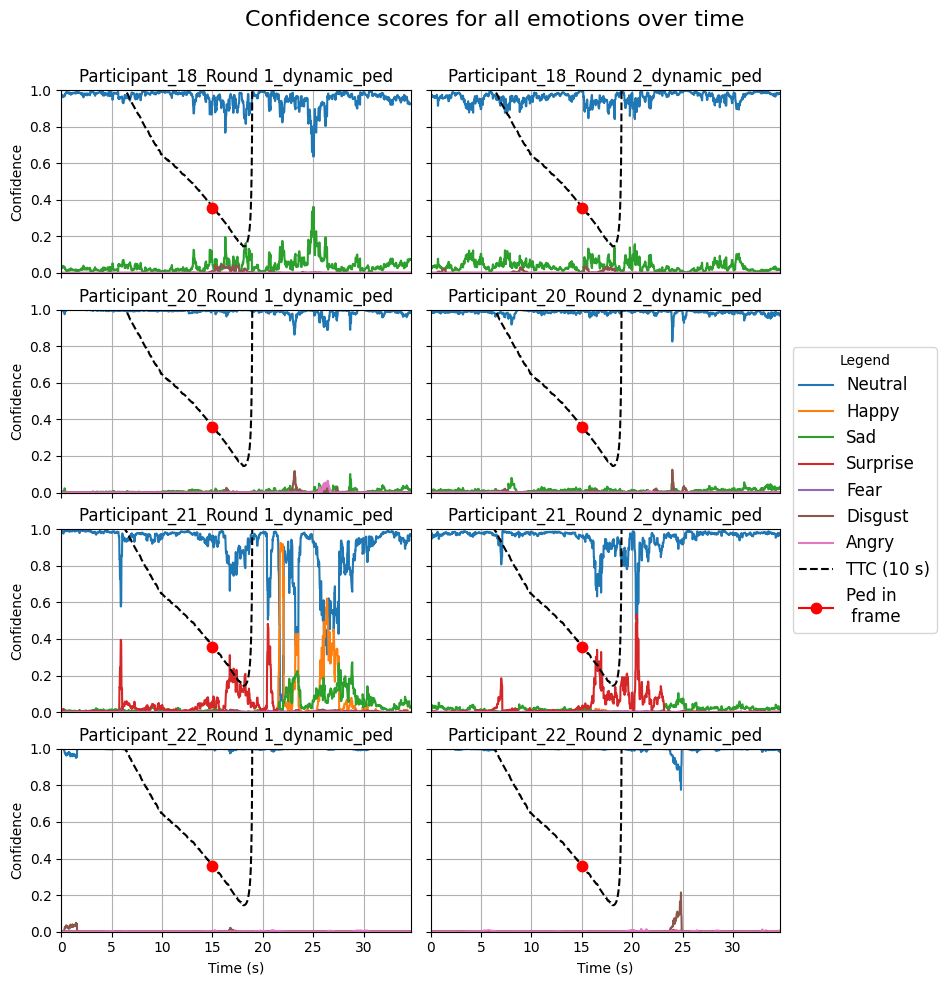}
    \caption{All emotions confidence scores during the dynamic scenario with the pedestrian in the scene, after applying the moving average filter, for participants 18 through 22, excluding 19.}
    \label{fig:allemotions1822}
\end{figure}

\begin{figure}[H]
    \centering
    \includegraphics[width=\linewidth]{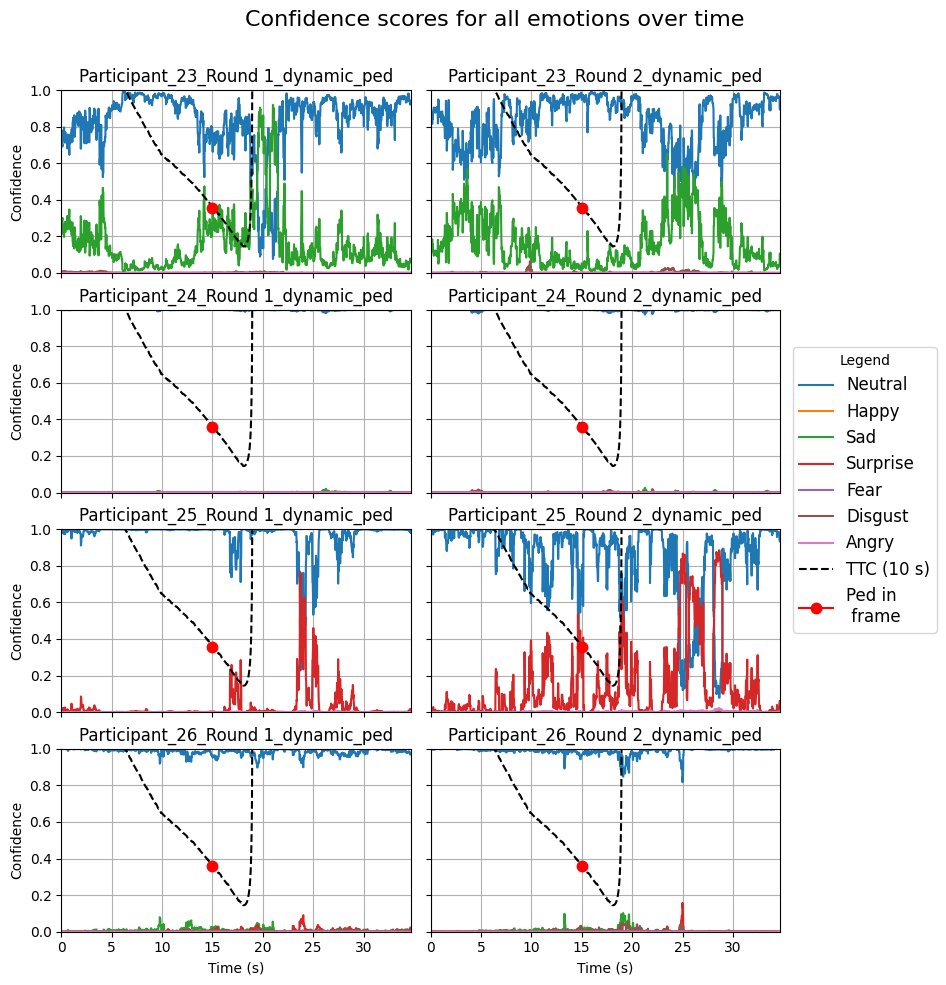}
    \caption{All emotions confidence scores during the dynamic scenario with the pedestrian in the scene, after applying the moving average filter, for participants 23 through 26.}
    \label{fig:allemotions2326}
\end{figure}

\begin{figure}[H]
    \centering
    \includegraphics[width=\linewidth]{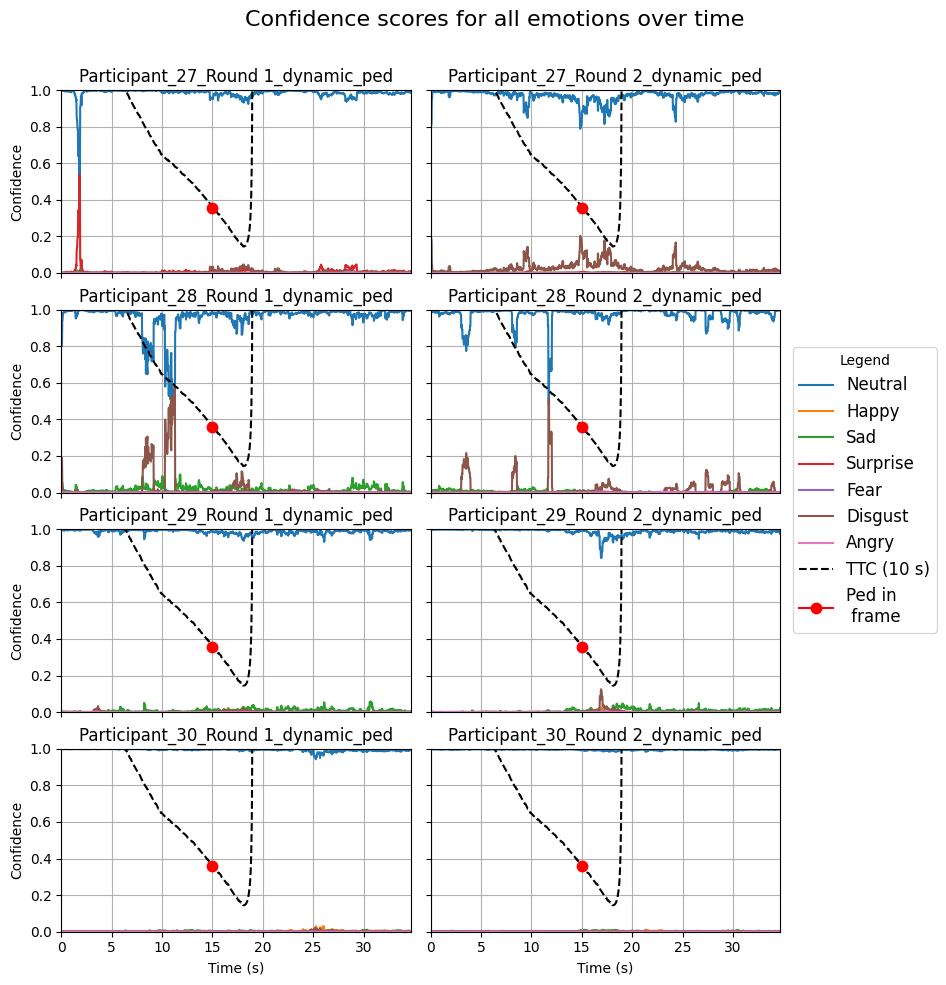}
    \caption{All emotions confidence scores during the dynamic scenario with the pedestrian in the scene, after applying the moving average filter, for participants 27 through 30.}
    \label{fig:allemotions2730}
\end{figure}

\begin{figure}[H]
    \centering
    \includegraphics[width=\linewidth]{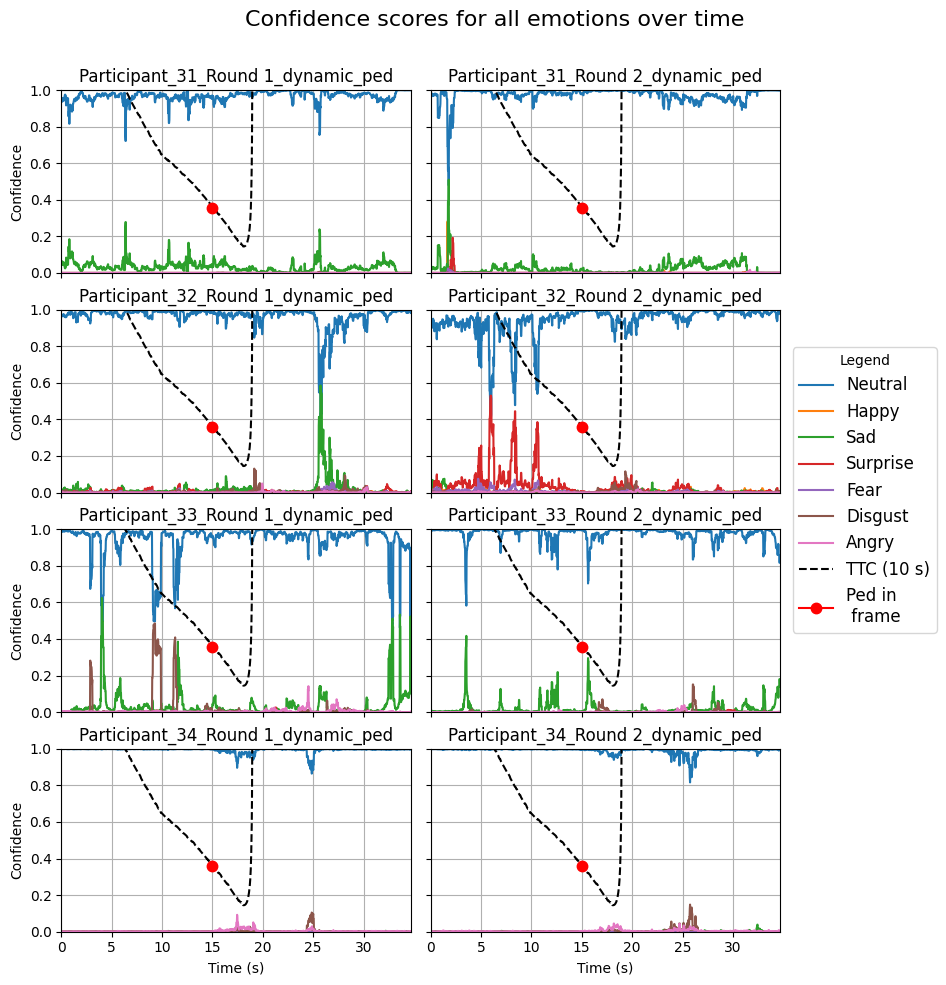}
    \caption{All emotions confidence scores during the dynamic scenario with the pedestrian in the scene, after applying the moving average filter, for participants 31 through 34.}
    \label{fig:allemotions3134}
\end{figure}

\begin{figure}[H]
    \centering
    \includegraphics[width=\linewidth]{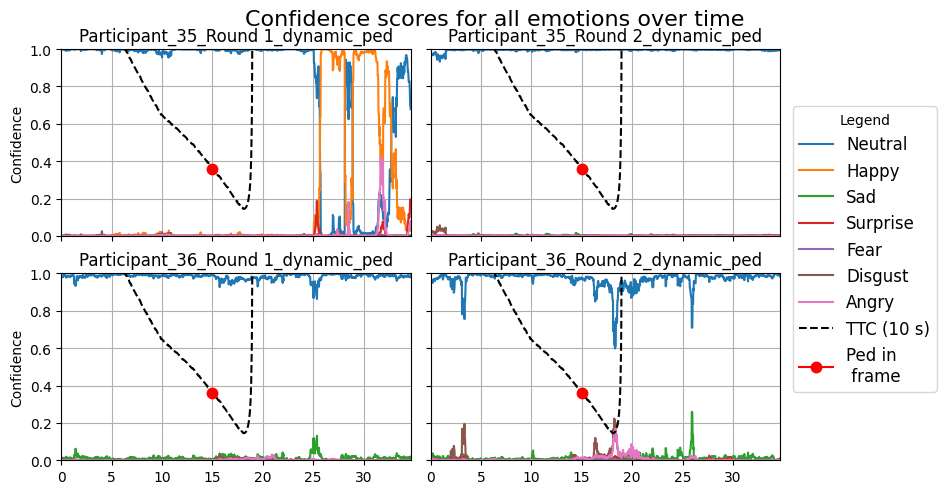}
    \caption{All emotions confidence scores during the dynamic scenario with the pedestrian in the scene, after applying the moving average filter, for participants 35 and 36.}
    \label{fig:allemotions3536}
\end{figure}


\section{Neural Network Model Architecture}
\label{appendix:architecture}

\begin{figure}[H]
    \centering
    \includegraphics[angle=-90, width=0.7\linewidth]{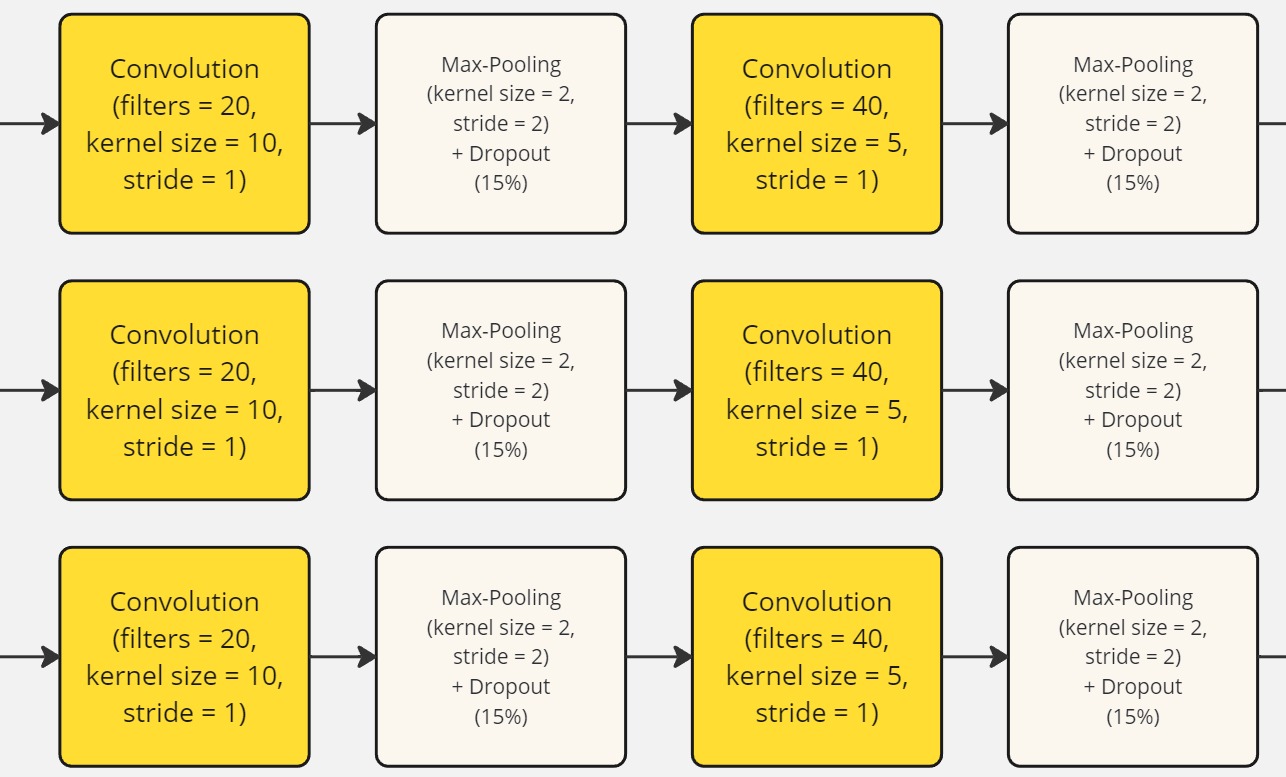}
    \caption{Part 1/3 of the architecture.}
    \label{fig:arch1}
\end{figure}

\begin{figure}[H]
    \centering
    \includegraphics[angle=-90, width=0.7\linewidth]{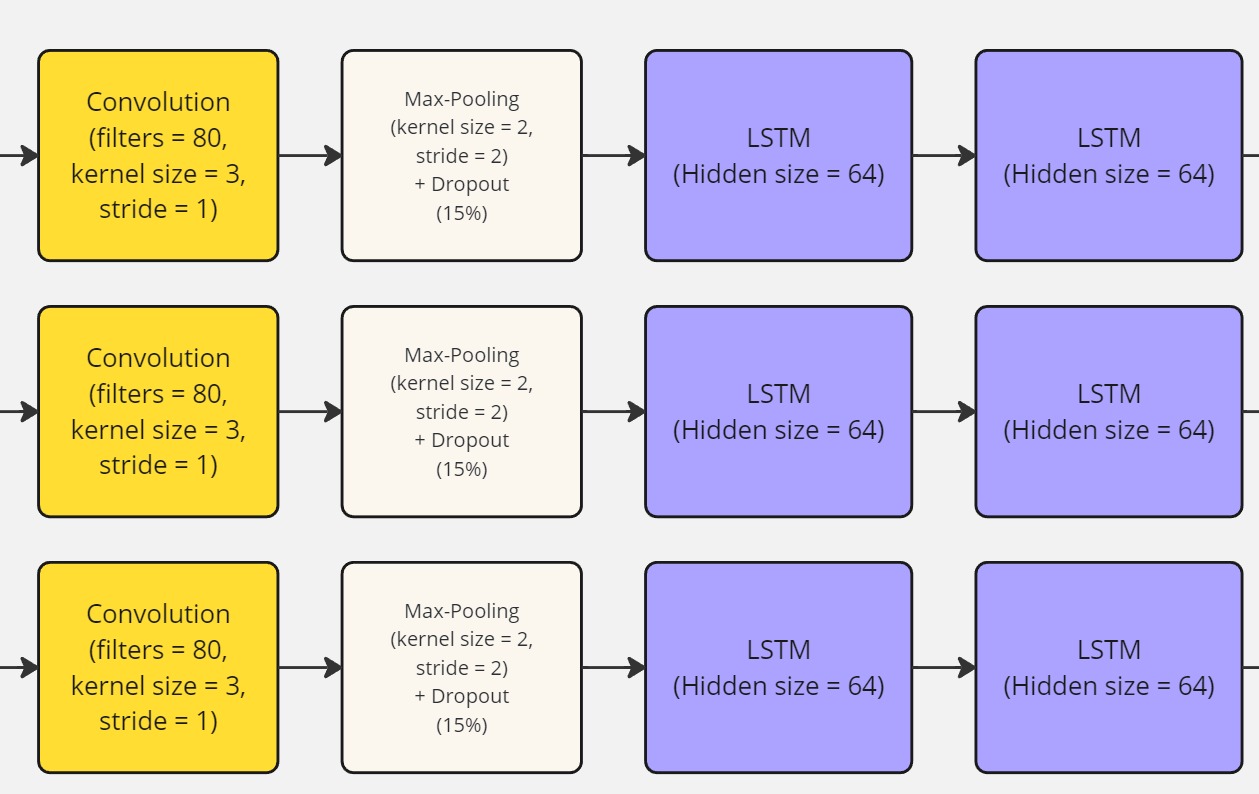}
    \caption{Part 2/3 of the architecture.}
    \label{fig:arch2}
\end{figure}

\begin{figure}[H]
    \centering
    \includegraphics[angle=-90, width=0.7\linewidth]{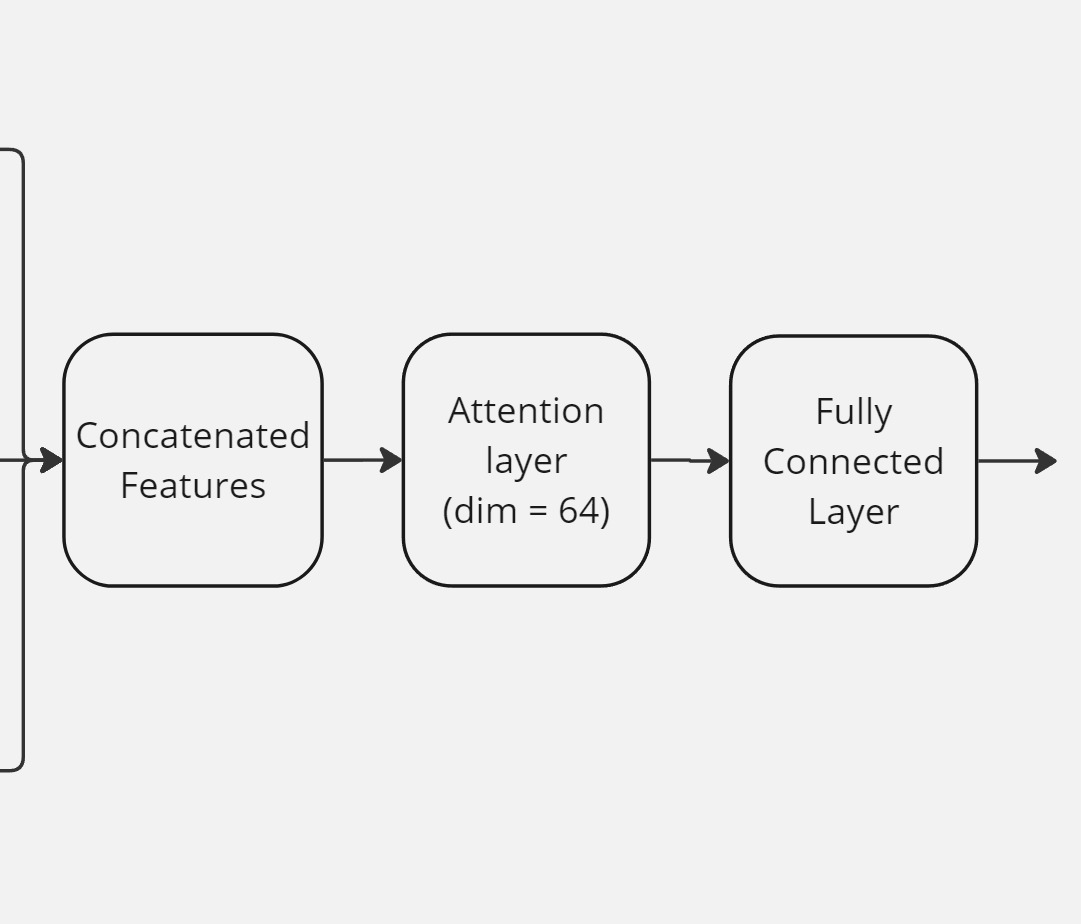}
    \caption{Part 3/3 of the architecture.}
    \label{fig:arch3}
\end{figure}

\section{Distribution of the data in training and validating the neural network}
\begin{figure}[H]
    \centering    \includegraphics[width=0.9\linewidth]{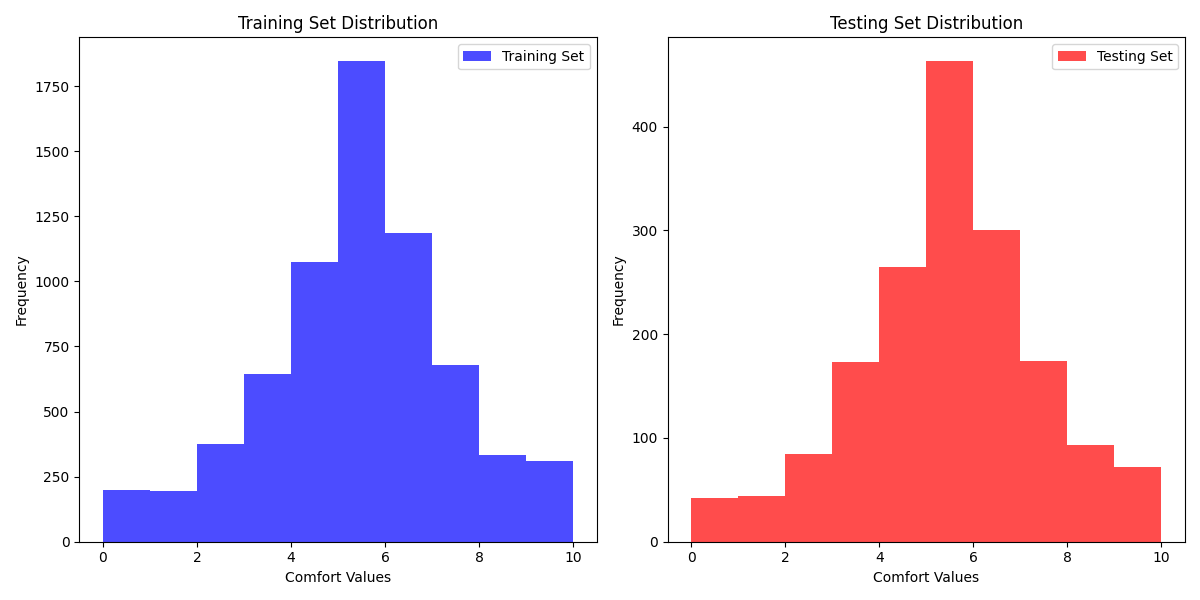}
    \caption{Data distribution of the training and validation set, when sampling with a constant stride of 0.5 seconds.}
    \label{fig:distribution}
\end{figure}

\begin{figure}[H]
    \centering    \includegraphics[width=0.9\linewidth]{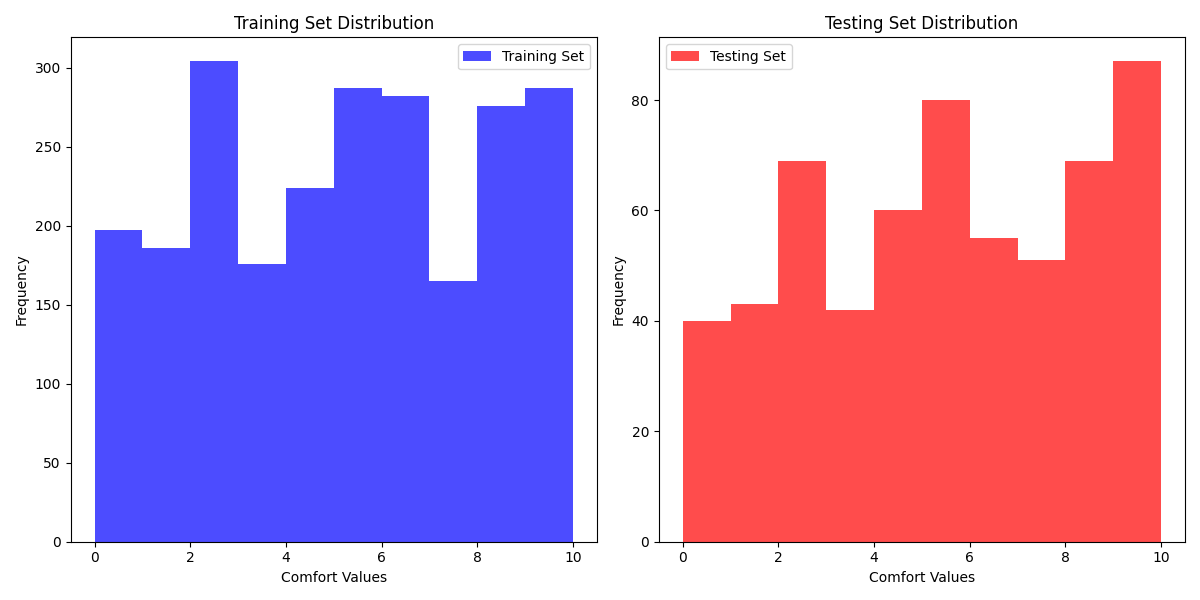}
    \caption{Data distribution of the training and validation set, after applying different strides to different regions. The distribution is much more equal.}
    \label{fig:distributionequal}
\end{figure}

20\% of the data was set aside as a validation set, and 80\% was used as training data. In this split, it was ensured that both sets contained roughly the same distribution for the labels. This can be seen in Figure \ref{fig:distribution}.
The stride of the sequence sampling is how much the window shifts to get the next sequence. This stride was made dependent on the label. For the labels that were far from the center a smaller stride was applied, and for the more common labels a larger stride was applied. This balanced the data out more. A minimum stride of 0.5 seconds was chosen, which is 10\% of the sequence length, to avoid different sequences from being too much alike which would result in overfitting. This stride of 0.5 seconds was applied to sequences with a label below 3 and above 8, as these were the most extreme regions with the lowest amount of samples. For labels between 3 and 5, and between 6 and 8, a stride of 2 seconds was applied. For the very common region between 5 and 6, a stride of 4 seconds was applied. These different strides resulted in a division as can be seen in Figure \ref{fig:distributionequal}. It is clear that the labels were now represented much more equally in the dataset.

\end{appendices}

\end{document}